\documentclass{article} 
\usepackage{iclr2027_conference,times}

\usepackage{amsmath,amsfonts,bm}

\def\eqref#1{equation~\ref{#1}}

\def\1{\bm{1}}

\def\mA{{\bm{A}}}
\def\mB{{\bm{B}}}

\def\mL{{\bm{L}}}
\def\mM{{\bm{M}}}

\def\mP{{\bm{P}}}

\DeclareMathAlphabet{\mathsfit}{\encodingdefault}{\sfdefault}{m}{sl}
\SetMathAlphabet{\mathsfit}{bold}{\encodingdefault}{\sfdefault}{bx}{n}

\usepackage{hyperref}
\usepackage{graphicx}
\usepackage{wrapfig}
\usepackage{subcaption}
\usepackage{booktabs}
\usepackage{multirow}
\usepackage{diagbox}
\usepackage{adjustbox}
\usepackage{array}
\usepackage{algorithm}
\usepackage{algorithmicx}
\usepackage{algpseudocode}
\usepackage{arydshln}
\usepackage{amssymb}

\title{Why Do Video Diffusion Models Violate Physics? Unveiling the Flaws in Attention Mechanisms}

\author{Yueyan Li, Haibo Wang, Caixia Yuan\thanks{Corresponding author: yuancx@bupt.edu.cn}, Xiaojie Wang \\
Beijing University of Posts and Telecommunications\\
\texttt{\{siriuslala,wanghb,yuancx,xjwang\}@bupt.edu.cn} \\
}

\iclrfinalcopy 
\begin{document}

\maketitle

\vspace{-10pt}
\begin{abstract}
\vspace{-5pt}
Despite impressive visual quality, state-of-the-art video diffusion models often generate content that violates real-world physical laws. While existing solutions rely on external priors or specialized data, we investigate the root cause by exploring the internal mechanisms of these models. Specifically, we present the first interpretability study on the ``motion planning'' process of text-to-video diffusion models, revealing how motion trajectories form during early denoising stages. Building upon the ``first shape, then details'' finding, we combine cross-attention trajectory patterns with causal head contributions to identify a specific subset of attention heads driving motion planning. Further, our self-attention analysis shows that Rotary Position Embedding (RoPE) induces excessive spatial attention decay. This causes early candidate regions to prematurely lock into physically implausible positions, suppressing reasonable trajectories in adjacent frames and triggering generation failure modes. To address this fundamental flaw, we propose a lightweight architectural modification that scales the frequency of RoPE across different denoising steps. This strategy reduces excessive attention decay, helping the model explore better candidate regions to establish coherent physical motion. Finally, training-free and training-based experiments confirm the effectiveness of our approach in enhancing the physical commonsense of generated videos \footnote{Code is available at \url{https://github.com/Siriuslala/physics}}.
\end{abstract}

\vspace{-10pt}
\section{Introduction}
\label{intro}
\vspace{-5pt}
Video generation models have advanced rapidly in recent years and have been hailed as a ``world simulator'' \citep{sora,veo3_1,seedance2_0}. However, although existing models perform well in aesthetics, motion stability, and instruction following, even the state-of-the-art models frequently generate content that violates real-world physical laws, reflecting a lack of \textit{physical commonsense} \citep{phygenbench,survey_physics}. Researchers have attempted to address this issue by introducing external physical priors, rewriting condition prompts, or adding specialized data rich in physical phenomena. However, these methods are either difficult to scale or do not resolve the problem from its root. Therefore, in this work, we shift our perspective to the interior of video diffusion models, aiming to understand the underlying mechanisms behind the failure modes of video generation and attempting to improve model architectures or algorithmic designs.

To ensure rigor, we need to clearly define what the ``physical commonsense'' of the model specifically refers to before commencing our study. According to \citet{phygenbench}, physical commonsense refers to the basic understanding of physical objects in daily life and the physical laws governing their interactions, mainly including mechanics, optics, thermodynamics, and material properties. Meanwhile, \cite{kangbingyi} primarily considered three categories of classical mechanics scenarios: uniform linear motion, perfectly elastic collision, and parabolic motion. In this work, we mainly focus on solid dynamics because it provides easily trackable motion trajectories, facilitating our investigation. For a video of $F$ frames, its underlying physical laws manifest as the continuous motion of objects in the video from frame 0 to frame $F-1$, externally reflecting a temporal evolution process. For video diffusion models, although physical knowledge is not explicitly introduced in their architectural design or training paradigms, they learn to induce physical laws from massive datasets via denoising during training in order to generate reasonable videos, thereby giving rise to the emergence of basic physical commonsense.
\begin{figure}[!t]
\centering
\includegraphics[width=1.0\columnwidth]{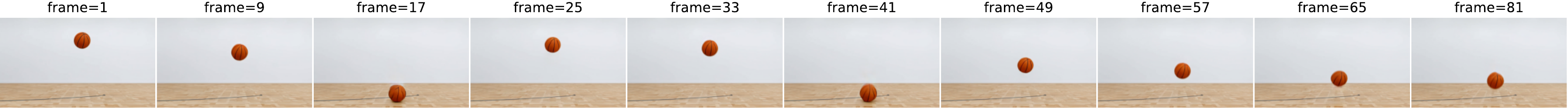}\\
\vspace{1pt}
\includegraphics[width=1.0\columnwidth]{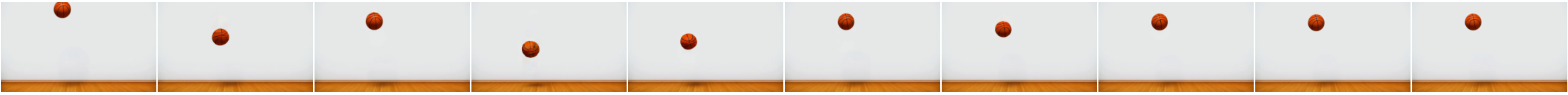}\\
\vspace{1pt}
\includegraphics[width=1.0\columnwidth]{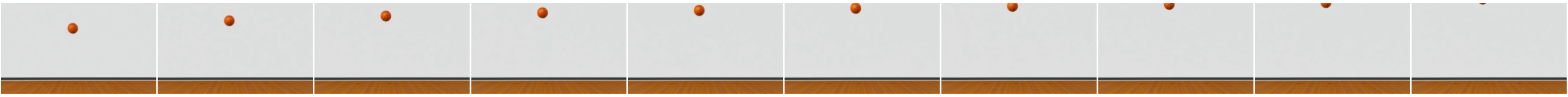}\\
\vspace{1pt}
\includegraphics[width=1.0\columnwidth]{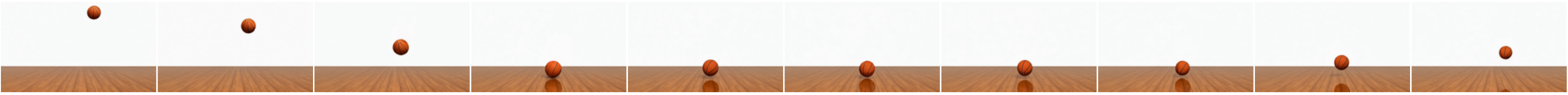}
\caption{From top to bottom are the videos generated by Wan-T2V-1.3B on an A800 GPU using seeds 26, 20, 8, and 29, respectively. Except for the first row, the others visibly violate physical laws, exhibiting behaviors such as mid-air bouncing, anti-gravity floating or sudden freezing.}
\label{fig:bad_case}
\vspace{-10pt}
\end{figure}

However, since existing models can hardly guarantee that all generated content strictly adheres to physical laws, we do not overemphasize physical commonsense for the time being, but instead first explore the mechanisms behind video generation. To sample a video containing object motion from Gaussian noise (regardless of whether the trajectory is physically reasonable), the model needs to: i) introduce semantic information from the condition into the video latent, and ii) allocate semantic information to different frames to determine the position of the object in each frame, while maintaining inter-frame coherence of the object motion as much as possible. We refer to the process covering the above two stages as the \textit{``motion planning’’ of video diffusion models}. Since this process is closely related to the model's physical commonsense, we focus our subsequent research here and pose three questions: \textit{(1) Where does motion planning happen? (2) How does this process happen? (3) Can we gain inspiration from it, such as better architectural designs or algorithmic optimizations?}

For (1) and (2), 
we start with a simple example of ``a basketball falling freely and bouncing'' based on Wan2.1-T2V-1.3B. 
While prior work \citep{sae_diffusion,towards_und,demystify} established that denoising follows a ``first shape, then details’’ progression where spatial layouts are finalized in early steps, they leave the underlying formation mechanisms unexplored. To bridge this gap, we extend this finding by exploring how motion planning unfolds from the view of model architecture.
Specifically, we first focus on cross-attention, as it is the sole source of video semantics. We observe the evolution of cross-attention maps in a layer from the video latent to object tokens across denoising steps. We find that an object possesses multiple \textit{candidate regions} per frame early on, which gradually converge into a deterministic shape at around step 5 (out of 50). 
We further quantify this process and find that not all attention heads exhibit a clear trajectory pattern.
To analyze head functions at a finer grain, 
we measure the \textit{convergence speed} of all heads toward the final trajectory.
Meanwhile, we design a causal intervention algorithm to measure the \textit{head contribution} to velocity prediction in flow matching.
Scatter plots in Figure~\ref{fig:ca_head_speed_vs_contri} show that, 
among all heads, only a small subset of heads with visible trajectory patterns impact motion planning, and having a clear trajectory pattern is not a sufficient condition for a head to be responsible for motion planning.

Next, we turn to self-attention, which is responsible for coordinating inter-frame relationships and serves as the underlying driver of the aforementioned findings. We address a critical question: \textit{how does the model select a deterministic object position from the candidate regions of each frame?} Since cross-attention patterns reflect self-attention outcomes, we first extract the candidate regions of each layer at each diffusion step based on the former. We then design a series of metrics based on self-attention to measure the confidence of each region. Visualization in Figure ~\ref{fig:candidate_evolution_seed20}, ~\ref{fig:sa_coupling_storyboard}, \ref{fig:candidate_evolution_seed26} reveals that the candidate regions in the early stages of denoising are in a highly sensitive state of competition, with no stable advantages or disadvantages. Once a few regions in certain frames gain higher confidence, their positions stabilize, prompting regions in other frames to put more attention to them. Crucially, if the early-stabilized positions are physically implausible, they can suppress regions with more reasonable positions in adjacent frames but larger relative distances due to RoPE-based attention decay. This could lead to content that violates physical laws, as illustrated in Figure ~\ref{fig:bad_case}.

Based on the above mechanistic analysis, we attribute most failure modes to the inflexibility of RoPE-induced attention decay along spatial dimensions in self-attention. As shown in Figure ~\ref{fig:sa_map}, attention attenuation along the height and width directions causes certain self-attention heads to excessively focus on the same regions across all frames, which in turn prevents some physically more appropriate candidate regions from receiving sufficient attention during the early stages of denoising. Therefore, for question (3), we propose a lightweight RoPE modification scheme, which sets different scalings for the frequency of RoPE across different denoising steps. This encourages the model to explore more candidate regions during motion planning by moderately reducing the attenuation speed of spatial attention. Both training-free and training-based experiments validate the effectiveness of this method. Overall, our contributions are as follows:
\begin{itemize}
    \item To the best of our knowledge, we present the first interpretability study on motion planning for text-to-video diffusion models, providing a practical analytical framework and toolkit.
    \item We extend the empirical finding of ``first shape then details’’ in reverse diffusion to a mechanistic level, showing from a more microscopic scale how the model forms the ``shape''.
    \item We uncover a hidden flaw in self-attention that triggers generation failure modes and enhance the physical commonsense of the model via lightweight modifications.
\end{itemize}

\section{Related Work}
\label{related}
\textbf{Intepretability for diffusion models.} Existing studies mainly focus on image generation. \citet{daam} attribute the influence of condition words on generated content via cross-attention. \citet{diff_interp_localize,diff_interp_circuit} locate knowledge in generative models via causal intervention. \citet{sae_diffusion,tide} study features inside diffusion models from a more granular perspective by training sparse autoencoders. Such studies are often associated with downstream applications such as image editing \citep{stable_flow,SAeUron,pci}. For video generation, \citet{temporal_correspondance} explore temporal correspondences across frames in video diffusion models. \citet{vid_diffusion_attn} study the impact of attention on video quality in text-to-video (T2V) tasks. \citet{maze_plan} discover the phenomenon of early plan in the maze solving task. Almost concurrently, \citet{demystify} propose ``Chain-of-Steps'' and find that several plausible paths emerge in parallel during early denoising in image-to-video (I2V) tasks.
However, they all stop at this finding, and the mechanism behind it remains unclear.

\textbf{Physics-aware video generation} aims to move beyond pixel-level visual fidelity and ensure object dynamics and interactions conform to real-world physical laws, serving as a critical step toward general-purpose world simulators. Existing studies fall into two paradigms. Explicit physics-driven methods integrate physics simulators as conditional guidance \citep{gpt4motion,physgen,motioncraft} or training constraints \citep{synthetic,newtongen}, offering precise physical control but suffering from limited generalizability and scalability. Implicit methods inject physical priors via curated datasets \citep{wisa}, LLM-guided prompt refinement \citep{phyt2v,vlipp}, or external foundation models \citep{videorepa,wwreward}, yet fail to address the root architectural cause of physical inconsistency. 
Others add specialized structures like physics experts \citep{prophy} or plug-in memory modules \citep{dit_mem}. These task-specific designs introduce extra computational overhead and lack flexibility, hindering general scaling. In contrast, we address a self-attention flaw by simply rescaling the frequency of RoPE. This lightweight adjustment enhances physical consistency while preserving native scalability.

\section{Preliminaries}
\label{headings}
This section briefly reviews the foundational framework of Flow Matching \citep{flow_matching} and the architecture of video diffusion models we study in this work. Flow matching provides a theoretically grounded framework for learning continuous-time generative processes in diffusion models. 
Specifically, given a data latent $x_1$ and a random noise $x_0 \sim \mathcal{N}(0, \mathbf{I})$, the Rectified Flow formulation defines an intermediate latent state $x_t$ at timestep $t \in [0, 1]$ via a linear interpolation: $x_t = (1 - t)x_0 + t x_1$.
The corresponding ground-truth velocity $v_t$ is defined as the time derivative of $x_t$, which simplifies to: $v_t = \frac{d x_t}{d t} = x_1 - x_0$.
To model the generative trajectory, a neural network $u(x_t, t, c; \theta)$ parameterized by $\theta$ is trained to predict this velocity field, conditioned on the intermediate state $x_t$, timestep $t$, and contextual conditioning $c$ (e.g., text embeddings). The optimization objective is formulated as the mean squared error (MSE) loss:
\begin{equation}
    \mathcal{L} = \mathbb{E}_{x_0, x_1, t, c} \left\| u(x_t, t, c; \theta) - v_t \right\|^2
\end{equation}
The model we study is based on Diffusion Transformer (DiT) \citep{dit}, represented by Wan2.1-T2V \citep{wan}. Given an input video $X \in \mathbb{R}^{(1+F) \times H \times W \times 3}$ with $1+F$ frames, height $H$, and width $W$, a 3D Variational Autoencoder (VAE) encodes it into a video latent. The latent is patchified and unfolded into a sequence of tokens $z_{\text{video}} \in \mathbb{R}^{(1+f)hw \times D}$. 
The conditioning prompt is embedded by a text encoder into a text embedding $z_{\text{text}} \in \mathbb{R}^{S \times D_{\text{T}}}$. 
In the DiT backbone, the video latent is first processed by bidirectional self-attention to achieve spatio-temporal interaction, then passes through cross-attention to acquire the semantics of the condition, and finally undergoes FFN to produce the output. The output of the final layer is linearly projected and normalized to yield the predicted velocity $u(z_t, t, c; \theta)$ for flow matching at each timestep $t$.
During inference, the predicted velocity is integrated via an ODE solver to generate the fully denoised latent, which is finally mapped back to the pixel space by the 3D VAE decoder to reconstruct the final video after $T$ denoising steps. Notations and model details are provided in Appendix~\ref{appn:notations} and~\ref{appn:wan_details}.

\section{Cross-Attention Mechanisms}
\label{sec:ca_interp}
Regarding motion planning, we first want to know where the moving object and the fixed background should respectively appear. Since cross-attention is the only module that can introduce condition semantics and allocate them to different regions of each frame of the video latent, we start here \footnote{Unlike I2V, directly decoding early latents into pixel-space in T2V introduces significant noise. Therefore, we investigate the denoising dynamics indirectly via cross-attention. See Appendix~\ref{appn:ca_maps} for more details.}. 

In the following analysis, we use Wan2.1-T2V-1.3B instead of 14B because we find that the model of larger size does not perform better in physics. The prompt of the main case for our analysis is ``\textit{Against a pure white background, a basketball falls vertically from mid-air onto a wooden floor and bounces up several times.}'', with a default random seed of 26 for 50-step inference. All experiments are conducted on an A800. Since the model uses classifier-free guidance (CFG) for generation, we study the conditional branch by default unless otherwise specified. See Appendix~\ref{appn:wan_details} for more details.
\subsection{Temporal Evolution of Cross-Attention}
\label{sec:ca_evolution}
\begin{figure}[!t]
\centering
\begin{tabular}{@{}m{0.04\columnwidth} @{\hspace{0.5pt}} m{0.955\columnwidth} @{}}
\centering\footnotesize T1 &
\includegraphics[width=\linewidth]{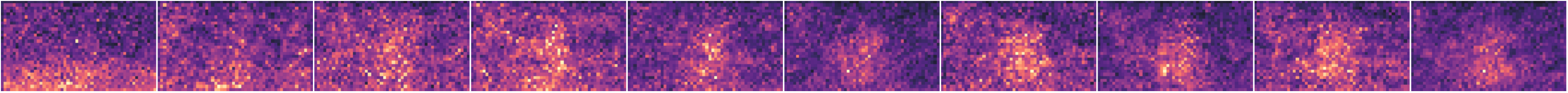} \\
\vspace{1pt}
\centering\footnotesize T3 &
\includegraphics[width=\linewidth]{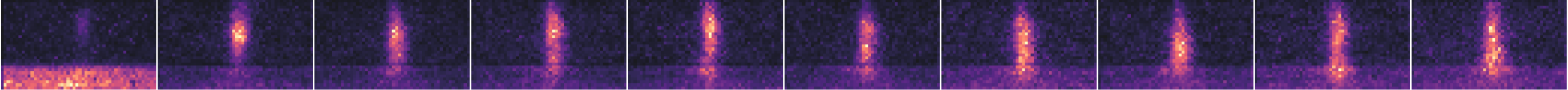} \\
\vspace{1pt}
\centering\footnotesize T5 &
\includegraphics[width=\linewidth]{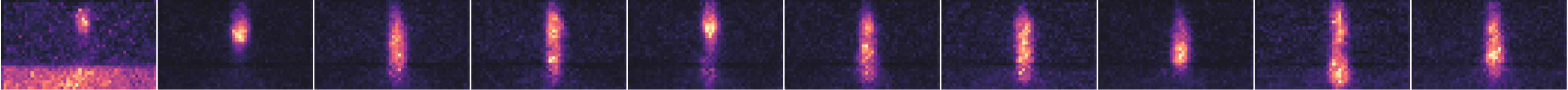} \\
\vspace{1pt}
\centering\footnotesize T7 &
\includegraphics[width=\linewidth]{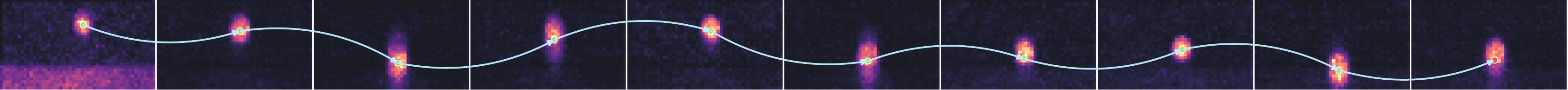} \\
\vspace{1pt}
\centering\footnotesize T50 &
\includegraphics[width=\linewidth]{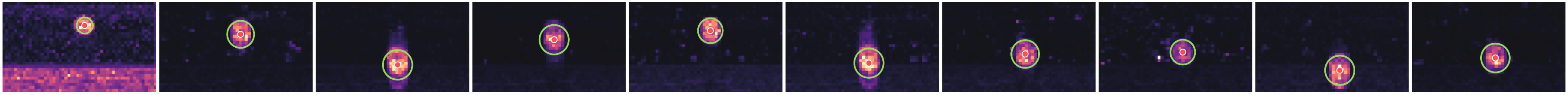} \\

\end{tabular}
\caption{Head-averaged cross-attention maps in layer 27. The trajectory appears within the first 5 denoising steps (annotated at T7). The green circle at T50 marks the reference area in Section~\ref{sec:ca_evolution}.}
\label{fig:ca_map}
\vspace{-10pt}
\end{figure}

As described in Section~\ref{intro}, the denoising process usually follows a ``first shape then details'' process, which is exactly the stage of the formation of the motion trajectory. Before a detailed analysis of this process, we first observe the overall evolution trend of the cross-attention map. We visualize the cross-attention $\mA^{\mathrm{CA}} \in \mathbb{R}^{f \times h \times w}$ of the video latent to the object token (e.g. ``basketball'') in the condition step-by-step and layer-by-layer. As shown in Figure~\ref{fig:ca_map}, we find that in the first 5 denoising steps, the position of the object in each frame changes from random noise (T1) to multiple highlighted regions distributed on the motion trajectory (T3), then gradually converges to the final position (T5), and finally displays a clear trajectory (T7).
To quantify this process, we design two metrics: \textit{\textbf{attention entropy}} and \textit{\textbf{support quality}}. Formally, let the spatial-temporal index set be $\mathcal{V}=\{1, \dots, f\} \times \Omega$, where the spatial index set $\Omega=\{1, \dots, h\} \times \{1, \dots, w\}$. For the attention map of each head, we first normalize it at the video-level: $\mP^{\mathrm{vid}}(z,y,x)=\frac{1}{\|\mA\|_1}\mA^{\mathrm{CA}}(z,y,x) \ (\sum_{(z,y,x) \in \mathcal{V}} \mP^{\mathrm{vid}}(z,y,x) = 1)$, where $\| \mA \|_1=\sum_{(z,y,x) \in \mathcal{V}} \mA^{\mathrm{CA}}(z,y,x)$. Then, the attention entropy is defined as the normalized spatial-temporal entropy: $\frac{H^{\mathrm{vid}}}{\log(fhw)}$, where $H^{\mathrm{vid}} = - \sum_{(z,y,x) \in \mathcal{V}} \mP^{\mathrm{vid}}(z,y,x) \log \mP^{\mathrm{vid}}(z,y,x)$.

Since attention entropy cannot capture the geometric structure in 2D space, we further use support quality to measure the overlap between the attention distribution and the final trajectory. As shown in the last row of Figure~\ref{fig:ca_map}, we first extract a binary support mask $\mM^{\mathrm{obj}} \in \{0,1\}^{f \times h \times w}$, which defines the final region of the object (\textit{\textbf{reference area}}) in each frame of the final trajectory (hereafter referred to as the \textit{reference trajectory}). Details of this process are provided in Algorithm~\ref{alg:support_mask}. Then, the video-level support quality is defined as: $Q^{\mathrm{vid}}=\sum_{(z,y,x) \in \mathcal{V}} \mP^{\mathrm{vid}}(z,y,x)\mM^{\mathrm{obj}}(z,y,x)$.

\begin{figure*}[t]
\centering
\begin{subfigure}{0.46\linewidth}
    \includegraphics[width=\linewidth]{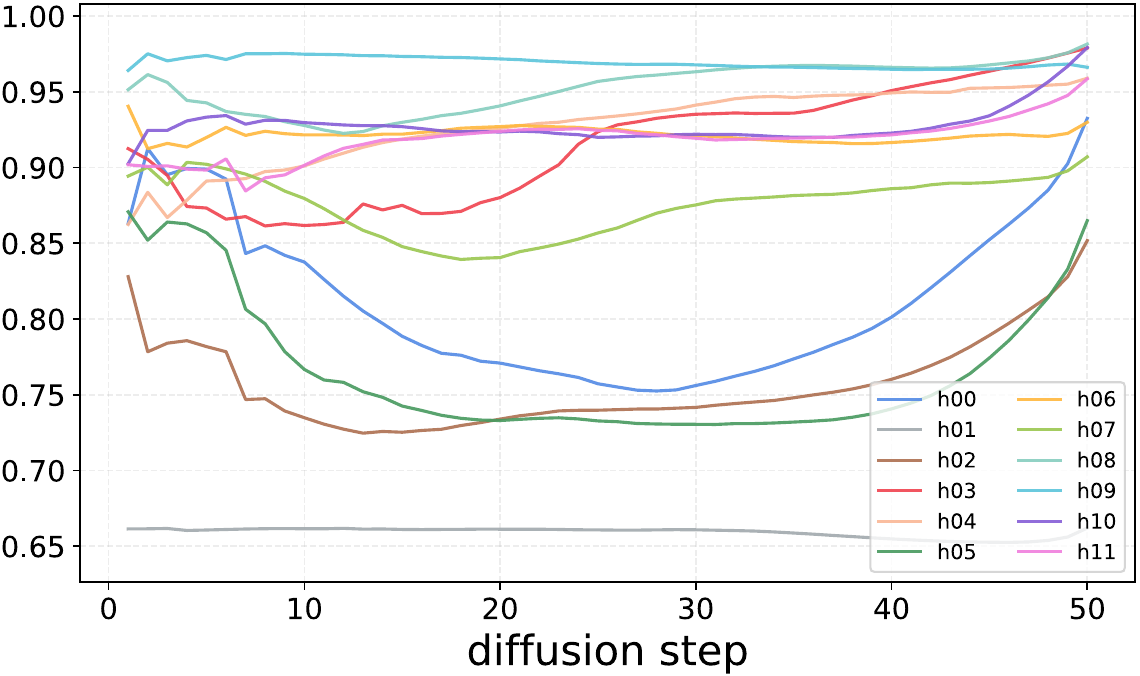}
    \caption{Video-level normalized entropy}
    \label{fig:ca_entropy_layer15}
\end{subfigure}
\hspace{10pt}
\begin{subfigure}{0.46\linewidth}
    \includegraphics[width=\linewidth]{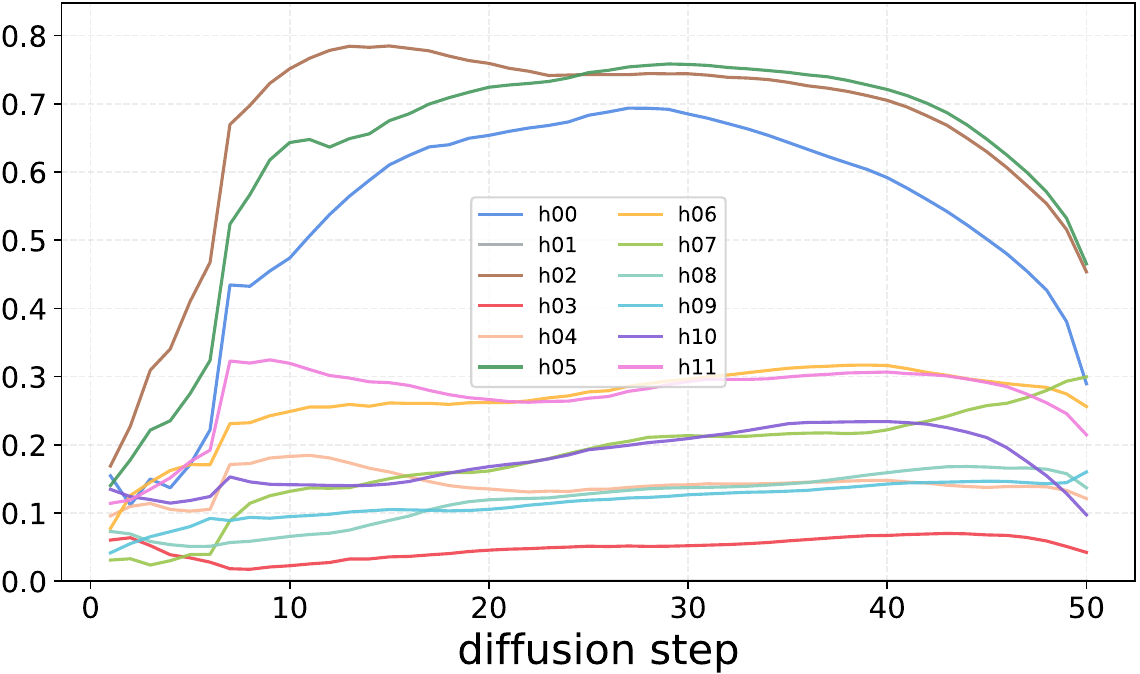}
    \caption{Video-level trajectory support quality}
    \label{fig:ca_support_quality_layer15}
\end{subfigure}
\vspace{-5pt}
\caption{The entropy and support quality of cross-attention heads in layer15. Overall, both metrics show distinct turning points around denoising step5, marking the emergence of the trajectory pattern.}
\label{fig:ca_temporal_evolution}
\vspace{-15pt}
\end{figure*}

Figure~\ref{fig:ca_temporal_evolution} shows the variation of the two metrics in layer 15 during denoising. We observe a jump of both metrics around denoising step 5. This indicates that most cross-attention heads focus their attention on the trajectory in the first 5 denoising steps, especially L15H2, L15H5, and L15H0 (Figure~\ref{fig:ca_map_l15h2}).
In addition, some heads do not show the above phenomenon. For example, although L15H1 maintains a low attention entropy, the support quality of this head remains 0, indicating that the attention distribution of this head is not aligned with the trajectory (Figure~\ref{fig:ca_map_l15h1}). This motivates us to identify the heads that truly affect the motion planning through a more detailed exploration.

\vspace{-5pt}
\subsection{Attention Heads for Motion Planning}
\label{sec:ca_motion_heads}
In this section, we aim to answer: which heads are responsible for motion planning? One idea is that such heads direct more condition semantics to the trajectory. Therefore, the attention of these heads from the video latent to the object token might concentrate more on the reference trajectory. However, does a head with an obvious trajectory pattern necessarily contribute significantly to motion planning? 
To answer this question, we define two additional metrics: \textit{\textbf{convergence speed}} and \textit{\textbf{head contribution}}. Convergence speed measures the speed of an attention pattern converging to the reference trajectory. It is defined as the mean of the support quality in the first 10 denoising steps. 

For head contribution, we adopt the idea of causal intervention \citep{causal_inf} and propose an attribution patching method for motion planning. Specifically, if we view the model $M$ as a directed acyclic graph, where each node $n$ is a component of the model (e.g., an attention head), we can measure the contribution of a node by ablating it and quantifying the change of the output before and after ablation through a patching metric $\mathcal{L}_m$. 
To focus the patching target more on motion planning, we first set the patching position to the reference area $\mM^{\mathrm{obj}}$, and then define the patching metric as:
\begin{equation}
\label{eq: head_contri}
\setlength{\abovedisplayskip}{4pt}
\setlength{\belowdisplayskip}{2pt}
    \mathcal{L}_m = \sum\nolimits_{p \in \mathcal{V}} \mM^{\mathrm{obj}}(p) \ [ u_t^{\mathrm{cond}}(p)^\top \cdot \operatorname{stopgrad}(\Delta u_t^{\mathrm{clean}}(p)) ]
\end{equation}
where $\Delta u_t^{\mathrm{clean}}\!=\!u_t^{\mathrm{cond, clean}} - u_t^{\mathrm{uncond, clean}} \in \mathbb{R}^{D}$ is the difference of the predicted velocity between the conditional branch and the unconditional branch at time step $t$ before ablation. This metric is intended to measure whether a head changes the condition-induced semantic increment written into the reference area. Besides, since this metric is merely an estimation of the actual impact of an attention head and is not necessarily completely accurate, we also ablate attention heads by setting the output of them to zero, and then directly observe the quality of the generated video. Details of the head contribution and the head zero ablation method are provided in Appendix~\ref{appn:head_contri} and \ref{appn:zero_ablation}.

\begin{wrapfigure}[]{l}{0.454\textwidth}
\centering
\begin{subfigure}{\linewidth}
    \includegraphics[width=\linewidth]{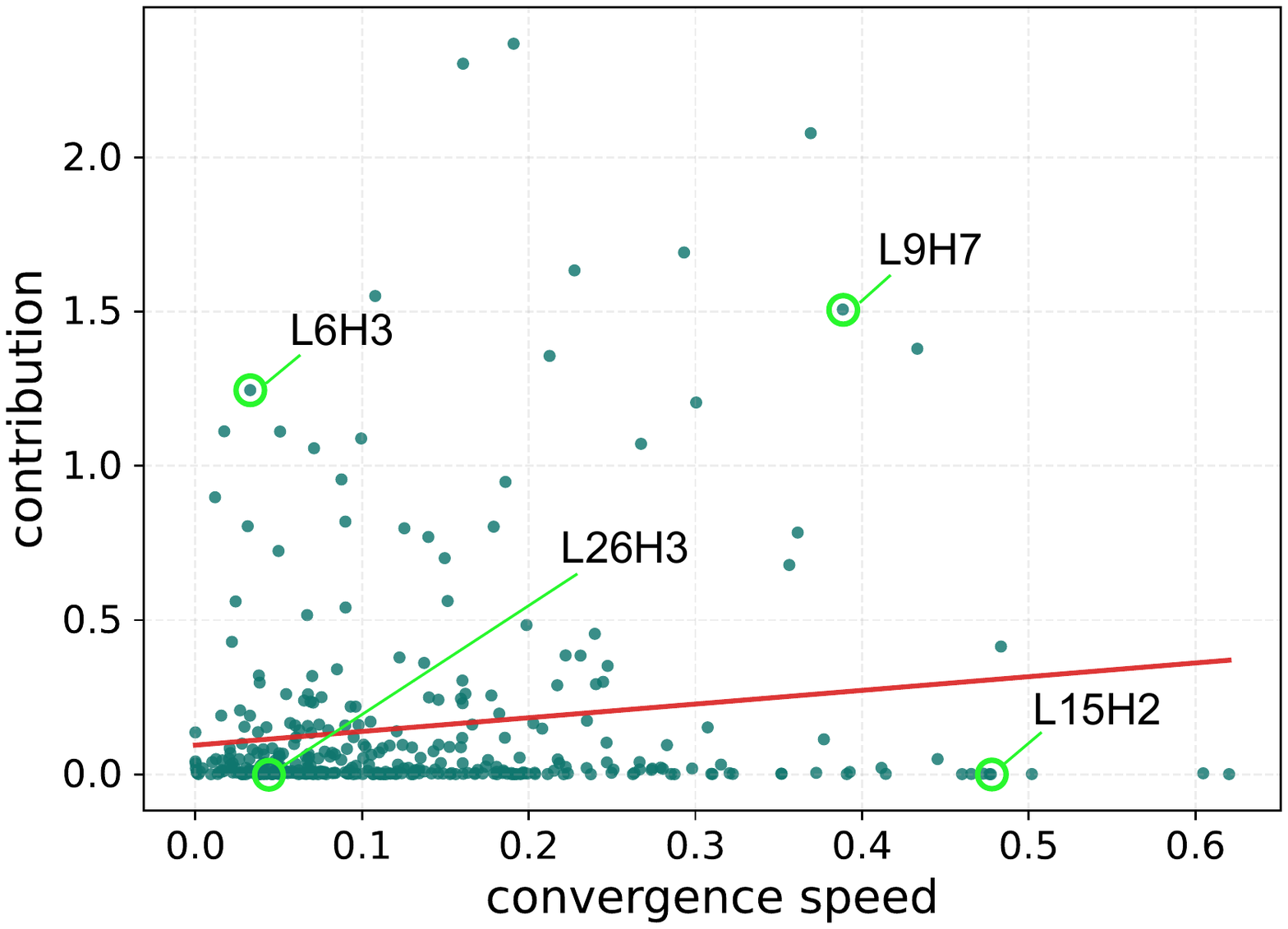}
\end{subfigure}
\vspace{-15pt}
\caption{Convergence speed vs. contribution of cross-attention heads at denoising step 2.}
\label{fig:ca_head_speed_vs_contri}
\vspace{-5pt}
\end{wrapfigure}

We visualize the results of convergence speed and head contribution in Figure~\ref{fig:ca_head_speed_vs_contri}. The scatter plot can be divided into 4 regions corresponding to 4 types of heads: (1) heads in the lower left corner with convergence speed $<$ 0.1 and contribution $<$ 0.5; (2) heads with convergence speed $<$ 0.1 but contribution $>$ 0.5; (3) heads in the upper right corner with convergence speed $>$ 0.1 and contribution $>$ 1.0; (4) heads close to the horizontal axis with convergence speed $>$ 0.1 but very low contribution.
\begin{figure}[!t]
\centering
\begin{tabular}{@{}m{0.04\columnwidth} @{\hspace{0.5pt}} m{0.955\columnwidth} @{}}

\centering\footnotesize (a) &
\includegraphics[width=\linewidth]{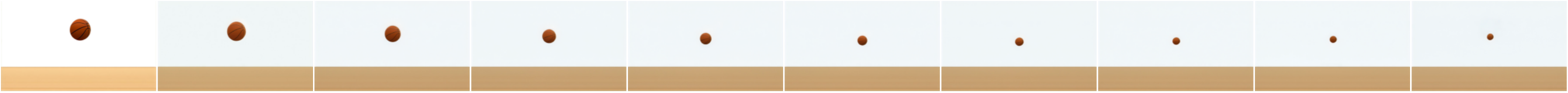} \\
\vspace{1pt}

\centering\footnotesize (b) &
\includegraphics[width=\linewidth]{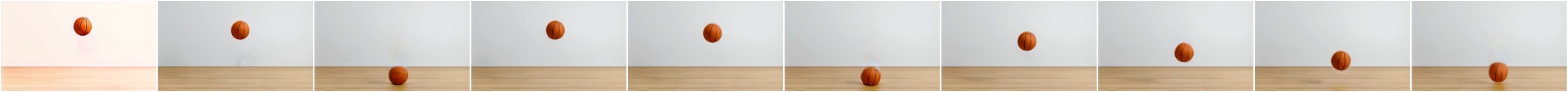} \\
\vspace{1pt}

\centering\footnotesize (c) &
\includegraphics[width=\linewidth]{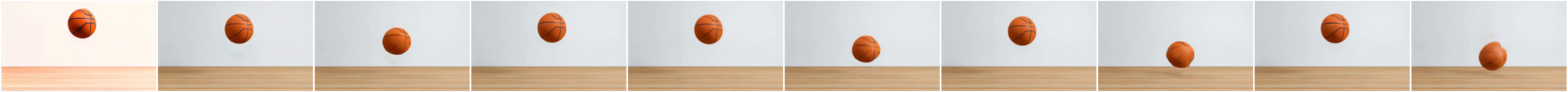} \\
\vspace{1pt}

\centering\footnotesize (d) &
\includegraphics[width=\linewidth]{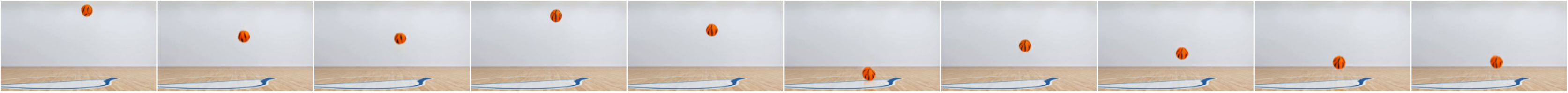} \\
\vspace{1pt}

\centering\footnotesize (e) &
\includegraphics[width=\linewidth]{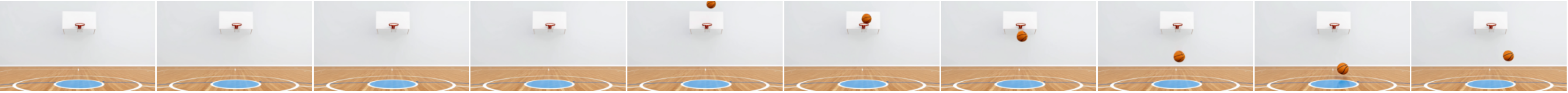} \\

\end{tabular}
\caption{The generated videos after the zero ablation of cross-attention heads in \ref{sec:ca_motion_heads}.}
\label{fig:ablate_results}
\vspace{-15pt}
\end{figure}

For Type (1) and (2), 
the attention patterns of these heads usually do not show a clear trajectory at denoising step 10. Most of them are relatively chaotic, or the highlights are in regions outside the object (Figure~\ref{fig:ca_map_examples} (a,b)). We first ablate Type (1) and find that the object has no displacement (Figure~\ref{fig:ablate_results} (a)). We hypothesize that this is because Type (1) contains many heads of layer 0 and 1, which appear at the very beginning of denoising and are important for the initialization of motion. Thus, we remove these heads from Type (1) and conduct the ablation again. As shown in Figure~\ref{fig:ablate_results} (b), the shape and the trajectory of the object have no changes.
When we ablate the heads of Type (2), we find identical results. Further, we ablate the heads of both Type (1) and (2) and find that except for some changes in the size of the object, the motion of it is basically not affected, as shown in Figure~\ref{fig:ablate_results} (c). This indicates that the impact of these heads focuses on static content such as the appearance of the object and the background, rather than motion planning. Even if the contribution is large, it does not mean that their real impact on the trajectory is large.

For the heads in Type (3) and (4), the attention patterns of them gradually show the trajectory during denoising (Figure~\ref{fig:ca_map_examples} (c)). We first ablate Type (3) and find that the trajectory of the object shows obvious collapse, as shown in Figure~\ref{fig:ablate_results} (e). Although these heads are rarer in quantity compared to Type (1) and (2), the impact of them is very large. This indicates that a clear trajectory pattern has a certain connection with motion planning. However, when we ablate Type (4), we find that although the initial position of the object has a slight shift compared to that before ablation, the overall trajectory has no changes (Figure~\ref{fig:ablate_results} (d)). This indicates that a clear trajectory pattern does not necessarily mean that this head is responsible for motion planning.
For the emergence of their trajectory pattern, the driving factor behind it is more that the Type (3) heads actively introduce motion-related semantic information into the position of the object in the residual stream of the video latent.
To verify the universality of the conclusion, we conduct the above experiments on other cases and random seeds. Results show that the classification of cross-attention heads holds in the absolute majority of scenarios. For more details and visualizations, please refer to Appendix~\ref{appn:ca_head_contri_details}.

\section{Self-Attention Mechanisms}
\label{sec:sa_interp}
\begin{figure}[!t]
\centering
\begin{tabular}{@{}m{0.04\columnwidth} @{\hspace{0.5pt}} m{0.955\columnwidth} @{}}
\centering\footnotesize T3 &
\includegraphics[width=\linewidth]{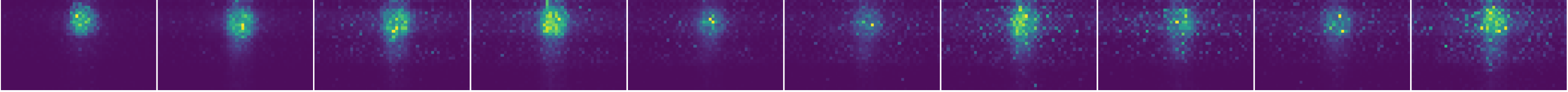} \\
\vspace{1pt}
\centering\footnotesize T20 &
\includegraphics[width=\linewidth]{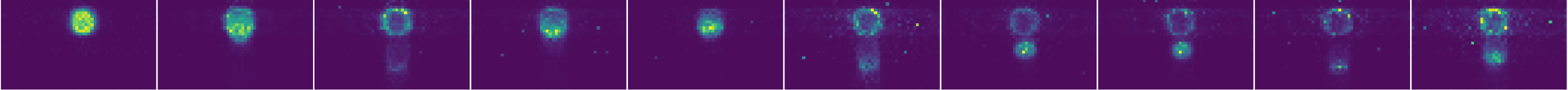} \\

\end{tabular}
\vspace{-10pt}
\caption{Head-averaged self-attention map from a region in frame 0 to other frames in layer 20.}
\label{fig:sa_map}
\vspace{-5pt}
\end{figure}

Cross-attention informs the model of the content to generate, but cannot reasonably arrange the content in video frames. Self-attention is the only module that enables communication among video tokens, thereby determining the position of the entity in the condition in different frames. It is the direct cause of the convergence of the cross-attention patterns to the object trajectory. In Section~\ref{sec:ca_interp}, we observe that in the early stage of denoising, one or multiple possible positions of the object simultaneously exist in each frame of the video latent, which we term as \textit{\textbf{candidate regions}}. In this section, we aim to answer: how does the model successfully select a physically plausible position for the object from the candidate regions, or \textit{how does it fail}? 
To begin with, we first observe the self-attention map. We randomly select a part from the query (e.g., the reference area in a frame) and visualize its attention to all frames. We observe a phenomenon different from cross-attention: as shown in Figure~\ref{fig:sa_map}, most self-attention does not clearly exhibit a clear trajectory pattern, but instead focuses on the same position of different frames. Specifically, for region $k$ in frame $f_i$ ($0 \le i < f$), its attention to other frames concentrates within region $k$ of other frames, while the attention to other regions outside of $k$ is small. We attribute this to the decay of RoPE in the spatiotemporal dimension. Taking a query $q=[q^f, q^h, q^w]$ at position $p=(p^f, p^h, p^w)$ as an example, RoPE injects positional information into it via $f(q, p)=[q^f e^{i p^f \theta}, q^h e^{i p^h \theta}, q^w e^{i p^w \theta}]$ (where $q^{f/h/w}$ are 2D vectors; see Appendix~\ref{appn:rope} for details). Consequently, the dot product of query $q$ and key $k$ can be expressed as: $\operatorname{Re}[\sum_{a \in \{f,h,w\}} q^{a^{}} {k^{a}}^{*} e^{i \Delta p^a \theta}]$. Due to long-range decay, region $a$ favors regions across all frames that are spatially close to itself, i.e., those with small $\Delta p^h, \Delta p^w$. This property is intuitively reflected in Figure~\ref{fig:rope_spatial_decay_map}, and we call it the \textit{\textbf{spatial anchoring effect}} of RoPE. Consequently, in the middle and late stages of denoising, even if some self-attention heads can display the trajectory, additional highlighted regions due to the spatial anchoring effect still exist outside of the trajectory.

\begin{figure}[!t]
\centering
\includegraphics[width=1.0\columnwidth]{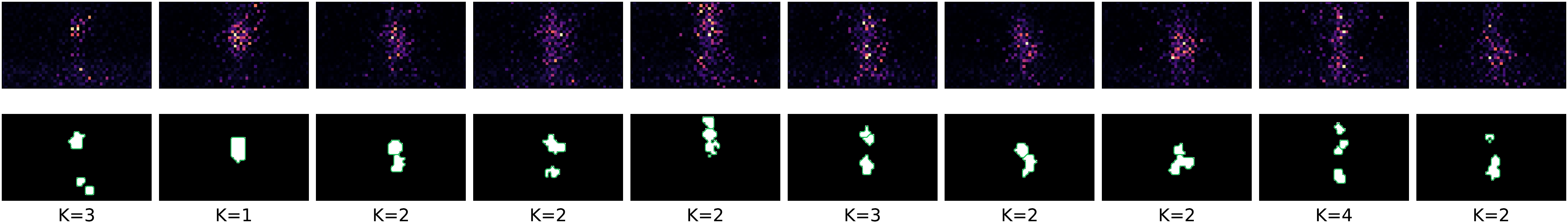}
\vspace{-15pt}
\caption{Candidate regions (bottom) extracted from the head-averaged cross-attention in layer 16.}
\label{fig:candidate_region_contour}
\vspace{-15pt}
\end{figure}

\begin{figure}[!b]
\vspace{-10pt}
\centering
\includegraphics[width=1.0\columnwidth]{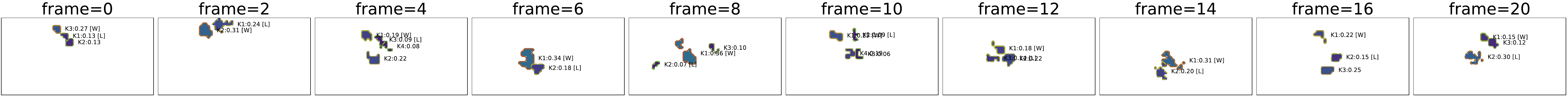}
\includegraphics[width=1.0\columnwidth]{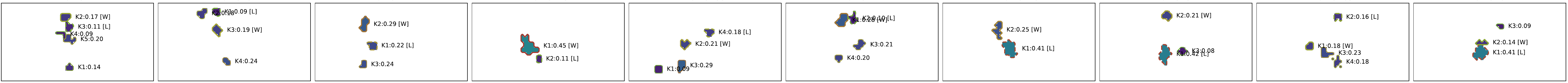}
\includegraphics[width=1.0\columnwidth]{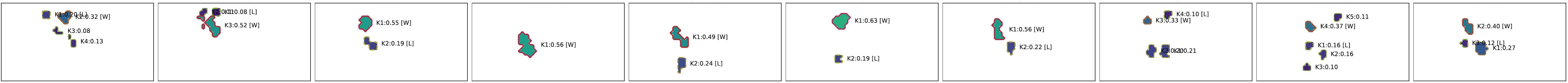}
\includegraphics[width=1.0\columnwidth]{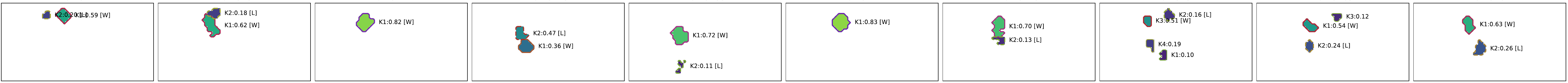}
\includegraphics[width=1.0\columnwidth]{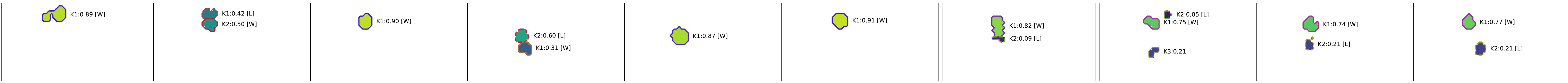}
\includegraphics[width=1.0\columnwidth]{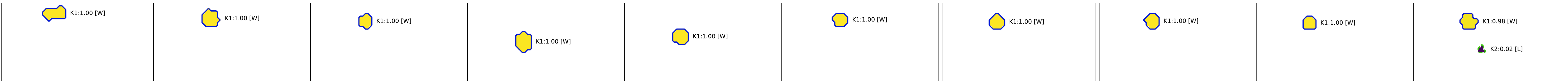}
\includegraphics[width=1.0\columnwidth]{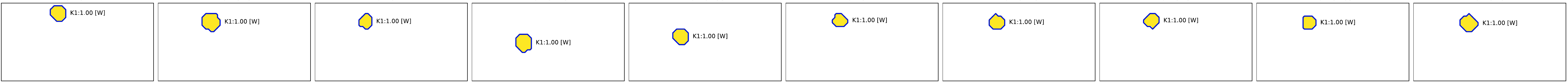}
\vspace{-15pt}
\caption{The evolution of candidate regions during denoising. Better viewed when zoomed in.}
\label{fig:candidate_evolution_seed20}
\vspace{-10pt}
\end{figure}
To quantify the dynamics of early candidate regions, we extract $\{\Omega_{i,k}\}_{k=1}^{K_i}$ for each frame $f_i$ from the head-averaged cross-attention map of each layer at each denoising step via Algorithm~\ref{alg:candidate_region_extraction}, as shown in Figure~\ref{fig:candidate_region_contour}. 
We use the candidates at step 49, layer 27 as the anchor regions $\{\Omega_{i,1}^{\mathrm{ref}}\}_{i=1}^{f}$, where each frame has only 1 region. 
For each candidate $\Omega_{i,k}$ in frame $f_i$, we calculate the \textit{anchor-distance}: $d_k = \Vert{} \mu(\Omega_{i,k}) - \mu(\Omega_{i,1}^{\mathrm{ref}}) \Vert{}$, where $\mu(\cdot)$ denotes the geometric center of a region. Then, the \textit{\textbf{winner}} $k^{+} = \arg\min_k d_k$ is the candidate closest to the reference area in frame $f_i$, with the \textit{\textbf{strongest loser}} as the non-winner candidate with the smallest anchor-distance.
Next, we need to measure how self-attention selects a candidate during denoising. 
Consider region $\Omega_{i,k_i}$, $\Omega_{j,k_j}$ in frame $f_i$, $f_j$. 
We first define the normalized self-attention intensity of $\Omega_{i,k_i}$ to $\Omega_{j,k_j}$ as $\widetilde C =\frac{C(\Omega_{i,k_i} \to \Omega_{j,k_j})}{M(\Omega_{i,k_i} \to f_j)}$, 
where $C(\Omega_{i,k_i} \to \Omega_{j,k_j})=\frac{1}{\vert{}\Omega_{i,k_i}\vert{}} \sum_{i \in \Omega_{i,k_i}, j \in \Omega_{j,k_j}} \mA_{ij}^{\mathrm{SA}}$ and $M(\Omega_{i,k_i} \to f_j)= \sum_{i \in \Omega_{i,k_i}, j \in \Omega_{j, \cdot}} \mA_{ij}^{\mathrm{SA}}$. 
Then, we define \textit{\textbf{mutual consistency}} as $\mathrm{MC}(\Omega_{i,k_i},\Omega_{j,k_j}) \!=\! \overline C(\Omega_{i,k_i} \!\to\! \Omega_{j,k_j})\,\overline C(\Omega_{j,k_j} \!\to\! \Omega_{i,k_i})$, where $\overline C$ is the average of $\widetilde C$ over self-attention heads. 
This metric measures the confidence of the mutual selection between two candidates.
For region $\Omega_{i,k_i}$, 
the mutual consistency of it is defined as: $\mathrm{MC}(\Omega_{i,k_i}) \!=\! \frac{1}{f} \sum_{j=1}^{f} \max_{k_j} \mathrm{MC}(\Omega_{i,k_i},\Omega_{j,k_j})$.
A higher value indicates that the region is more likely to be the winner.
To validate this metric, we plot mutual consistency vs. anchor distance for each denoising step. As shown in Figure~\ref{fig:mutual_consistency}, their correlation strengthens as denoising progresses, showing that mutual consistency can measure the confidence of self-attention for each candidate.

To study why an early candidate region later wins or fails, we visualize the dynamics of denoising. We select a failed case of seed 20, where the basketball bounces in the air and then stops.
Figure~\ref{fig:candidate_evolution_seed20} shows the candidate regions of layer 16 with their mutual consistency in the first 7 steps. [W] and [L] denote the winner and the strongest loser. Brighter regions indicate higher mutual consistency. We find that: \textbf{(1)} In early denoising, multiple candidate regions compete unstably with no clear advantage, causing the winner-loser gap to oscillate around 0 (see Figure~\ref{fig:mutual_consistency_layerwise}). \textbf{(2)} A small number of frames determine the position of the object earlier, such as frames 4, 8, and 10. \textbf{(3)} Early dominant and physically reasonable regions can be suppressed later. For instance, region K1 in frame 14 dominates at step 2, but its advantage is ``robbed'' by region K3 at step 3 (K1: 0.21 vs. K3: 0.33). 

\begin{figure}[!t]
\centering
\begin{tabular}{@{}m{0.04\columnwidth} @{\hspace{0.5pt}} m{0.955\columnwidth} @{}}

\centering\footnotesize T2 &
\includegraphics[width=\linewidth]{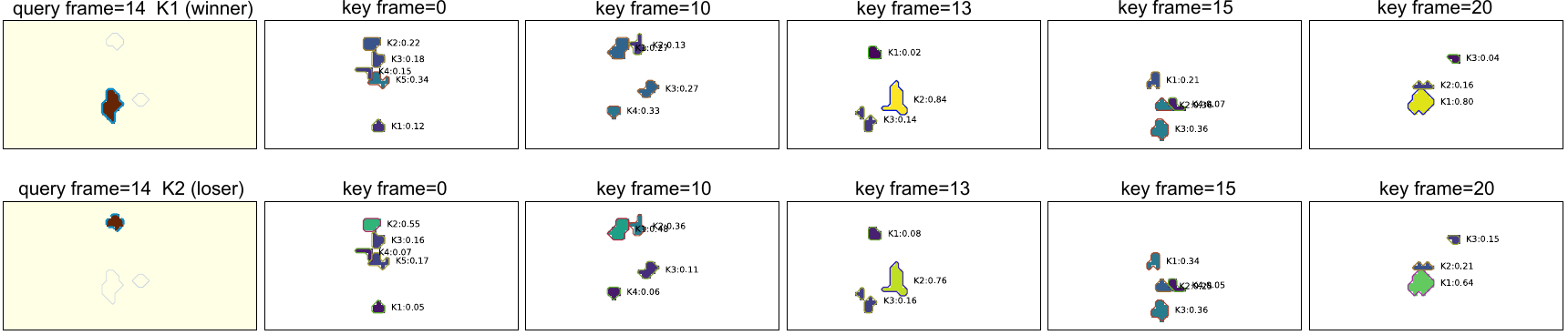}\vspace{3pt} \\

\hdashline
\centering\footnotesize T3 &
\vspace{3pt}\includegraphics[width=\linewidth]{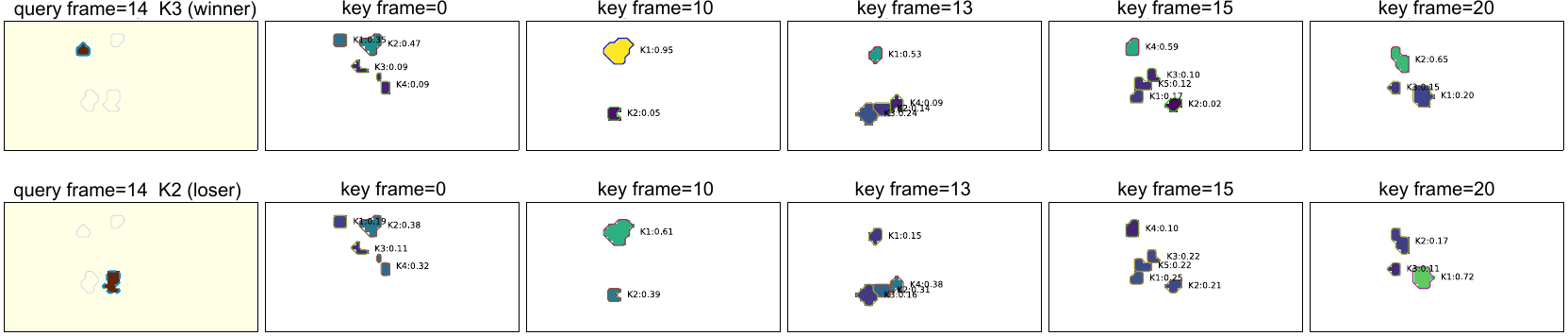} \\

\end{tabular}
\vspace{-5pt}
\caption{The mutual consistency between regions in frame 14 and others at denoising step 2 and 3.}
\label{fig:sa_coupling_storyboard}
\vspace{-15pt}
\end{figure}

Detailed in Figure~\ref{fig:sa_coupling_storyboard}, both K1 and K3 in frame 14 tend to attend to regions in other frames close to their respective positions: K1 focuses on the lower regions in each frame at step 2, while K3 targets the upper regions at step 3. This behavior stems from the spatial anchoring effect of self-attention. Since region K1 in frame 10 stabilizes its position early at step 3 with a high confidence of 0.63, the same effect steadily raises confidence of spatially adjacent regions in other frames (e.g., K3 in frame 14), driving them to align with its position and even overriding the model’s physical prior. This enables K3 of frame 14 to surpass K1, leading to the failure mode where the basketball remains stationary in the air.
The other failure cases in Figure~\ref{fig:bad_case} share this mechanism.

\vspace{-5pt}
\section{RoPE Modification}
\vspace{-5pt}
\label{sec:rope_modify}
\subsection{Method}
\vspace{-5pt}
In this section, we leverage the conclusions derived from the mechanistic analysis to improve the model architecture, thereby optimizing the physical commonsense of the model. Our idea is straightforward: to mitigate the spatial anchoring effect found in Section~\ref{sec:sa_interp}, we weaken the spatial decay of RoPE to encourage the model to explore more candidate regions of moving objects during motion planning. Formally, for a query $q=[q^f, q^h, q^w]$ at position $p=(p^f, p^h, p^w)$, we apply modulation factors $\lambda^h, \lambda^w < 1.0$ on the height and width dimensions in the RoPE transformation of it:
\begin{equation}
\setlength{\abovedisplayskip}{4pt}
\setlength{\belowdisplayskip}{4pt}
    f^h(q, p) = q^he^{ip^h\lambda^h\theta}, f^w(q, p) = q^we^{ip^w\lambda^w\theta}.
\end{equation}
This formulation is equally applicable to the key in self-attention, and more details of RoPE are provided in Appendix~\ref{appn:rope}. 
This method mitigates the decay rate of RoPE by reducing the magnitude of the rotation angle.
In practice, we explored various approaches to parameterize $\lambda^{h/w}$, among which some methods seemed reasonable but yielded poor performance. In the main text, we search for a preset  $\lambda^{h/w}$ on the first 5 denoising steps for both training-free and training-based methods. For other approaches (e.g., adaptively adjusting $\lambda^{h/w}$), we discuss in detail in Appendix~\ref{appn:rope_modify_train_lambda}.

\begin{table}[htbp]
\centering
\caption{Experimental results of modified RoPE on VideoPhy. ``PR'' means \textit{prompt refinement}.}
\vspace{-5pt}
\label{tab:model_result}
\small
\setlength{\tabcolsep}{4.5 pt}
\resizebox{\columnwidth}{!}{
\begin{tabular}{l cccccccc}
\toprule
\multirow{2}{*}{\diagbox[width=3.2cm]{Model}{Metrics (\%)}}
& \multicolumn{2}{c}{Overall}
& \multicolumn{2}{c}{Solid-Solid}
& \multicolumn{2}{c}{Solid-Fluid}
& \multicolumn{2}{c}{Fluid-Fluid} \\
\cmidrule(l{0.4em}r{0.4em}){2-3}
\cmidrule(l{0.4em}r{0.4em}){4-5}
\cmidrule(l{0.4em}r{0.4em}){6-7}
\cmidrule(l{0.4em}r{0.4em}){8-9}
& SA & PC & SA & PC & SA & PC & SA & PC \\
\midrule
Wan2.1-T2V-1.3B & 53.64 & 29.45 & 46.89 & 24.52 & 55.89 & 31.41 & 63.74 & 36.55 \\
Wan2.1-T2V-1.3B-PR \citep{phyt2v} & 80.47 & 44.31 & 70.48 & 28.88 & 87.30 & 53.47 & 90.94 & 61.87 \\
Wan2.1-T2V-1.3B-LoRA \citep{lora} & 55.69 & 35.57 & 47.82 & 27.95 & 61.32 & 38.97 & 63.74 & 45.61 \\
Wan2.1-T2V-1.3B-VideoREPA \citep{videorepa} & 56.56 & 37.90 & 48.69 & 30.61 & 62.68 & 42.57 & 64.72 & 46.65 \\
\midrule
Wan2.1-T2V-1.3B-modified RoPE & 57.43 & 39.94 & 48.75 & 32.60 & 66.46 & 48.33 & 61.87 & 43.74 \\
Wan2.1-T2V-1.3B-modified RoPE+PR & 86.30 & 58.89 & 82.92 & 45.32 & 87.30 & 68.88 & 90.94 & 72.81 \\
Wan2.1-T2V-1.3B-LoRA+modified RoPE & 64.72 & 41.98 & 54.33 & 35.39 & 77.94 & 49.54 & 67.48 & 45.61 \\
Wan2.1-T2V-1.3B-LoRA+modified RoPE+PR & \textbf{87.46} & \textbf{62.68} & \textbf{83.85} & \textbf{50.61} & \textbf{87.30} & \textbf{72.81} & \textbf{94.68} & \textbf{72.81} \\
\bottomrule
\end{tabular}
}
\vspace{-10pt}
\end{table}

\subsection{Implementation details}
\vspace{-1pt}
For evaluation, we first test our method on the ``basketball free fall'' case with various random seeds for quick verification, 
and then use \textbf{VideoPhy} \citep{videophy} for systematic evaluation, which contains 343 test cases that involve interactions between various material types in the physical world (e.g., solid-solid, solid-fluid, fluid-fluid). We use Semantic Adherence (SA) and Physical Commonsense (PC) defined by the authors as metrics.
For training, we utilize \textbf{WISA} \citep{wisa} which contains 80K human-curated videos of 17 physical laws. See Appendix~\ref{appn:rope_modify_data} for details.

We validate our method under both training-free and training-based settings. For inference, we only apply the RoPE modification during the first 5 steps of denoising. For training, we design a customized timestep sampler, which samples the first 10\% denoising timesteps with a high probability $p^{\mathrm{early}}$, thereby focusing the training on the early stage of motion planning in denoising. 
We slowly increase $\lambda^{h/w}$ from 0 to a predefined value (0.75 by default) and fine-tune the model via LoRA with a batch size of 32 and a learning rate of 1e-4. Experimental settings are detailed in Appendix~\ref{appn:rope_modify_train}.

\vspace{-4pt}
\subsection{Main results \& Analysis}
\vspace{-1pt}
For comparison, training-free baselines include the base model and prompt refinement (following the method in \citep{phyt2v}), while training-based methods include LoRA \citep{lora} and VideoREPA \citep{videorepa} which uses an external video foundation model for guidance during fine-tuning.
We first tested the ``basketball free fall'' case.
As shown in Figure~\ref{fig:rope_res_basketball}, our method indeed improves the motion trajectories, but requires manual tuning of $\lambda^{h/w}$. Therefore, we adopt fine-tuning to enable the model to gradually adapt to $\lambda^{h/w}$. As shown in Table~\ref{tab:model_result}, the training-based results are generally superior to those of training-free and other baselines in terms of both instruction following and physical consistency. Specifically, the advantage of our method is mainly reflected in the \textit{solid-*} subset, as it contains more \textit{large-magnitude motions} compared with the \textit{fluid-fluid} subset, which is precisely the focus of our method.
Ablation studies on the choice of $\lambda^{h/w}$, as well as $p^{\text{early}}$ in our custom sampler are detailed in Appendix~\ref{appn:rope_modify_train}. We also performed validation on a larger model of size 14B. 
Comparisons of the generated videos are shown in Appendix~\ref{appn:rope_modify_viz_videophy}.

Beyond the main results, we provide another 3 key findings: \textbf{(1)} Physical consistency improvements brought by prompt refinement mainly stem from enhanced instruction following, while our method can be combined with it to further boost physical consistency. \textbf{(2)} We train the model under a fixed random seed of 42, but the trained model improves the motion trajectories generated by other seeds with the same $\lambda^{h/w}$, as shown in Figure~\ref{fig:rope_res_train_basketball}, suggesting the good generalizability of our method. 
\textbf{(3)} For LoRA, tuning only the attention modules yields better results without degrading aesthetics, indicating that the physical failures stem more from attention rather than FFN, which guarantees the completeness of our interpretability analysis. This is discussed in detail in Appendix~\ref{appn:rope_modify_train_lora_config}

\vspace{-5pt}
\section{Limitations and Discussions}
\vspace{-5pt}
In conclusion, starting from the finding of ``first shape, then details'', we present a comprehensive interpretability analysis of motion planning in video diffusion models. Based on this, we identify a flaw of self-attention during the denoising process and propose a RoPE modification method, which is proven to make generated video content adhere better to physical laws with only a scaling factor. Moving beyond this, our findings can be extended to broader scenarios (e.g., I2V) and other model architectures (e.g.,  few-step autoregressive diffusion models), which we will explore in future work.

\subsection*{AI use statement}
In this work, we used generative AI tools for proposing and refining hypotheses regarding the internal mechanisms of video generation models. We have not used generative AI tools for designing research methodology or experiments, implementing methods, or interpreting results, and the rest of the required disclosure tasks are not applicable to this work. Additionally, we used generative AI tools for creating and editing software code and editing the research paper to improve readability. We have reviewed all AI-assisted work: all AI-proposed hypotheses were independently verified through rigorous empirical experiments, all code was thoroughly tested and verified by the authors, and all polished text was reviewed to ensure accuracy. We take responsibility for the final content of this work, including text, claims or artifacts produced with the aid of generative AI.

\subsection*{Ethics statement}
This work complies with the Code of Ethics. Our study focuses on understanding the internal mechanisms and improving the physical consistency of video generation models. All experiments utilize open-source models and publicly available benchmarks. For human evaluation, assessments were conducted voluntarily by qualified researchers. No personally identifiable information was collected, and the evaluated video prompts contained no harmful, sensitive, or offensive content.

\subsection*{Reproducibility statement}
To ensure the reproducibility of our results, complete source code covering both the interpretability analysis and the interpretability-guided architectural modifications is provided in the Supplementary Material. The base models used throughout our experiments (Wan2.1-T2V-1.3B/14B) are officially open-sourced on Hugging Face, and their detailed architecture specifications are elaborated in Appendix~\ref{appn:model_details}. The formal mathematical modeling and derivations of the RoPE analysis introduced in Section~\ref{sec:sa_interp} and \ref{sec:rope_modify} are detailed in Appendix~\ref{appn:rope}. Furthermore, comprehensive experimental setups for Section~\ref{sec:rope_modify} including the parameterization of our method, training data preprocessing pipelines, the choice of hyperparameters, and evaluation protocols are fully described in Appendix~\ref{appn:rope_modify}.



\bibliography{iclr2027_conference}
\bibliographystyle{iclr2027_conference}

\newpage
\appendix
\section{Notations}
\label{appn:notations}
\begin{table}[htbp]
  \centering
  \caption{Summary of Notations}
  \label{tab:notations}
  \begin{tabular}{l l}
    \toprule
    Symbol & Description \\
    \midrule
    $F$ & Number of frames in a video \\
    $H$ & Height of a video \\
    $W$ & Width of a video \\
    $f$ & Number of frames in a video latent \\
    $h$ & Height of a video latent \\
    $w$ & Width of a video latent \\
    $T$ & Number of denoising steps \\
    $N_L$ & Number of layers in DiT \\
    $N_H$ & Number of self / cross-attention heads \\
    $D$ & Model dimension (dimension of the residual stream in DiT) \\
    $D_H$ & Dimension of a self / cross-attention head \\
    $D_T$ & Dimension of the text embedding \\
    L$x$H$y$ & Head $y$ in layer $x$ \\
    $\mA^{\mathrm{CA}}$ & Cross-attention map of a head \\
    $\mA^{\mathrm{SA}}$ & Self-attention map of a head \\
    $\Omega$ & Spatial index set $\{1, \dots, h\} \times \{1, \dots, w\}$ \\
    $\mathcal{V}$ & Spatial-temporal index set $\{1, \dots, f\} \times \Omega$ \\
    \bottomrule
  \end{tabular}
\end{table}

\section{Model details}
\label{appn:wan_details}

\begin{table}[t]
\centering
\small
\caption{Main architecture and inference parameters of Wan2.1-T2V-1.3B.}
\label{tab:wan21_t2v_13b_params}
\begin{tabular}{lll}
\toprule
Parameter & Value & Description \\
\midrule
Video shape & \(81\times480\times832\times3\) & Default size of generated video ($F \times H \times W \times C$) \\
Frame rate & \(16\) fps & Default sample FPS in shared config \\
Text encoder & UMT5-XXL encoder & Frozen prompt encoder \\
\(S\) & \(512\) & Maximum text sequence length \\
\(D_T\) & \(4096\) & UMT5-XXL text embedding dimension \\
\(C_z\) & \(16\) & VAE latent channel dimension \\
VAE stride & \((4,8,8)\) & Temporal, height, and width downsampling ratio of VAE \\
VAE latent shape & \(16\times21\times60\times104\) & Shape of \(\bar z_t\) before DiT patchification \\
Patch size & \((1,2,2)\) & DiT 3D patch size and stride in latent space \\
DiT token grid & \(21\times30\times52\) & Patchified spatio-temporal grid for \(832\times480\) output \\
\(L\) & \(32760\) & Token sequence length, \(L=(1+f)hw=21\times30\times52\) \\
\(N_L\) & \(30\) & Number of DiT transformer layers \\
\(N_H\) & \(12\) & Number of self/cross-attention heads \\
\(D\) & \(1536\) & Model dimension (dimension of the residual stream in DiT ) \\
\(D_H\) & \(128\) & Attention head dimension, \(D_H=D/N_H\) \\
\(D_{\mathrm{ffn}}\) & \(8960\) & FFN intermediate dimension \\
\(D_{\mathrm{freq}}\) & \(256\) & Sinusoidal timestep embedding dimension before time MLP \\
Inference steps \(T\) & \(50\) & Default T2V sampling steps \\
Default solver & FlowUniPC (default) & Sampler \\
Sample shift & \(5.0\) (default) & Flow-matching timestep schedule shift \\
$s_\mathrm{CFG}$ & \(5.0\) (default) & Classifier-free guidance scale \\
\bottomrule
\end{tabular}
\end{table}

\paragraph{Wan2.1-T2V architecture.}
\label{appn:model_details}
We study the Wan2.1-T2V-1.3B model, whose denoising backbone is a Diffusion Transformer operating on a patchified video latent. For clarity, we distinguish the VAE latent grid from the DiT token grid. Given a video \(X \in \mathbb{R}^{(1+F)\times H\times W\times 3}\), the Wan-VAE maps it to a latent tensor \(\bar z_0 \in \mathbb{R}^{C_z \times (1+f) \times 2h \times 2w}\), where \(C_z=16\), \(1+f=(F/4)+1\), \(2h=H/8\), and \(2w=W/8\). Equivalently, after the DiT patch embedding with patch size \((1,2,2)\), the DiT token grid has size \((1+f)\times h\times w\), where \(h=H/16\) and \(w=W/16\). During inference, the sampler initializes a Gaussian latent \(\bar z_{t_T} \sim \mathcal{N}(0,I)\) with the same shape \(\mathbb{R}^{C_z \times (1+f) \times 2h \times 2w}\).

The text prompt is tokenized to at most \(S=512\) tokens and encoded by an UMT5-XXL text encoder. Let \(c_{\mathrm{raw}} \in \mathbb{R}^{S_c \times D_T}\) denote the non-padding text embedding, where \(S_c \leq S\) and \(D_T=4096\). Wan pads this sequence to length \(S\), then projects it into the DiT hidden dimension using a two-layer MLP, giving \(C=\operatorname{MLP}_{\mathrm{text}}(c_{\mathrm{raw}}) \in \mathbb{R}^{S \times D}\), where \(D=1536\). 

At denoising time \(t\), the latent \(\bar z_t \in \mathbb{R}^{C_z \times (1+f) \times 2h \times 2w}\) is patchified by a 3D convolution with kernel size and stride \((1,2,2)\). This produces a hidden tensor in \(\mathbb{R}^{D \times (1+f) \times h \times w}\), which is flattened in spatio-temporal order into the initial token sequence \(X^{(0)}_t \in \mathbb{R}^{L \times D}\), where \(L=(1+f)hw\). The timestep is embedded by a sinusoidal embedding of dimension \(D_{\mathrm{freq}}=256\), followed by an MLP to obtain \(e_t \in \mathbb{R}^{D}\). A second projection maps this vector to \(e^{\mathrm{blk}}_t \in \mathbb{R}^{6 \times D}\), which provides adaptive shift, scale, and gate parameters for the self-attention and FFN sublayers in every transformer block.

The DiT backbone contains $N_L=30$ transformer layers. Let $X_{t,\ell-1}\in\mathbb{R}^{L\times D}$ be the residual stream entering layer $\ell$ at denoising step $t$. The layer has $N_H=12$ attention heads, each with head dimension $D_H=D/N_H=128$. Its timestep modulation is $m_{t,\ell}=e^{\mathrm{blk}}_t+m_{\ell}\in\mathbb{R}^{6\times D}$, where $m_{\ell}$ is a learned layer-specific modulation parameter. Splitting $m_{t,\ell}$ gives $(a^{\mathrm{sa}}_{t,\ell}, b^{\mathrm{sa}}_{t,\ell}, g^{\mathrm{sa}}_{t,\ell}, a^{\mathrm{ffn}}_{t,\ell}, b^{\mathrm{ffn}}_{t,\ell}, g^{\mathrm{ffn}}_{t,\ell})$, each in $\mathbb{R}^{D}$. The self-attention input is:
\begin{equation}
    \widehat X^{\mathrm{sa}}_{t,\ell}
    =
    \operatorname{LN}_1(X_{t,\ell-1})\odot(1+b^{\mathrm{sa}}_{t,\ell})
    +a^{\mathrm{sa}}_{t,\ell}.
\end{equation}
For head $k$, the query, key, and value tensors are $Q^{\mathrm{sa}}_{t,\ell,k},K^{\mathrm{sa}}_{t,\ell,k},V^{\mathrm{sa}}_{t,\ell,k}\in\mathbb{R}^{L\times D_H}$. Wan applies RMS normalization to queries and keys and applies 3D RoPE according to the token grid $(1+f,h,w)$. The self-attention map is:
\begin{equation}
    A^{\mathrm{sa}}_{t,\ell,k}
    =
    \operatorname{Softmax}
    \left(
        Q^{\mathrm{sa}}_{t,\ell,k}
        (K^{\mathrm{sa}}_{t,\ell,k})^\top
        /\sqrt{D_H}
    \right)
    \in\mathbb{R}^{L\times L}.
\end{equation}
The corresponding per-head attention output before the output projection is:
\begin{equation}
    Z^{\mathrm{sa}}_{t,\ell,k}
    =
    A^{\mathrm{sa}}_{t,\ell,k}V^{\mathrm{sa}}_{t,\ell,k}
    \in\mathbb{R}^{L\times D_H}.
\end{equation}
After concatenating all heads and applying the output projection, the self-attention residual update is:
\begin{equation}
    X'_{t,\ell}
    =
    X_{t,\ell-1}
    +
    g^{\mathrm{sa}}_{t,\ell}\odot
    \operatorname{SA}_{t,\ell}(\widehat X^{\mathrm{sa}}_{t,\ell}).
\end{equation}

Cross-attention injects condition information into the video tokens. Its input is:
\begin{equation}
    \widehat X^{\mathrm{ca}}_{t,\ell}
    =
    \operatorname{LN}_2(X'_{t,\ell})
    \in\mathbb{R}^{L\times D},
\end{equation}
and the text context is $C\in\mathbb{R}^{S\times D}$. For each head $k$, $Q^{\mathrm{ca}}_{t,\ell,k}\in\mathbb{R}^{L\times D_H}$ is computed from video tokens, while $K^{\mathrm{ca}}_{t,\ell,k},V^{\mathrm{ca}}_{t,\ell,k}\in\mathbb{R}^{S\times D_H}$ are computed from text tokens. The cross-attention map is:
\begin{equation}
    A^{\mathrm{ca}}_{t,\ell,k}
    =
    \operatorname{Softmax}
    \left(
        Q^{\mathrm{ca}}_{t,\ell,k}
        (K^{\mathrm{ca}}_{t,\ell,k})^\top
        /\sqrt{D_H}
    \right)
    \in\mathbb{R}^{L\times S}.
\end{equation}
The corresponding per-head attention output before the output projection is:
\begin{equation}
    Z^{\mathrm{ca}}_{t,\ell,k}
    =
    A^{\mathrm{ca}}_{t,\ell,k}V^{\mathrm{ca}}_{t,\ell,k}
    \in\mathbb{R}^{L\times D_H}.
\end{equation}
After concatenating all heads and applying the output projection, the residual stream becomes:
\begin{equation}
    X''_{t,\ell}
    =
    X'_{t,\ell}
    +
    \operatorname{CA}_{t,\ell}(\widehat X^{\mathrm{ca}}_{t,\ell},C).
\end{equation}
Finally, the FFN input is:
\begin{equation}
    \widehat X^{\mathrm{ffn}}_{t,\ell}
    =
    \operatorname{LN}_3(X''_{t,\ell})\odot(1+b^{\mathrm{ffn}}_{t,\ell})
    +a^{\mathrm{ffn}}_{t,\ell},
\end{equation}
and the layer output is:
\begin{equation}
    X_{t,\ell}
    =
    X''_{t,\ell}
    +
    g^{\mathrm{ffn}}_{t,\ell}\odot
    W_{out,\ell}\operatorname{GELU}
    \left(
        W_{in,\ell}\widehat X^{\mathrm{ffn}}_{t,\ell}
    \right),
\end{equation}
where $W_{in,\ell}:\mathbb{R}^{D}\rightarrow\mathbb{R}^{D_{\mathrm{ffn}}}$, $W_{out,\ell}:\mathbb{R}^{D_{\mathrm{ffn}}}\rightarrow\mathbb{R}^{D}$, and $D_{\mathrm{ffn}}=8960$.

After the final layer, $X_{t,N_L}\in\mathbb{R}^{L\times D}$ is passed through the output head. The head uses another timestep-adaptive normalization, then applies a linear projection from $D$ to $64$ channels per token. The projected sequence $Y_t\in\mathbb{R}^{L\times 64}$ is unpatchified back to the VAE-latent grid, yielding the predicted flow-matching velocity $u_\theta(\bar z_t,t,C)\in\mathbb{R}^{C_z\times(1+f)\times 2h\times 2w}$. With classifier-free guidance scale $s_{\mathrm{cfg}}$, Wan evaluates the denoiser twice and uses $u_{\mathrm{CFG}}=u_\theta(\bar z_t,t,C^-)+s_{\mathrm{CFG}}\left(u_\theta(\bar z_t,t,C^+)-u_\theta(\bar z_t,t,C^-)\right)$. A flow-matching ODE solver, by default FlowUniPC, integrates this velocity over $T$ denoising steps to obtain the final clean latent $\bar z_{t_0}$. The Wan-VAE decoder then maps $\bar z_{t_0}$ back to pixel space, producing the generated video $\widehat X\in\mathbb{R}^{(1+F)\times H\times W\times 3}$.

\section{Examples of Cross-Attention Maps}
\label{appn:ca_maps}

\begin{figure*}[htbp]
\centering
\includegraphics[width=\linewidth]{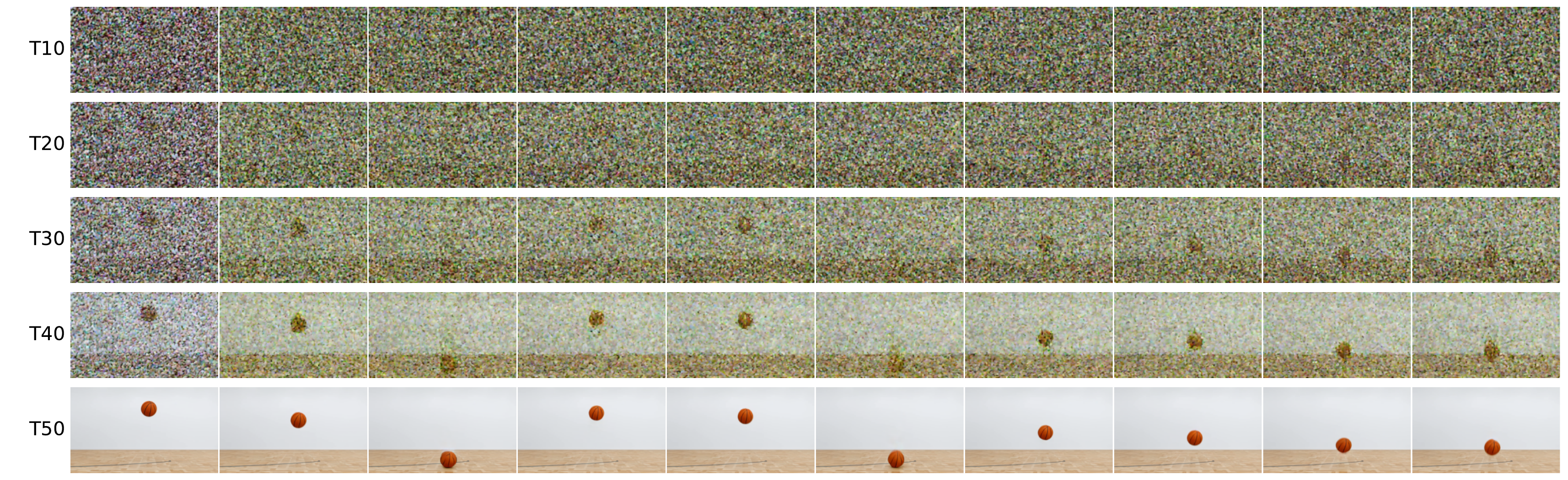}
\caption{The denoising process of Wan2.1-T2V-1.3B. After each denoising step $t$, we directly pass the denoised video latent to the VAE decoder to get the denoised video. Compared to I2V, the initial noise in T2V does not contain any semantic information. Consequently, the video decoded in the early stage of denoising contains a large amount of noise, which is inconvenient for the study of motion planning. In contrast, the pattern of the cross-attention map enables the observation of the process of motion planning at an earlier stage of denoising.}
\label{fig:diffusion_process}
\end{figure*}

\begin{figure}[htbp]
\centering
\begin{tabular}{@{}m{0.04\columnwidth} @{\hspace{0.5pt}} m{0.955\columnwidth} @{}}
\centering\footnotesize T1 &
\includegraphics[width=\linewidth]{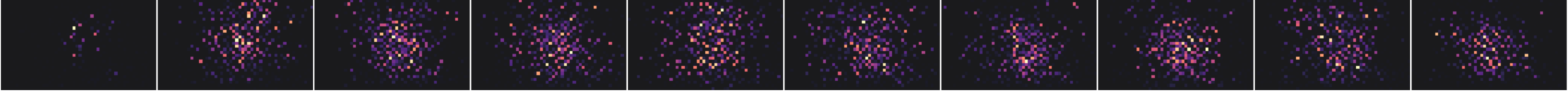} \\
\vspace{1pt}
\centering\footnotesize T3 &
\includegraphics[width=\linewidth]{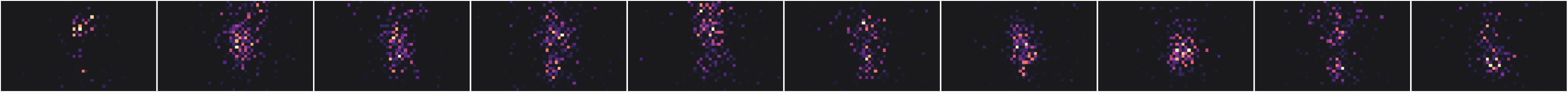} \\
\vspace{1pt}
\centering\footnotesize T5 &
\includegraphics[width=\linewidth]{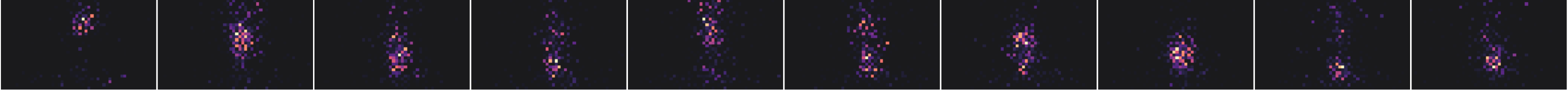} \\
\vspace{1pt}
\centering\footnotesize T7 &
\includegraphics[width=\linewidth]{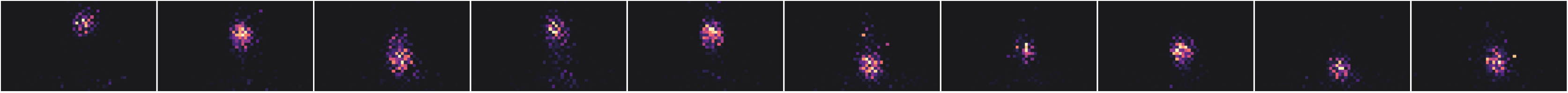} \\
\vspace{1pt}
\centering\footnotesize T9 &
\includegraphics[width=\linewidth]{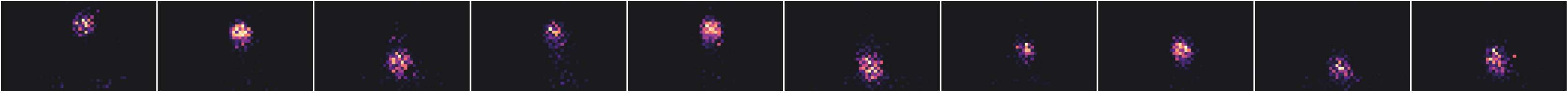} \\

\end{tabular}
\vspace{-5pt}
\caption{The evolution of cross-attention head L15H2 during denoising.}
\label{fig:ca_map_l15h2}
\end{figure}

\begin{figure}[htbp]
\centering
\includegraphics[width=\linewidth]{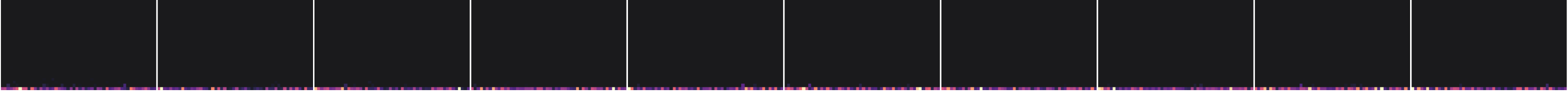}
\caption{The attention pattern of cross-attention head L15H1. During the denoising process, its attention pattern consistently remains like this.}
\label{fig:ca_map_l15h1}
\end{figure}

\begin{figure}[!htbp]
\centering
\begin{tabular}{@{}m{0.04\columnwidth} @{\hspace{0.5pt}} m{0.955\columnwidth} @{}}

\centering\footnotesize (a) &
\includegraphics[width=\linewidth]{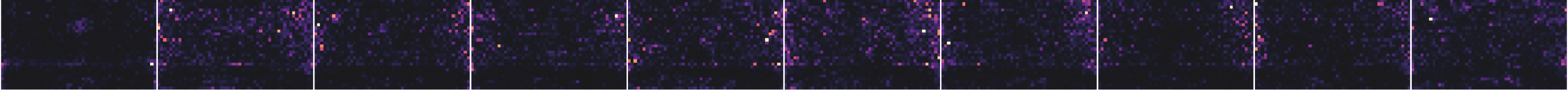} \\
\vspace{1pt}

\centering\footnotesize (b) &
\includegraphics[width=\linewidth]{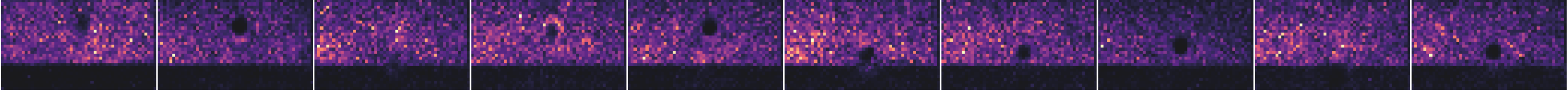} \\
\vspace{1pt}

\centering\footnotesize (c) &
\includegraphics[width=\linewidth]{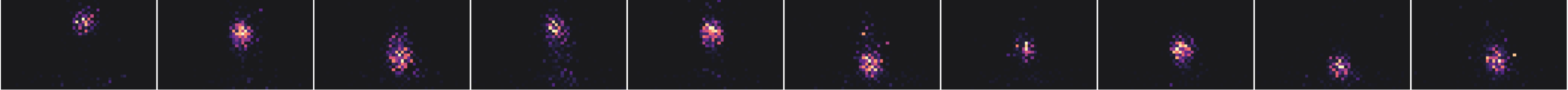} \\
\vspace{1pt}

\centering\footnotesize (d) &
\includegraphics[width=\linewidth]{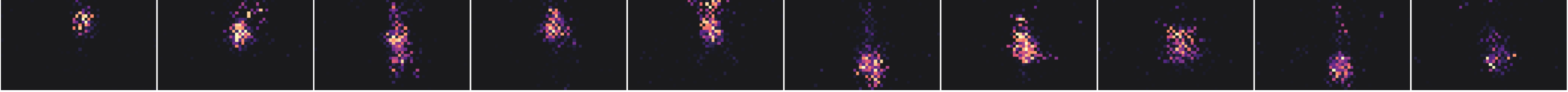} \\

\end{tabular}
\caption{Examples of the cross-attention heads mentioned in section~\ref{fig:ca_head_speed_vs_contri}. Specifically, (a): L6H3, (b): L26H3, (c): L15H2, (d): L9H7. The attention maps are selected at denoising step 7.}
\label{fig:ca_map_examples}
\end{figure}

\clearpage
\section{Processing of Cross / Self-Attention}


\begin{algorithm}[htbp]
\caption{Reference-Trajectory Support Mask Construction}
\label{alg:support_mask}
\begin{algorithmic}
\State \textbf{Input:}
\State \hspace{1em} Reference cross-attention map $\mA^{\mathrm{ref}} \in \mathbb{R}^{f \times h \times w}$ (default: head-averaged map of T50L27);
\State \hspace{1em} Winsorization quantile $q_{\mathrm{win}} \in [0,1]$ (default: $0.995$);
\State \hspace{1em} Despiking quantile $q_{\mathrm{despike}} \in [0,1]$ (default: $0.98$);
\State \hspace{1em} Minimum component area $a_{\min} \in \mathbb{N}$ (default: $2$);
\State \hspace{1em} Trajectory quantile $q_{\mathrm{traj}} \in [0,1]$ (default: $0.95$);
\State \hspace{1em} Radius scale $\alpha > 0$ (default: $1.0$);
\State \hspace{1em} Radius lower bound $r_{\min} > 0$ (default: $1.0$);
\State \hspace{1em} Radius upper-bound ratio $\gamma > 0$ (default: $0.25$)
\State \textbf{Output:}
\State \hspace{1em} Support mask $\mM^{\mathrm{obj}} \in \{0,1\}^{f \times h \times w}$;
\State \hspace{1em} Reference centers $\{c_i\}_{i=1}^f$, where $c_i = (\hat y_i^{\mathrm{ref}}, \hat x_i^{\mathrm{ref}})$;
\State \hspace{1em} Support radii $\{r_i\}_{i=1}^f$
\State
\end{algorithmic}

\begin{algorithmic}[1]
\For{$i = 1$ to $f$}
    \State $\mA_i \gets \mA_i^{\mathrm{ref}} \in \mathbb{R}^{h \times w}$  \textcolor{blue}{\Comment{Step 1: winsorization}}
    \State $\tau_i^{\mathrm{win}} \gets \mathrm{Quantile}(\{\mA_i(y,x)\}_{y,x}, q_{\mathrm{win}})$
    \State $\widetilde{\mA}_i(y,x) \gets \min(\mA_i(y,x), \tau_i^{\mathrm{win}}), \ \forall (y,x)$
    \\
    \State $\tau_i^{\mathrm{despike}} \gets \mathrm{Quantile}(\{\widetilde{\mA}_i(y,x)\}_{y,x}, q_{\mathrm{despike}})$  \textcolor{blue}{\Comment{Step 2: despiking mask}}
    \State $\mM_i^{\mathrm{despike}}(y,x) \gets \mathbf{1}\!\left[\widetilde{\mA}_i(y,x) \ge \tau_i^{\mathrm{despike}}\right]$
    \\
    \State Extract all $8$-connected components of $\mM_i^{\mathrm{despike}}$  \textcolor{blue}{\Comment{Step 3: remove tiny components}}
    \State Remove every component whose area is smaller than $a_{\min}$
    \State Zero out all removed locations in $\widetilde{\mA}_i$
    \\
    \State $(y_i^{\mathrm{peak}}, x_i^{\mathrm{peak}}) \gets \arg\max_{y,x} \widetilde{\mA}_i(y,x)$  \textcolor{blue}{\Comment{Step 4: find the peak-containing component}}
    \State $\tau_i^{\mathrm{traj}} \gets \mathrm{Quantile}(\{\widetilde{\mA}_i(y,x)\}_{y,x}, q_{\mathrm{traj}})$
    \State $\mM_i^{\mathrm{traj}}(y,x) \gets \mathbf{1}\!\left[\widetilde{\mA}_i(y,x) \ge \tau_i^{\mathrm{traj}}\right]$
    \State Force $\mM_i^{\mathrm{traj}}(y_i^{\mathrm{peak}}, x_i^{\mathrm{peak}}) \gets 1$
    \State $\Omega_i^{\mathrm{peak}} \gets$ the $8$-connected component of $\mM_i^{\mathrm{traj}}$ that contains $(y_i^{\mathrm{peak}}, x_i^{\mathrm{peak}})$
    \State $a_i \gets |\Omega_i^{\mathrm{peak}}|$
    \\
    \State
    $
    c_i \gets
    \left(
    \frac{1}{a_i}\sum_{(y,x)\in\Omega_i^{\mathrm{peak}}} y,\ 
    \frac{1}{a_i}\sum_{(y,x)\in\Omega_i^{\mathrm{peak}}} x
    \right)
    $  \textcolor{blue}{\Comment{Step 5: extract the reference center}}
    \\
    \State $r_i^{\mathrm{eq}} \gets \sqrt{a_i / \pi}$  \textcolor{blue}{\Comment{Step 6: determine the support radius}}
    \State $r_i \gets \alpha\, r_i^{\mathrm{eq}}$
    \State $L \gets \min(h, w)$
    \State $r_{\max} \gets \max(r_{\min}, \gamma L)$
    \State $r_i \gets \min(r_{\max}, \max(r_{\min}, r_i))$
    \\
    \For{$y = 1$ to $h$}  \textcolor{blue}{\Comment{Step 7: build the frame-wise circular support mask}}
        \For{$x = 1$ to $w$}
            \State $\mM_i(y,x) \gets \mathbf{1}\!\left[(y-\hat y_i^{\mathrm{ref}})^2 + (x-\hat x_i^{\mathrm{ref}})^2 \le r_i^2\right]$
        \EndFor
    \EndFor
\EndFor
\State \Return $\mM^{\mathrm{obj}} = \{\mM_i\}_{i=1}^f$, $\{c_i\}_{i=1}^f$, $\{r_i\}_{i=1}^f$
\end{algorithmic}
\end{algorithm}

\begin{algorithm}[htbp]
\caption{Candidate Region Extraction}
\label{alg:candidate_region_extraction}
\begin{algorithmic}
\State \textbf{Input:}
\State \hspace{1em} Shared head-mean object-token cross-attention map $\mA^{(s,\ell,\mathrm{mean})} \in \mathbb{R}^{f \times h \times w}$;
\State \hspace{1em} Reference object boxes $\{B_i^{\mathrm{ref}}\}_{i=1}^f$ (computed from Algorithm~\ref{alg:support_mask});
\State \hspace{1em} Base quantile $q_{\mathrm{base}} \in [0,1]$, seed quantiles $\{q_{\mathrm{seed},r}\}_{r=1}^{R}$ (default: $0.85$, $(0.92, 0.95, 0.97)$);
\State \hspace{1em} Smoothing radius $r_{\mathrm{smooth}} \in \mathbb{N}$, winsorization quantile $q_{\mathrm{win}} \in [0,1]$, despiking quantile $q_{\mathrm{despike}} \in [0,1]$ (default: $1$, $0.995$, $0.98$);
\State \hspace{1em} Minimum component area $a_{\min} \in \mathbb{N}$, minimum stable seed levels $n_{\mathrm{stable}} \in \mathbb{N}$, seed merge distance $d_{\mathrm{merge}} > 0$, merge overlap threshold $\tau_{\mathrm{merge}} \in [0,1]$ (default: $4$, $2$, $2.0$, $0.75$)
\State \textbf{Output:}
\State \hspace{1em} Candidate label map $\mL^{(s,\ell)} \in \mathbb{N}_0^{f \times h \times w}$, candidate regions $\{\Omega_{i,k}\}_{k=1}^{K_i}$ for each frame $i$
\State
\end{algorithmic}

\begin{algorithmic}[1]
\State Set $K_{\max} \gets 5$, $\eta \gets 0.80$, $\rho_w \gets 0.40$, $\rho_h \gets 0.95$
\For{$i = 1$ to $f$}
    \State $\mA_i \gets \mA_i^{(s,\ell,\mathrm{mean})}$  \textcolor{blue}{\Comment{Step 1: preprocessing and smoothing}}
    \State Apply winsorization to $\mA_i$ with quantile $q_{\mathrm{win}}$
    \State Apply despiking to $\mA_i$ with quantile $q_{\mathrm{despike}}$ and minimum component area $a_{\min}$
    \State Obtain $\bar{\mA}_i$ by $(2r_{\mathrm{smooth}}+1)\times(2r_{\mathrm{smooth}}+1)$ uniform averaging with zero padding
    
    \\
    \State $q_{\mathrm{bg}} \gets \max(\ 0.50, \min(q_{\mathrm{base}} - 0.10, 0.92))$  \textcolor{blue}{\Comment{Step 2: background suppression}}
    \State $t_i^{\mathrm{bg}} \gets \mathrm{Quantile}(\{\bar{\mA}_i(y,x)\}_{y,x}, q_{\mathrm{bg}})$
    \State $\mB'_i(y,x) \gets \max(\bar{\mA}_i(y,x)-t_i^{\mathrm{bg}}, 0)$
    \State $\mB_i(y,x) \gets \left(\mB'_i(y,x)/\max_{u,v}\mB'_i(u,v)\right)^2,\ \forall (y,x)$
    
    \\
    \State $t_i^{\mathrm{sup}} \gets \mathrm{Quantile}(\{\mB_i(y,x): \mB_i(y,x) > 0\}, q_{\mathrm{base}})$    \textcolor{blue}{\Comment{Step 3: support set}}
    \State $\mathcal{S}_i \gets \{(y,x): \mB_i(y,x) \ge t_i^{\mathrm{sup}}\}$
    
    \\
    \State $\mathcal{P}_i \gets \varnothing$  \textcolor{blue}{\Comment{Step 4: multi-level peak proposals}}
    \For{each seed quantile $q_{\mathrm{seed},r}$}
        \State $\hat q_{\mathrm{seed},r} \gets \min(0.995,\ \max(q_{\mathrm{base}}, q_{\mathrm{seed},r}))$
        \State $t_{i,r}^{\mathrm{seed}} \gets \mathrm{Quantile}(\{\mB_i(y,x): \mB_i(y,x) > 0\}, \hat q_{\mathrm{seed},r})$
        \State Extract 8-neighborhood local maxima of $\mB_i$ above $t_{i,r}^{\mathrm{seed}}$ inside $\mathcal{S}_i$
        \State Add one peak proposal from each connected local-maximum component to $\mathcal{P}_i$
    \EndFor

    \\
    \State Greedily merge proposals in $\mathcal{P}_i$ within distance $d_{\mathrm{merge}}$  \textcolor{blue}{\Comment{Step 5: stable seeds}}
    \State Keep seeds appearing on at least $n_{\mathrm{stable}}$ levels and retain at most $K_{\max}$ seeds
    \State Run seeded weighted $k$-means on support points in $\mathcal{S}_i$ with weights $\mB_i$

    \\
    \For{each cluster $\mathcal{C}_{i,k}$}  \textcolor{blue}{\Comment{Step 6: clustering and core trimming}}
        \State Keep the smallest-radius subset containing at least $\eta$ of the cluster mass
        \State $\Omega_{i,k} \gets$ the largest connected component inside that compact core
    \EndFor

    \\
    \State Discard region $\Omega_{i,k}$ such that $|\Omega_{i,k}| \!<\! a_{\min}$, $\frac{\sum_{(y,x)\in\Omega_{i,k}} \mB_i(y,x)}{|\Omega_{i,k}|} \!<\!
 \frac{\sum_{(y,x)\in\mathcal{C}_{i,k}} \mB_i(y,x)}{|\mathcal{C}_{i,k}|}$, $\mathrm{bbox\_width}(\Omega_{i,k})/w > \rho_w$, or $\mathrm{bbox\_height}(\Omega_{i,k})/h > \rho_h$    \textcolor{blue}{\Comment{Step 7: strong pruning}}
    
    \\
    \State $\mathcal{E}_i \gets \{k : |\Omega_{i,k} \cap B_i^{\mathrm{ref}}| / |\Omega_{i,k}| \ge \tau_{\mathrm{merge}}\}$  \textcolor{blue}{\Comment{Step 8: reference-box merge}} 
    \If{$|\mathcal{E}_i| \ge 2$}
        \State Merge all $\{\Omega_{i,k}\}_{k \in \mathcal{E}_i}$ into one region
    \EndIf
    \State Write the surviving regions into the frame label map $\mL_i$
\EndFor
\State \Return $\mL^{(s,\ell)} = \{\mL_i\}_{i=1}^{f}$ and $\{\Omega_{i,k}\}_{k=1}^{K_i}$
\end{algorithmic}
\end{algorithm}

\newpage
\section{Motion Planning Heads in Cross-Attention}
\subsection{Attribution Patching for Motion Planning}
We view a model $M$ as a computational graph $\mathcal{G} = \{ \mathcal{V}, \mathcal{E} \}$, where $\mathcal{V}$ and $\mathcal{E}$ denote the sets of nodes and edges, respectively. A node $n$ represents a model component, such as a neuron, an attention head, or an MLP layer, depending on the granularity of the analysis. An edge $e: n_1 \rightarrow n_2$ describes the information flow, specifically the path from the output of an upstream node $n_1$ to the input of a downstream node $n_2$.
We aim to determine whether a node significantly influences the final output of the model. This requires measuring the contribution of each node in the computational graph to the final output. This can be achieved through causal intervention \citep{causal_inf}, which involves perturbing a node and quantifying the resulting change in the final output to calculate its contribution, or \textit{indirect effect}, denoted by $c(\cdot)$.
The metric $\mathcal{L}_m$ to quantify the final output often relies on the downstream task. For example, \citet{ioi} use the logit difference as the metric for the indirect object identification task in GPT-2 Small. 

In this work, our aim is to find the important nodes for motion planning in video generation. Unlike language models, the output of a flow matching model is the velocity during denoising, which requires considering multiple denoising timesteps rather than a single forward pass. Furthermore, since video generation lacks tokens with explicit meanings like those in language models, it is difficult to isolate a distinct component from the velocity to measure an object's motion state (such as motion direction or magnitude). 
Therefore, as shown in Equation~\ref{eq: head_contri}, we mainly consider three aspects when designing $\mathcal{L}_m$: 
\begin{itemize}
    \item We focus the contribution computation on the first 5 steps of the 50-step denoising process and perform causal intervention on each step individually, because motion planning primarily occurs during these initial 5 steps;
    \item We focus on the object regions within the video latent rather than the background regions, allowing $\mathcal{L}_m$ to concentrate more on object motion;
    \item We measure the difference in the model's final output before and after causal intervention using the dot product between two velocity vectors within the object region: the velocity $u_t^{\mathrm{cond}}$ predicted by the conditional branch after intervention, and $\Delta u_t^{\mathrm{clean}}$, which is the difference between the velocities predicted by the conditional and unconditional branches before intervention. Intuitively, the dot product of two vectors positively correlates with their similarity. Besides, we adopt $\Delta u_t^{\mathrm{clean}}$ as the reference velocity because this difference eliminates general information in the velocity, retaining more semantic information related to the object. This prevents $\mathcal{L}_m$ from being perturbed by static, constant information present in the unconditional velocity.
\end{itemize}

Next, we introduce the specific implementation of causal intervention. The most fundamental and straightforward algorithm is \textbf{\textit{activation patching}}. We first consider the scenario in language models. 
Given an input $x^{\text{clean}}$ (e.g., a prompt), we aim to perturb a specific node $n$. The recommended approach is to construct a corrupted prompt $x^{\text{noise}}$ with the same format as $x^{\text{clean}}$ but with different or opposite semantics. For instance, if $x^{\text{clean}}$ is ``The \textit{kids} on the beach" and the task is verb prediction, $x^{\text{noise}}$ could be ``The \textit{kid} on the beach". We first input $x^{\text{noise}}$ into the model to obtain the corrupted activation $n(x^{\text{noise}})$ at node $n$. Then, we input $x^{\text{clean}}$ and, upon reaching node $n$, apply the following intervention to compute the score for node $n$:
\label{appn:head_contri}
\begin{equation}
    c(n) = \mathcal{L}_m \big(M\big(x^{\text{clean}} | do(n \leftarrow n(x^{\text{noise}}))\big)\big) - \mathcal{L}_m(M(x^{\text{clean}})) 
\end{equation}
Here, we use do-calculus notation \citep{causal_graph} to represent the intervention process. We perform this intervention for each node $n \in \mathcal{V}$. Generally, we take the absolute value of the scores. Another intervention method is to set the corrupted value of node $n$ to a noise value. In this work, when intervening on an attention head, we use the average of its outputs across all positions in the video latent as the corrupted value, which is called \textit{mean ablation}.

Given $|\mathcal{V}|$ nodes in the computational graph, the time complexity of standard activation patching is $\mathcal{O}(|\mathcal{V}|)$. To reduce this complexity, \citet{atp} proposed \textbf{\textit{attribution patching}}, which uses a first-order Taylor expansion as a linear approximation of $c(n)$. Specifically, treating the corrupted value $n(x^{\text{noise}})$ as the independent variable and $c(n)$ as the dependent variable, we take the first-order Taylor expansion of $c(n)$ at $n(x^{\text{noise}}) = n(x^{\text{clean}})$:
\begin{align}
    c(n) &\nonumber \approx \mathcal{L}_m(M(x^{\text{clean}})) + [n(x^{\text{noise}}) - n(x^{\text{clean}})]^{\top} \cdot \nabla_n \mathcal{L}_m(M(x^{\text{clean}})) |_{n=n(x^{\text{clean}})} - \mathcal{L}_m(M(x^{\text{clean}})) \\
    &= [n(x^{\text{noise}}) - n(x^{\text{clean}})]^{\top} \cdot \nabla_n \mathcal{L}_m(M(x^{\text{clean}})) |_{n=n(x^{\text{clean}})}
\label{eq:attribution_patching}
\end{align}
This allows us to compute all node scores simultaneously using only one forward and backward pass on $x^{\text{clean}}$ when combined with mean ablation, reducing the time complexity to $\mathcal{O}(1)$.

\subsection{Zero Ablation of Cross-Attention Heads}
\label{appn:zero_ablation}

\paragraph{Per-head write}is the attention output from an attention head to the residual stream. For an attention layer $\ell$ with $N_H$ heads, let $Z_{t,\ell,k}\in\mathbb{R}^{L\times D_H}$ denote the attention output of head $k$ at denoising step $t$ after attention aggregation, i.e., $Z_{t,\ell,k}=A_{t,\ell,k}V_{t,\ell,k}$, where $A_{t,\ell,k}$ and $V_{t,\ell,k}$ are the attention weights and value in self/cross attention, respectively. The attention output projection $W_{O,\ell}$ maps the concatenated head outputs back to the residual-stream dimension $D$. We partition its weight by heads as $W_{O,\ell}=[W_{O,\ell,1},W_{O,\ell,2},\dots,W_{O,\ell,N_H}]$, where $W_{O,\ell,k}\in\mathbb{R}^{D_H\times D}$. The per-head write of head $k$ is then defined as:
\begin{equation}
    U_{t,\ell,k}
    =
    Z_{t,\ell,k}W_{O,\ell,k}
    \in\mathbb{R}^{L\times D}.
\end{equation}
If the attention sublayer applies a residual gate $g_{t,\ell}\in\mathbb{R}^{D}$ before the residual addition ($\sum_{k=1}^{N_H}U_{t,\ell,k}$), we absorb this gate into the definition and write:
\begin{equation}
    U_{t,\ell,k}
    =
    g_{t,\ell}\odot Z_{t,\ell,k}W_{O,\ell,k} \in\mathbb{R}^{L\times D},
\end{equation}
where $g_{t,\ell}$ is broadcast over the token dimension. Thus, the attention residual update can be decomposed as:
\begin{equation}
    X_{t,\ell}
    =
    X_{t,\ell-1}
    +
    \sum_{k=1}^{N_H}U_{t,\ell,k},
\end{equation}
Therefore, $U_{t,\ell,k}$ represents the actual direction and magnitude written by head $k$ into the residual stream, rather than the attention weights $A_{t,\ell,k}$ or the pre-projection head output $Z_{t,\ell,k}$.

\paragraph{Zero ablation}
Consider a subset of cross-attention heads, denoted as $\mathcal{H} \subseteq \{ \text{L}_x\text{H}_y \mid 0 \le x < N_L, 0 \le y < N_H \}$. In denoising step $t$, we zero out the per-head write of all heads in $\mathcal{H}$ and observe the changes in the generated video compared to the state before ablation. This approach allows us to directly evaluate the influence of the attention heads in $\mathcal{H}$ on the model's generation process. In practice, when using zero ablation, we apply the aforementioned intervention across all denoising steps by default.

\subsection{Experiment details}
\label{appn:ca_head_contri_details}
\begin{figure}[htbp]
\centering
\begin{tabular}{@{}m{0.04\columnwidth} @{\hspace{0.5pt}} m{0.955\columnwidth} @{}}

\centering\footnotesize &
\includegraphics[width=\linewidth]{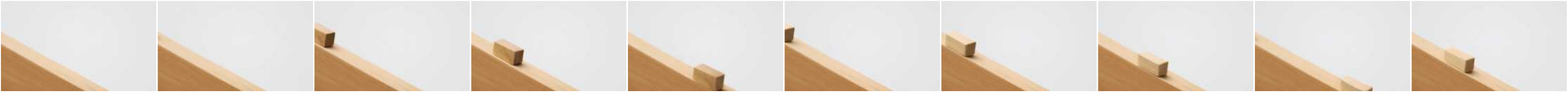} \\
\vspace{1pt}

\centering\footnotesize (a) &
\includegraphics[width=\linewidth]{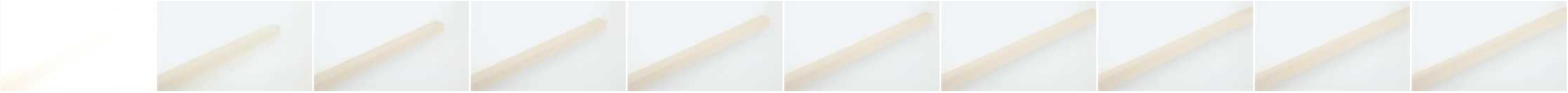} \\
\vspace{1pt}

\centering\footnotesize (b) &
\includegraphics[width=\linewidth]{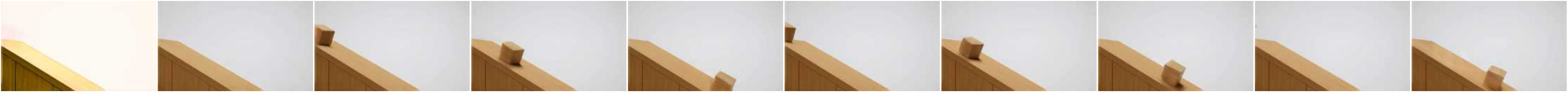} \\
\vspace{1pt}

\centering\footnotesize (c) &
\includegraphics[width=\linewidth]{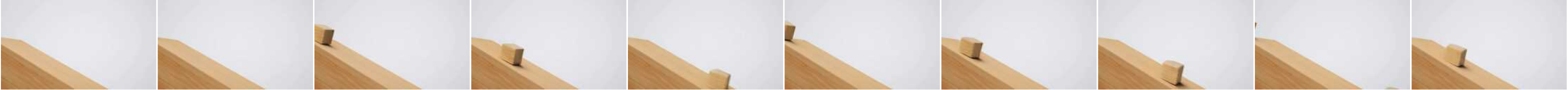} \\
\vspace{1pt}

\centering\footnotesize (d) &
\includegraphics[width=\linewidth]{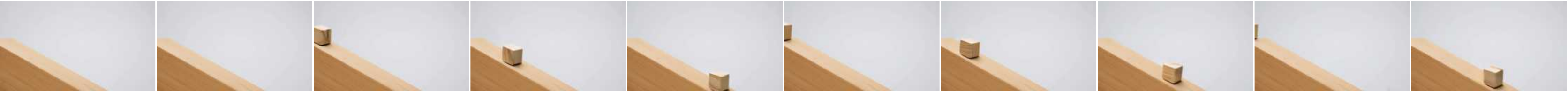} \\
\vspace{1pt}

\centering\footnotesize (e) &
\includegraphics[width=\linewidth]{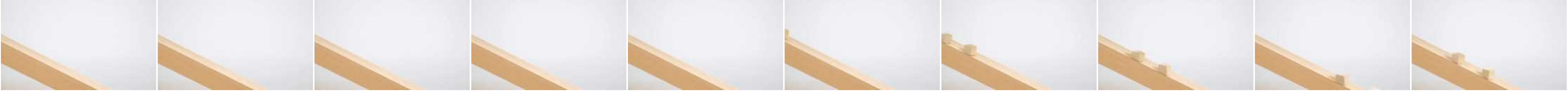} \\

\end{tabular}
\caption{Row1: The video generated by Wan2.1-T2V-1.3B with the prompt ``Against a pure white background, a wooden cube block at the top of a smooth slope slides straight down the slope with steadily and uniformly increasing speed.'' with a ramdom seed of 2. Row2-6: The generated videos corresponding to the various zero ablations of cross-attention heads in Section \ref{sec:ca_motion_heads}.}
\label{fig:ablate_results_cube}
\vspace{-15pt}
\end{figure}

\begin{figure}[htbp]
\centering
\begin{tabular}{@{}m{0.04\columnwidth} @{\hspace{0.5pt}} m{0.955\columnwidth} @{}}

\centering\footnotesize &
\includegraphics[width=\linewidth]{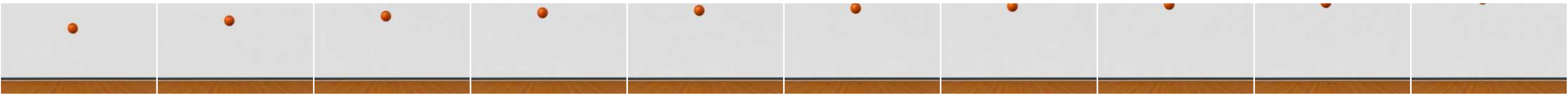} \\
\vspace{1pt}

\centering\footnotesize (a) &
\includegraphics[width=\linewidth]{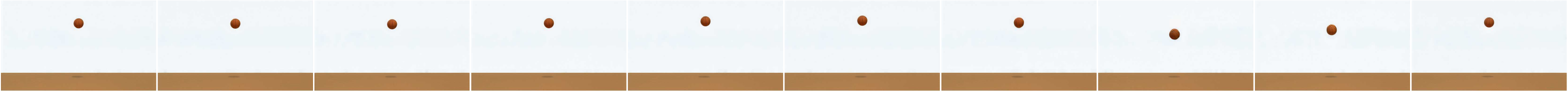} \\
\vspace{1pt}

\centering\footnotesize (b) &
\includegraphics[width=\linewidth]{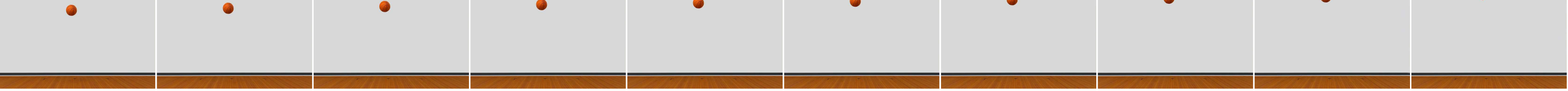} \\
\vspace{1pt}

\centering\footnotesize (c) &
\includegraphics[width=\linewidth]{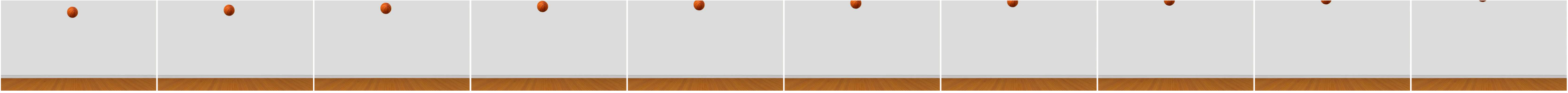} \\
\vspace{1pt}

\centering\footnotesize (d) &
\includegraphics[width=\linewidth]{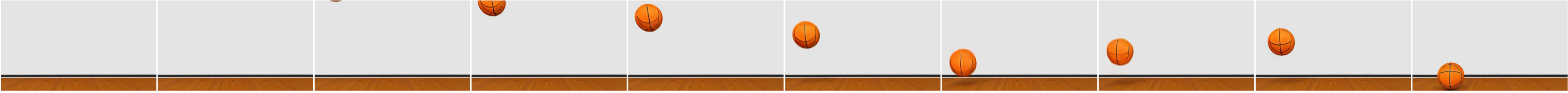} \\
\vspace{1pt}

\centering\footnotesize (e) &
\includegraphics[width=\linewidth]{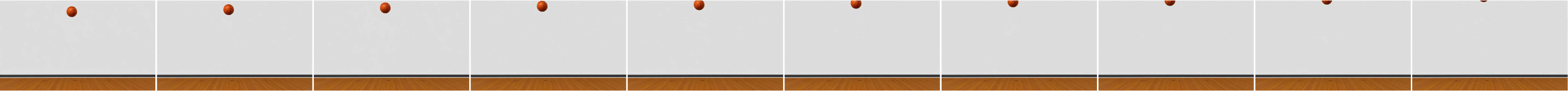} \\

\end{tabular}
\caption{Row1: The video generated by Wan2.1-T2V-1.3B with the prompt ``Against a pure white background, a basketball falls vertically from mid-air onto a wooden floor and bounces up several times.'' with a ramdom seed of 8. Row2-6: The generated videos corresponding to the various zero ablations of cross-attention heads in Section \ref{sec:ca_motion_heads}.}
\label{fig:ablate_results_seed8}
\vspace{-15pt}
\end{figure}

In Section~\ref{sec:ca_motion_heads}, our classification of cross-attention heads is based on a fixed case and random seed. However, we find that the previously drawn conclusions also hold true for different cases and seeds. Here, we provide some additional visualization results, as shown in Figure~\ref{fig:ablate_results_cube} and \ref{fig:ablate_results_seed8}.

\clearpage
\section{RoPE Details}
\label{appn:rope}
We briefly introduce the basic principles of RoPE \citep{rope}. The goal of RoPE is to introduce relative positional information into the dot product of the query and key in the attention module, thereby enabling the attention module to capture causal logic information between tokens (in language models) or spatial positional relationships (in vision models). First, we consider 1D-RoPE in language models. Let the model dimension be $D$. RoPE first performs pairwise grouping across the entire model dimension, injecting positional information with two dimensions as a unit. Therefore, we consider the simplest case where $D=2$, namely query $q = [q_0, q_1] \in \mathbb{R}^{2}$ and key $k = [k_0, k_1] \in \mathbb{R}^{2}$, with their respective position ids being $m$ and $n$. In complex form, $q$ and $k$ can be written as $q_0 + iq_1$ and $k_0 + ik_1$, respectively. RoPE first introduces positional information in the query and key, respectively, taking the query as an example:
\begin{equation}
\label{eq:rope_trans complex}
\begin{split}
f(q, m) &= qe^{im\theta} \\
&=(q_0 + iq_1)(cos(m\theta) + isin(m\theta)) \\
&=[q_0cos(m\theta) - q_1sin(m\theta)] + i[q_0sin(m\theta) + q_1cos(m\theta)]
\end{split}
\end{equation}
If represented by matrix multiplication, this transformation can also be written as:
\begin{equation}
\label{eq: rope_trans_vec}
\begin{split}
f(q, m)^{\top} &= \begin{pmatrix}
cos(m\theta) & -sin(m\theta) \\
sin(m\theta) & cos(m\theta)
\end{pmatrix} \begin{pmatrix}
q_0 \\
q_1 \\
\end{pmatrix} \\
&=\begin{pmatrix}
q_0cos(m\theta) - q_1sin(m\theta) \\
q_0sin(m\theta) + q_1cos(m\theta)
\end{pmatrix}
\end{split}
\end{equation}
The transformation RoPE applies to the query $q$ is a multiplication by a rotation matrix that carries its positional information, which is the origin of the name of RoPE. The frequency is typically $\theta_i=b^{-2i/D} (i = 0, ..., D/2-1)$, where $b$ is called the base frequency, and $i$ is the dimension group index (in our example, $i$ only takes the value of 0 because $D=2$).

Then, the dot product between $q$ and $k$ is:
\begin{equation}
\label{eq: rope_dot_product_matrix}
\begin{split}
&<f(q, m), f(k, n)> \\
=&f(q, m)f(k, n)^{\top} \\
=&\begin{pmatrix}
q_0 & q_1
\end{pmatrix} \begin{pmatrix}
cos(m\theta) & sin(m\theta) \\
-sin(m\theta) & cos(m\theta)
\end{pmatrix} \begin{pmatrix}
cos(n\theta) & -sin(n\theta) \\
sin(n\theta) & cos(n\theta)
\end{pmatrix} \begin{pmatrix}
k_0 \\
k_1
\end{pmatrix} \\
=&\begin{pmatrix}
q_0 & q_1
\end{pmatrix} \begin{pmatrix}
cos(m-n)\theta & sin(m-n)\theta \\
-sin(m-n)\theta & cos(m-n)\theta
\end{pmatrix} \begin{pmatrix}
k_0 \\
k_1
\end{pmatrix} \\
=&(q_0k_0 + q_1k_1)cos(m-n)\theta + (q_0k_1 - q_1k_0)sin(m-n)\theta
\end{split}
\end{equation}

Similarly, we can express this in the form of complex multiplication:
\begin{equation}
\label{eq: rope_dot_product_complex}
\begin{split}
&<f(q, m), f(k, n)> \\
=&f(q, m)f(k, n)^{\top} \\
=&\operatorname{Re}[qe^{im\theta} \cdot (ke^{in\theta})^{*}] \\
=&\operatorname{Re}[qk^{*}e^{i(m-n)\theta}]
\end{split}
\end{equation}
As can be seen, the dot product between the query and the key is the real part of the multiplication between the transformed query and the conjugate of the transformed key.
\begin{figure*}[!t]
\centering
\begin{subfigure}{0.46\linewidth}
    \includegraphics[width=\linewidth]{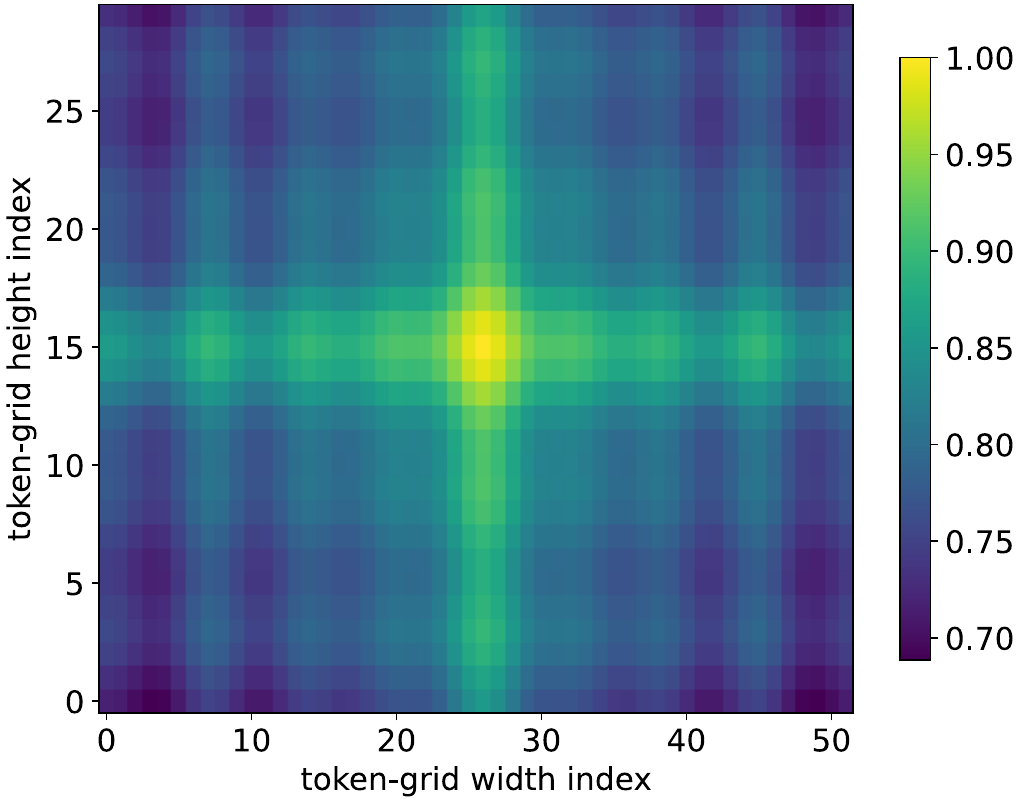}
    \caption{The spatial center heatmap of RoPE.}
    \label{fig:rope_spatial_decay}
\end{subfigure}
\hspace{10pt}
\begin{subfigure}{0.46\linewidth}
    \includegraphics[width=\linewidth]{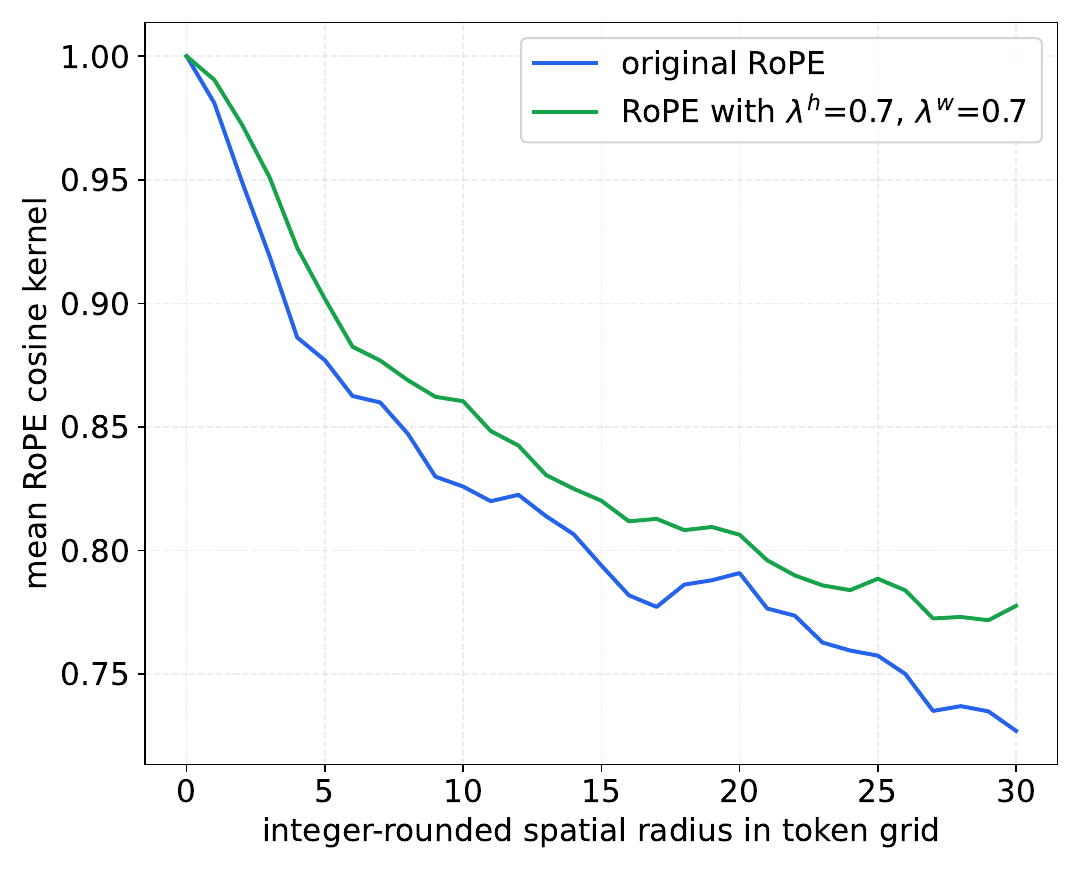}
    \caption{The spatial RoPE decay curve.}
    \label{fig:rope_spatial_decay_curve}
\end{subfigure}
\caption{Visualizations of the spatial decay in 3D-RoPE in Wan2.1-T2V-1.3B. }
\label{fig:rope_spatial_decay_map}
\end{figure*}

Similar to the basic form of 1D RoPE, 3D-RoPE introduces three-dimensional positional information to separately represent information along the frame, height and width directions for each patch in an image. Specifically, 3D-RoPE divides the model dimension into three equal halves, corresponding to frame, height and width, respectively. Suppose we have a six-dimensional query $q \in \mathbb{R}^6$ at coordinates $p=(p^f, p^h, p^w)$. Its vector form is $[q^f, q^h, q^w]=[ q_0, q_1, q_2, q_3, q_4, q_5 ]$, where $q^f=[ q_0, q_1 ]$, $q^h=[q_2, q_3]$, $q^w=[q_4, q_5]$. Its complex form is $[q^f, q^h, q^w]=[q_0 + iq_1, q_2 + iq_3, q_4 + iq_5]$, where $q^f=q_0 + iq_1$, $q^h=q_2 + iq_3$, $q^w=q_4 + iq_5$. The transformation $f(q, p)$ that 3D-RoPE applies to the query is:
\begin{equation}
\label{eq: 3d_rope_dot_product_matrix}
    f(q, p)^{\top} \!=\! \begin{pmatrix}
    cos(p^f\theta) & -sin(p^f\theta) & 0 & 0 & 0 & 0 \\
    sin(p^f\theta) & cos(p^f\theta) & 0 & 0 & 0 & 0 \\
    0 & 0 & cos(p^h\theta) & -sin(p^h\theta) & 0 & 0 \\
    0 & 0 & sin(p^h\theta) & cos(p^h\theta) & 0 & 0 \\
    0 & 0 & 0 & 0 & cos(p^w\theta) & -sin(p^w\theta) \\
    0 & 0 & 0 & 0 & sin(p^w\theta) & cos(p^w\theta) \\
    \end{pmatrix} \begin{pmatrix}
    q_0 \\
    q_1 \\
    q_2 \\
    q_3 \\
    q_4 \\
    q_5
    \end{pmatrix}
\end{equation}
When written in the complex form, the dot product $<f(q, p_q), f(k, p_k)>$ between $q$ and $k$ is:
\begin{equation}
\label{eq: 2d rope dot product complex}
\begin{split}
     &\operatorname{Re}\big[\big(q^fe^{ip_q^f\theta}, q^he^{ip_q^h\theta}, q^we^{ip_q^w\theta}\big)\cdot\big( (k^fe^{ip_k^f\theta})^{*}, (k^he^{ip_k^h\theta})^{*}, (k^we^{ip_k^w\theta})^{*} \big)\big] \\
    =&\operatorname{Re}\big[ q^f{k^f}^{*}e^{i(p_q^f - p_k^f)\theta} + q^h{k^h}^{*}e^{i(p_q^h - p_k^h)\theta} + q^w{k^w}^{*}e^{i(p_q^w - p_k^w)\theta} \big] \\
    =&\operatorname{Re}\big[ q^f{k^f}^{*}e^{i\Delta p^f\theta} + q^h{k^h}^{*}e^{i\Delta p^h\theta}, q^w{k^w}^{*}e^{i\Delta p^w\theta} \big] \\
    =&\operatorname{Re}\big[\sum\nolimits_{a \in \{f,h,w\}} q^{a^{}} {k^{a}}^{*} e^{i \Delta p^a \theta}\big]
\end{split}
\end{equation}
Next, we visualize the effect of 3D-RoPE on the dot product between two tokens in the same frame. Let $m_f,m_h,m_w$ denote the number of dimension pairs for frame, height and width, respectively. 
For axis \(a\in\{f,h,w\}\), the \(i\)-th pair uses $\theta_{a,i}=10000^{-2i/d_a}, i=0,1,\ldots,m_a-1$,
where \(d_a=2m_a\) is the dimension of axis \(a\).
The original model uses the same base frequency for the three axes, i.e., \(\theta_{f,i}=\theta_{h,i}=\theta_{w,i}\) in the sense that no axis-specific base frequency is introduced.
For a dimension pair $i$ on axis \(a\), the RoPE dot-product formula gives $\operatorname{Re}\left[q_i^a {k_i^a}^{*} e^{i\Delta p^a\theta_{a,i}}\right]$.
Let $q_i^a {k_i^a}^{*}=A_i^a+iB_i^a$.
Then $\operatorname{Re}\left[(A_i^a+iB_i^a)e^{i\Delta p^a\theta_{a,i}}\right]=A_i^a\cos(\Delta p^a\theta_{a,i})-B_i^a\sin(\Delta p^a\theta_{a,i})$.
In the visualization, we do not model the content-dependent coefficients \(A_i^a\) and \(B_i^a\). 
Instead, we keep the cosine factor by considering the self-correlation case. When the two pre-RoPE vectors are the same, i.e., \(q_i^a=k_i^a=u_i^a\), then the dot product is $\|u_i^a\|^2\cos(\Delta_a\theta_{a,i})$.
Therefore, we define
\begin{equation}
    K(\Delta p^f,\Delta p^h,\Delta p^w) = \frac{1}{m_f+m_h+m_w} \sum\nolimits_{a\in \{f,h,w\}} \sum\nolimits_{i=0}^{m_a-1}\cos(\Delta p^a\theta_{a,i})
\end{equation}
This quantity only keeps the RoPE-dependent cosine factors and averages them over all dimension pairs in one head.
For the \textit{spatial center heatmap} (Figure~\ref{fig:rope_spatial_decay_map}), we fix the two tokens to be in the same frame, so \(\Delta_f=0\). Let the center spatial token be $(h_\ast,w_\ast)=\left( \left\lfloor\frac{h}{2}\right\rfloor,\left\lfloor\frac{w}{2}\right\rfloor \right).$
For each spatial position $p_s=(p_s^h,p_s^w)$, the plotted value is: $K_{\mathrm{heatmap}}(p_s)=K(0,p_s^h-p_\ast^h,p_s^w-p_\ast^w).$

The \textit{RoPE spatial decay curve} is derived from the heatmap. For each spatial position \(p_s=(p_s^h,p_s^w)\), define $r(p_s)\!=\!\operatorname{round}\big(\sqrt{(p_s^h-p_\ast^h)^2+(p_s^w-p_\ast^w)^2}\big)$.
For a radius $\rho$, let $\mathcal{S}_{\rho}\!=\!\{p_s: r(p_s)=\rho\}$.
The curve is $K_{\mathrm{curve}}(\rho)=\frac{1}{|\mathcal{S}_{\rho}|}\sum_{p_s\in\mathcal{S}_{\rho}}K_{\mathrm{heatmap}}(p_s)$.
Thus, the curve is the average of \(K_{\mathrm{heatmap}}(p_s)\) over all spatial positions whose rounded distance to the center position is \(\rho\), as shown in Figure~\ref{fig:rope_spatial_decay_curve}.

\clearpage
\section{More Results on Self-attention Mechanisms}

\vspace{-5pt}
\begin{figure}[htbp]
\centering
\includegraphics[width=1.0\columnwidth]{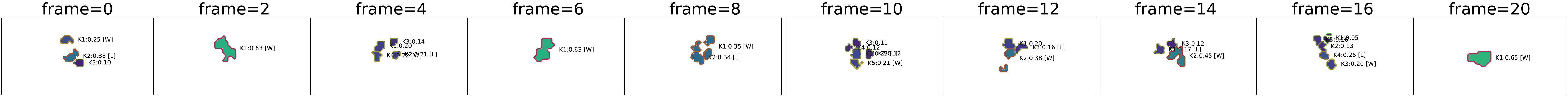}
\includegraphics[width=1.0\columnwidth]{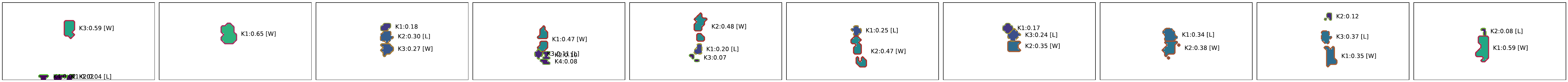}
\includegraphics[width=1.0\columnwidth]{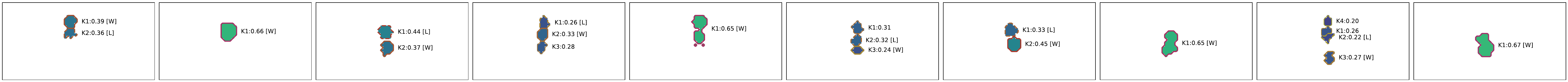}
\includegraphics[width=1.0\columnwidth]{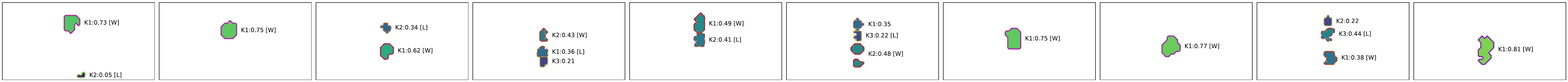}
\includegraphics[width=1.0\columnwidth]{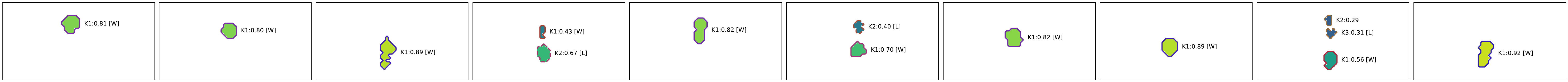}
\includegraphics[width=1.0\columnwidth]{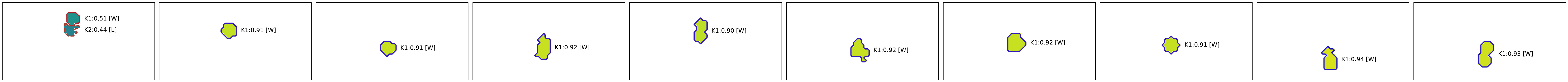}
\includegraphics[width=1.0\columnwidth]{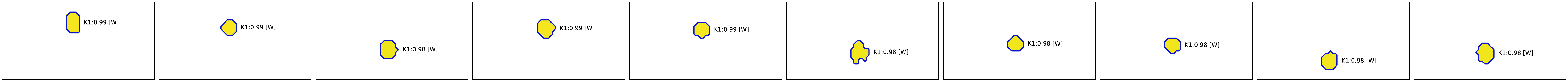}
\caption{The evolution of candidate regions during denoising (seed26, layer18, from T1 to T7). Better viewed when zoomed in.}
\label{fig:candidate_evolution_seed26}
\vspace{-5pt}
\end{figure}
\begin{figure*}[htbp]
\centering
\begin{subfigure}{0.46\linewidth}
    \includegraphics[width=\linewidth]{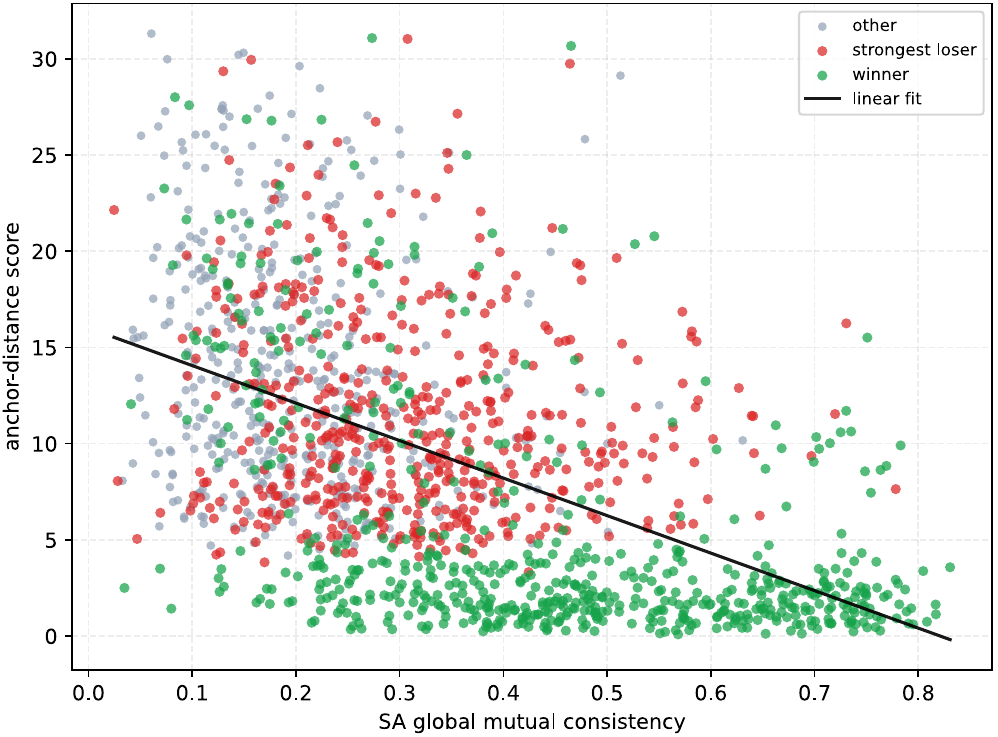}
    \caption{Denoising step 3}
    \label{fig:mutual_consistency_t3}
\end{subfigure}
\hspace{10pt}
\begin{subfigure}{0.46\linewidth}
    \includegraphics[width=\linewidth]{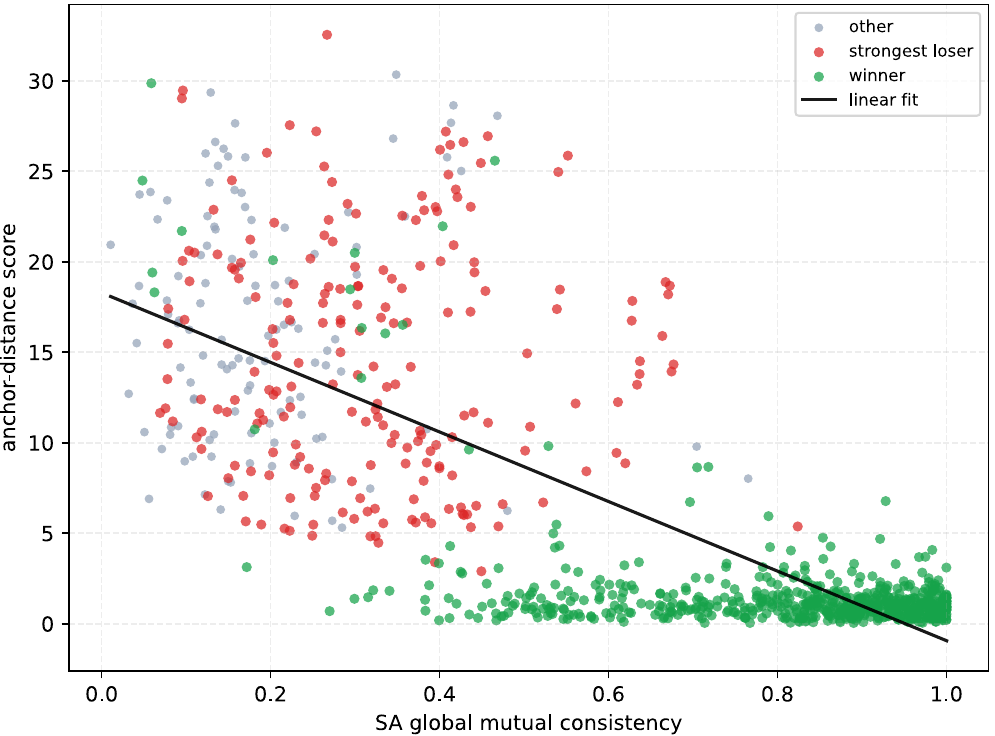}
    \caption{Denoising step 10}
    \label{fig:mutual_consistency_t10}
\end{subfigure}
\vspace{-5pt}
\caption{The relationship between mutual consistency and anchor distance.}
\label{fig:mutual_consistency}
\vspace{-15pt}
\end{figure*}

In Figure~\ref{fig:mutual_consistency}, we visualize the scatter plots of mutual consistency (x-axis) versus anchor distance (y-axis) for candidate regions at each denoising step, both of which are defined in detail in Section~\ref{sec:sa_interp}. In the scatter plot for each denoising step, each point represents a candidate region within a video latent frame at a specific layer. Here, green points represent the winners (i.e., regions belonging to the final trajectory), red points represent the strongest losers (i.e., regions second closest to the object position in the final trajectory for that frame), and gray points represent the remaining regions.

As can be seen, at denoising step 3, the mutual consistency of candidate regions other than the winners is mostly concentrated below 0.5. For the winners, while a portion exhibits higher mutual consistency (greater than 0.5), another portion shows mutual consistency comparable to that of the losers. This suggests that during the early stage of motion planning, the competition among candidate regions is generally intense and highly unstable. Only a fraction of the regions can establish their positions early on (i.e., those with higher mutual consistency). By denoising step 10, motion planning is complete, and the winners and losers have separated into two distinct clusters. This also validates that the mutual consistency metric effectively distinguishes winners from losers.

In Section~\ref{sec:sa_interp}, we visualize the denoising dynamics corresponding to the video generated with seed 20. Here, we additionally provide the denoising dynamics for the video generated with seed 26, as shown in Figure~\ref{fig:candidate_evolution_seed26}. In this video, the motion trajectory of the basketball generally adheres to physical laws. When observing the candidate regions in early denoising stages, we notice that at denoising step 3, region K1 in frame 4 exhibits a higher mutual consistency than K2 (0.44 vs. 0.37), despite being the strongest loser. However, by step 4, K1 is overtaken by K2 (0.34 vs. 0.62), and its region size becomes noticeably smaller than that of K2. This serves as another manifestation of the high instability of object motion trajectories during motion planning.

\begin{figure}[htbp]
\centering
\includegraphics[width=1.0\columnwidth]{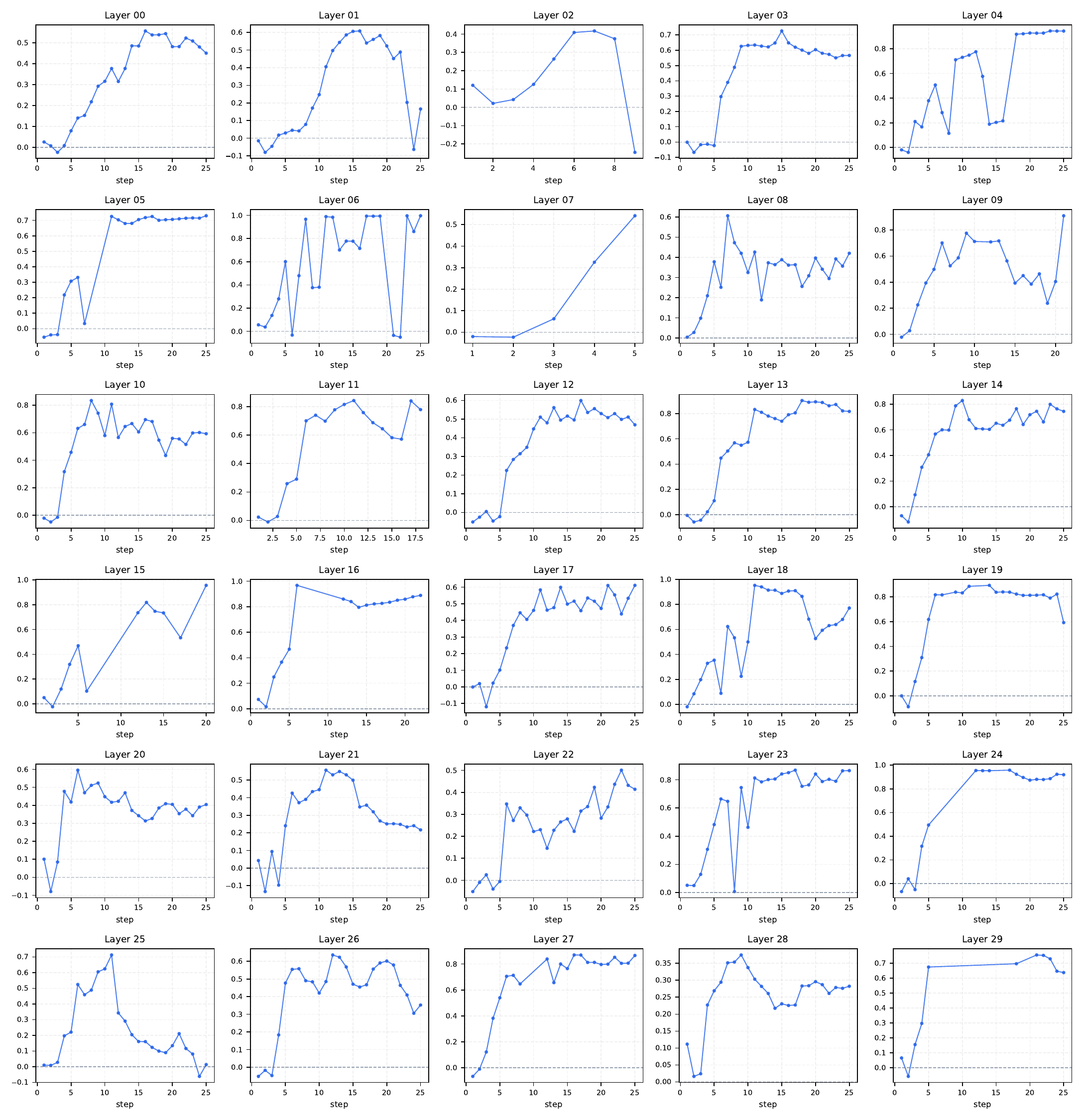}
\caption{The winner-loser gap of mutual consistency (i.e. the difference between the average mutual consistency of the winners in all frames and that of the strongest losers) with denoising steps in all layers. We only show the first 25 denoising steps for convenience since the mutual consistency doesn't change significantly in the later stages of denoising. As can be seen, during the first few denoising steps, the difference in mutual consistency between the winner and the strongest loser is often less than zero, meaning that the confidence at the object position in the final trajectory can be lower than that of other regions. This indicates that during the early motion planning phase, candidate regions undergo a highly sensitive competitive phase.}
\label{fig:mutual_consistency_layerwise}
\vspace{-10pt}
\end{figure}

\section{More Details on RoPE Modification}
\label{appn:rope_modify}

\subsection{Training Details}
\label{appn:rope_modify_train}

\subsubsection{General Settings}
\label{appn:rope_modify_train_general_settings}
For LoRA fine-tuning, we set the batch size to 32. The learning rate is set to 1e-4 with a warmup ratio of 0.03, remaining constant after reaching the maximum. We use the Adam optimizer with $\beta_1=0.9, \beta_2=0.999, \text{eps}=\text{1e-8}$. The default number of training steps is 1,500 steps (1 epoch), with the random seed set to 42. Training is conducted on 4 $\times$ A800 80G GPUs. In evaluation, we select the checkpoint at 800 steps, where the performance of the model peaks. For VideoREPA, following the settings of \citet{videorepa}, we adopt VideoMAEv2 \citep{videomae} as the alignment target encoder and set the alignment depth to 18. Other LoRA configurations are provided in Appendix~\ref{appn:rope_modify_train_lora_config}.

\subsubsection{LoRA Configurations}
\label{appn:rope_modify_train_lora_config}
For LoRA, the rank $r$ and alpha $\alpha$ are set to 64 and 32, respectively. For trainable modules, we compare training only cross- and self-attention (i.e., their respective $W_Q, W_K, W_V, W_O$) against training both attention and FFN (including $W_{in}$ and $W_{out}$). The rationale behind this design is that our research focuses on cross- and self-attention rather than FFN. This stems from our hypothesis that modules more relevant to motion planning are attention (especially self-attention responsible for inter-frame interactions) rather than FFN. \citet{ffn_interp} also present a similar perspective that FFNs in language models function as key-value memories, predominantly responsible for the storage of static knowledge rather than dynamic interactions among tokens. Nevertheless, to ensure the rigor of our study, we investigate whether FFN also exerts a non-negligible influence on the physical commonsense of the model.

\begin{figure}[htbp]
\centering

\begin{subfigure}{\linewidth}
    \includegraphics[width=\columnwidth]{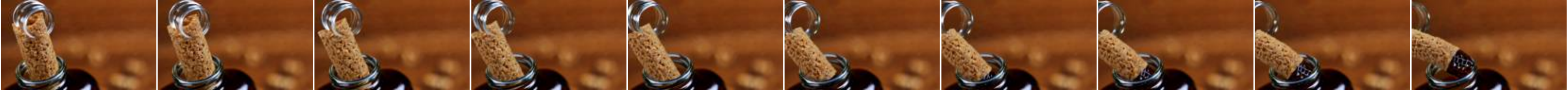}
    \caption{LoRA trainable modules: cross- and self-attention.}
\end{subfigure}
\vspace{5pt}

\begin{subfigure}{\linewidth}
    \includegraphics[width=\columnwidth]{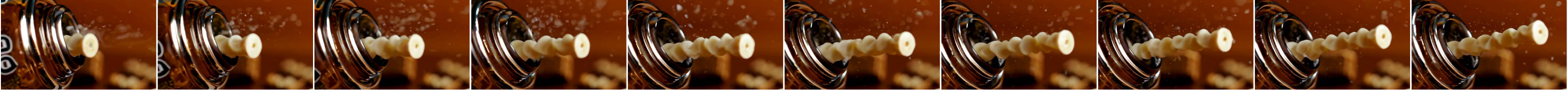}
    \caption{LoRA trainable modules: cross- and self-attention + FFN.}
\end{subfigure}

\caption{``Cork being twisted out of a bottle''. Under identical settings, additionally training FFN during LoRA fine-tuning tends to degrade aesthetic attributes such as the shapes of objects.}
\label{fig:lora_ffn_ablation}
\end{figure}

Through comparison, we observe that additionally fine-tuning the parameters of FFN not only fails to improve the motion trajectories of objects in the generated videos, but also degrades aesthetics, particularly the static properties of objects. As shown in the example in Figure~\ref{fig:lora_ffn_ablation}, the ``attention + FFN'' setting leads to distortion in the shape of the cork, while it still exhibits rotational motion. We attribute this to the fact that FFN mainly filters features in hidden states via non-linear transformations, predominantly governing the appearance of objects in videos rather than motion trajectories. Updating the parameters of FFN during fine-tuning introduces unnecessary alterations to the static knowledge stored in FFN, thereby undermining aesthetics. In contrast, restricting parameter updates solely to attention modules improves the process of inter-frame interactions without impairing the inherent shapes of objects in generated videos, robustly optimizing the motion trajectories of objects. This finding further corroborates the validity of our interpretability hypothesis: modules associated with the physical commonsense of the model are predominantly attention rather than FFN.

\subsubsection{\texorpdfstring{Parameterization of $\lambda^{h/w}$ for RoPE Modification}{Parameterization of RoPE modification}}
\label{appn:rope_modify_train_lambda}

In the main text, we primarily discuss using a fixed $\lambda^{h/w}$ during training. In fact, we explore different parameterization methods of $\lambda^{h/w}$. Here we mainly discuss two options: (1) fixed $\lambda^{h/w}$, which uses a fixed $\lambda^{h/w}$; (2) learnable $\lambda^{h/w}$, which allows the model to adaptively adjust $\lambda^{h/w}$ across denoising steps and self-attention heads guided by the flow-matching loss during training.

\paragraph{Fixed $\lambda^{h/w}$:} This method determines the value of $\lambda^{h/w}$ before training and keeps it unchanged during training. Specifically, given a value $\lambda$ of $\lambda^{h/w}$, if the sampled timestep falls within the initial 10\% timesteps of the denoising process during training, we apply RoPE modification with $\lambda^{h/w}=\lambda$ to the self-attention of the model. Otherwise, no RoPE modification is applied. This design accounts for the fact that motion planning predominantly occurs during the initial 10\% of the denoising process. In practice, for the selection of $\lambda^{h/w}$, we evaluate 0.55, 0.60, 0.65, 0.70, 0.75, 0.80, 0.85, and 0.90, and test across multiple random seeds on the ``basketball free fall'' case as well as examples selected from the training set of VideoPhy. Based on generation performance, we ultimately set $\lambda^{h/w}$ to 0.75.

\paragraph{Learnable \textbf{$\lambda^{h/w}$}:} This method aims to enable the model to learn to adjust the value of $\lambda^{h/w}$ by itself. The key idea is to replace a globally fixed spatial RoPE frequency with head-specific and timestep-conditioned scaling factors. This allows different self-attention heads to learn different degrees of spatial locality relaxation at different diffusion timesteps. Specifically, we replace $\lambda^{h/w}$ in fixed $\lambda^{h/w}$ with:
\begin{equation}
    \lambda_{\ell, k}^{a} (\tau) = \exp(\mu_{\ell, k}^{a} + g_\psi(e_\tau))
\end{equation}
where $a \in \{h, w\}$ denotes the dimension (height / width) to which RoPE modification is applied, $\ell$ is the layer index, and $k$ is the index of the self-attention head. $\mu_{\ell, k}$ is the base modulation factor of head $k$ in layer $\ell$, initialized to 0. $\tau$ denotes the timestep, $e_\tau$ is the sinusoidal embedding of the timestep, and $g_\psi(\cdot)$ is a single-layer timestep-conditioned MLP, i.e., $g_\psi(e_\tau)=W_2\,\mathrm{SiLU}(W_1 e_\tau+b_1)+b_2$, where $W_2=0, b_2=0$ upon initialization. Therefore, $\lambda_{\ell, k}^{a} (\tau)$ is initialized to 1, identical to the original model. During training, we expect the model to automatically identify suitable $\lambda^{h/w}$ for each self-attention head and timestep under the guidance of the loss function.

\begin{figure}[htbp]
\centering
\begin{subfigure}{0.325\linewidth}
    \includegraphics[width=\linewidth]{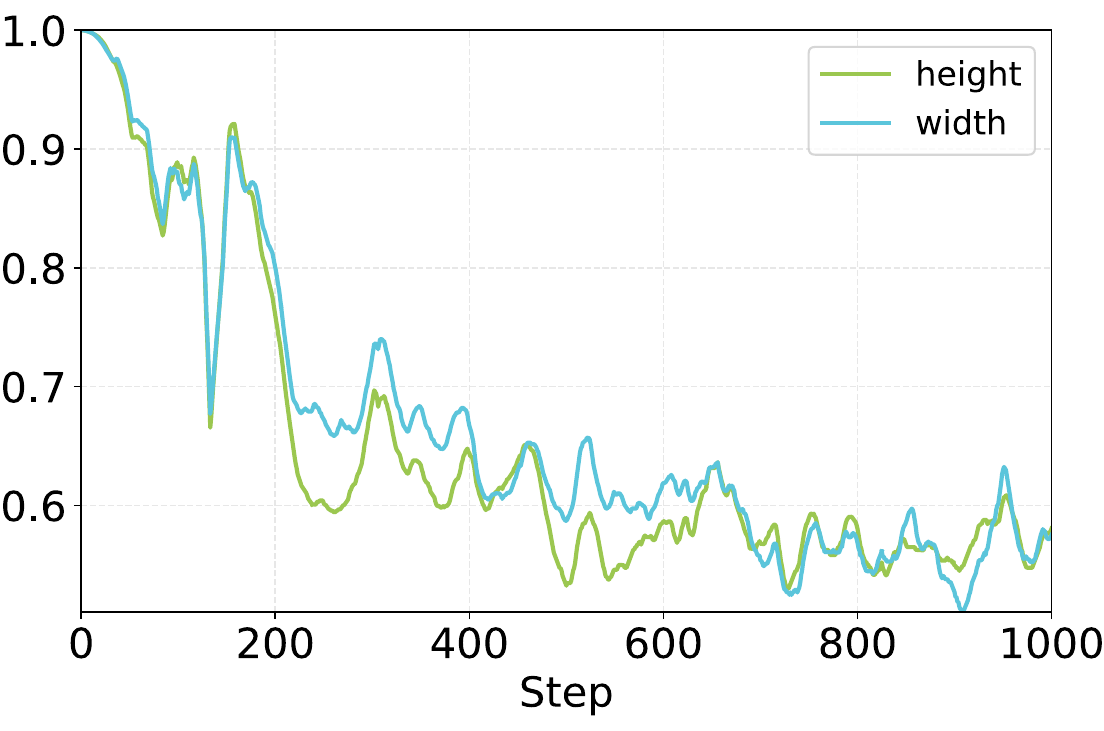}
    \caption{$\lambda^{h/w}$ min}
\end{subfigure}
\begin{subfigure}{0.325\linewidth}
    \includegraphics[width=\linewidth]{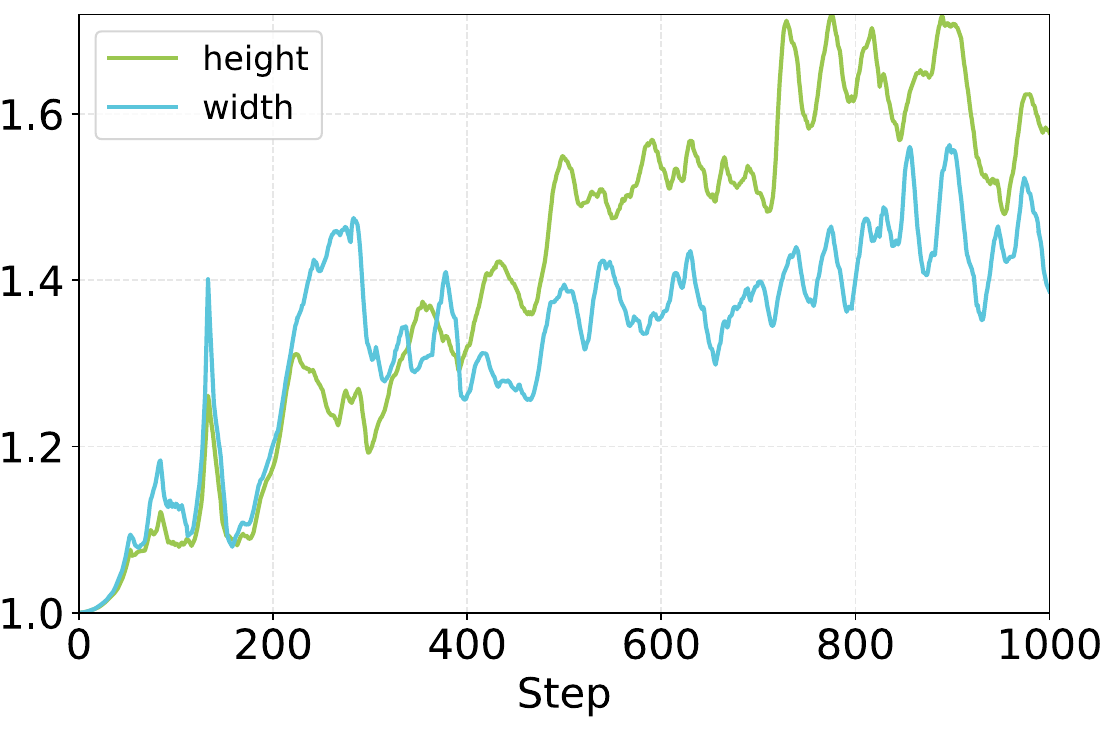}
    \caption{$\lambda^{h/w}$ max}
\end{subfigure}
\begin{subfigure}{0.325\linewidth}
    \includegraphics[width=\linewidth]{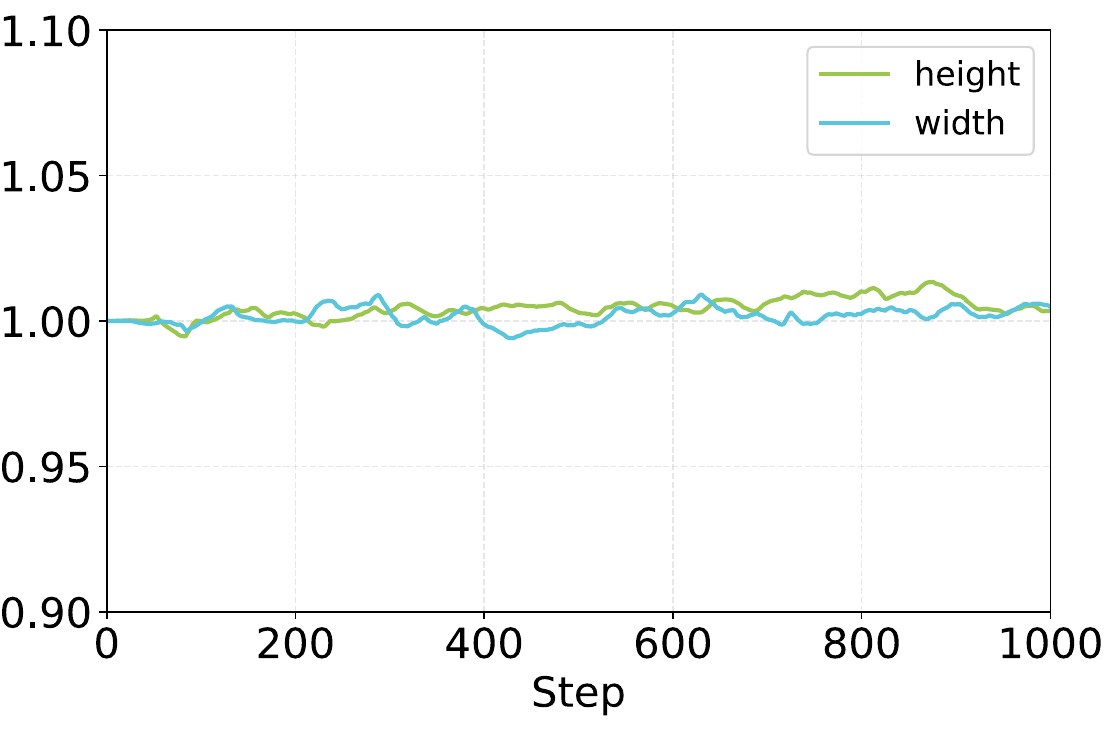}
    \caption{$\lambda^{h/w}$ mean}
\end{subfigure}
\caption{The minimum, maximum, and average values of $\lambda^{h/w}$ during training. Values of $\lambda^{h/w}$ are sampled at denoising timestep=900 (out of 1,000 steps), and trends of $\lambda^{h/w}$ across other timesteps are similar. The minimum and maximum values of $\lambda^{h/w}$ continuously decrease/increase, while the average value remains stable around 1.0, indicating that approximately half of the self-attention heads adjust their $\lambda^{h/w}$ to values less than 1.}
\label{fig:learnable_lambda}
\end{figure}

However, in practice, we find that this method deviates from expectations: as shown in Figure~\ref{fig:learnable_lambda}, during training, $\lambda^{h/w}$ of approximately half of the self-attention heads is less than 1, while that of the other half is greater than 1. Moreover, the physical commonsense of the final trained model shows no significant improvement. 

We argue that this phenomenon further underscores that the current model architecture and flow-matching loss lack sufficient capture of the motion trajectories of objects in videos. From the perspective of frame differences between adjacent frames, the motion information of objects can be regarded as subtle displacements from frame $f$ to $f+1$. Such motion information accounts for a minor proportion of the entire frame information that the model needs to predict. Consequently, even if we grant adjustable flexibility to the RoPE frequency of self-attention in the model, parameter updates of the model are predominantly influenced by gradients from information other than motion, thereby hindering effective learning of the motion patterns of objects. Therefore, based on our findings in Section~\ref{sec:sa_interp}, we consider it necessary to enforce the model to explore more candidate positions of objects in each frame via RoPE modification during early denoising stages.

\subsubsection{Custom Timestep Sampler}
\label{appn:rope_modify_train_custom_sampler}
According to our interpretability research, motion planning occurs during the initial 10\% of denoising timesteps, while the remaining denoising process is predominantly responsible for detail filling. Therefore, we design a dedicated timestep sampler to concentrate training on early denoising stages. Unlike uniform sampling, we sample the initial 10\% of denoising steps with probability $p^{\mathrm{early}}$ and the remaining steps with probability $1-p^{\mathrm{early}}$. 
In practice, for the value of $p^{\mathrm{early}}$, we explore 0.3, 0.5, 0.7, 0.9, and 1.0, ultimately setting $p^{\mathrm{early}}$ to 0.9. We observe that setting $p^{\mathrm{early}}$ excessively high degrades the aesthetics of generated videos, as the fine-tuning process lacks training samples from mid-to-late denoising stages for regularization, causing the model to lose its denoising capability for mid-to-late stages. Conversely, setting $p^{\mathrm{early}}$ too low (less than or equal to 0.5) leads to distortion in the motion trajectories of objects in generated videos. This likely occurs because an overly small $p^{\mathrm{early}}$ exposes the model to too few early-stage denoising samples with RoPE modification, resulting in insufficient training where samples with RoPE modification act as injected noise instead. Setting $p^{\mathrm{early}}$ to 0.9 allows the model to fully adapt to the altered RoPE frequencies without losing its capability to perform denoising in mid-to-late stages.

\subsection{Data Processing and Evaluation Details}
\label{appn:rope_modify_data}
\paragraph{Training data processing:} For training, we use WISA \citep{wisa}, a dataset comprising 80,000 manually collected videos that depict 17 fundamental physical laws across three domains of physics: dynamics, thermodynamics, and optics. For each video, physical information is annotated beyond the caption, including textual physical descriptions, qualitative physics categories, and quantitative physical properties. Although the authors conducted multiple rounds of filtering on the data, we still apply further strict filtering according to the following rules to ensure the reliability of the training data:

\begin{itemize}
    \item For all videos, we remove samples where the \texttt{duration} label is empty, as the length of these videos is typically under one second and the quality is very low. Furthermore, we require the duration of the video to be greater than 2.0s to ensure the video contains sufficient content. In addition, since the duration and resolution of videos in the dataset vary, and the default length of videos generated by the Wan2.1-T2V is 81 frames (5s, 16FPS), we adopt the following frame sampling scheme: crop the first 5 seconds of the video (no cropping for videos shorter than 5 seconds) and uniformly sample 81 frames. Videos with fewer than 81 sampled frames are discarded.
    \item For videos in the domain of dynamics, we additionally remove samples labeled as ``no obvious dynamic phenomenon'', and require $\texttt{motion\_score} > 0.10$ and $0.01 < \texttt{motion\_score\_v2} < 6.50$ to ensure that the video content exhibits distinct yet unexaggerated motion amplitudes.
    \item For the reflection subcategory in optics, since the sample count of this category is larger than that of other subcategories, we only sample 4,000 videos to ensure a balanced distribution of data samples.
\end{itemize}

After filtering, we obtain approximately 48,000 high-quality videos for training.

\paragraph{Evaluation details:} We first perform quick verification across multiple random seeds on the ``basketball free fall'' case to observe whether our method improves the motion trajectory of the basketball. For training-free RoPE modification, we manually tune the value of $\lambda^{h/w}$ within the range of [0.50, 0.90] to achieve the best generation quality. Since this method cannot scale in practice, we subsequently fine-tune the model on the basis of RoPE modification and use a fixed $\lambda^{h/w}=0.70$ in evaluation. It is worth noting that the value of $\lambda^{h/w}$ at test time does not need to be identical to the value during training (0.75 by default), and 0.70 is an empirically suitable value.

Subsequently, we conduct systematic validation with VideoPhy \citep{videophy}, a benchmark for physical commonsense in video generative models. The test set of VideoPhy contains 343 test samples, covering interaction scenarios among three real-world material types: solid-solid, solid-fluid, and fluid-fluid. The original test samples are relatively short prompts. For prompt refinement, we refer to PhyT2V \citep{phyt2v} to rewrite the prompts, expanding descriptions of the objects and scenes involved in the prompts as well as the motion process. 

For evaluation metrics, following the requirements of VideoPhy, we use Semantic Adherence (SA) and Physical Commonsense (PC). Specifically, SA evaluates the instruction-following ability of the model, while PC evaluates whether the motion trajectories of objects in the generated videos follow real-world physical laws. Both SA and PC are binary metrics, with values of 1 or 0. VideoPhy adopts fine-tuned vision-language models and closed-source models for automated evaluation. However, we find that such automated evaluation methods introduce significant bias. We observe severe hallucination issues in the judge model during video understanding, leading to low correlation with human evaluation results. Therefore, to ensure the rigor of the evaluation, we conduct human evaluation for all baselines and our method. To ensure fairness, human evaluators are blind to the version of the model and are only asked to score across the two dimensions of SA and PC. Additionally, we evaluate whether the aesthetics of generated videos show significant changes compared with the original model. We instruct evaluators to mark samples with noticeable degradation in aesthetic quality as the basis of ablation studies.

For the evaluation on VideoPhy, all videos are sampled with a random seed of 42. For our proposed RoPE modification method, we set $\lambda^{h/w}$ to 0.70 for both training-free and training-based settings.

\subsection{Visualizations}
\label{appn:rope_modify_viz}

\subsubsection{Results on the basketball falling case}

\begin{figure}[htbp]
\centering

\begin{subfigure}{\columnwidth}
    \begin{subfigure}{1.00\linewidth}
        \includegraphics[width=\columnwidth]{iclr2027/images/video_viz/basketball_seed8.pdf}
    \end{subfigure}

    \begin{subfigure}{1.00\linewidth}
        \includegraphics[width=\columnwidth]{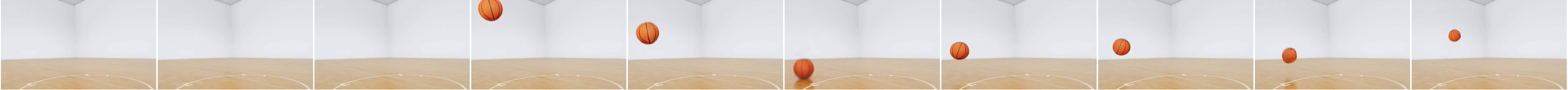}
    \end{subfigure}
\caption{seed = 8, $\lambda^{h/w}=0.55$}
\end{subfigure}
\\[7pt]

\begin{subfigure}{\columnwidth}
    \begin{subfigure}{0.99\linewidth}
        \includegraphics[width=\linewidth]{iclr2027/images/video_viz/basketball_seed20.pdf}
    \end{subfigure}
    \\
    \begin{subfigure}{0.99\linewidth}
        \includegraphics[width=\linewidth]{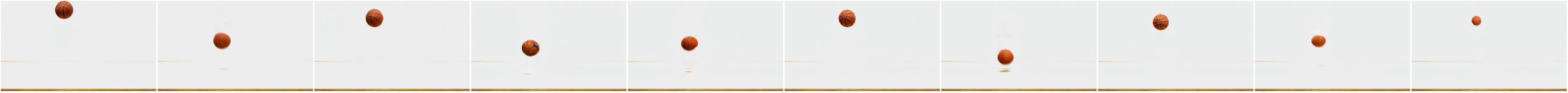}
    \end{subfigure}
\caption{seed = 20, $\lambda^{h/w}=0.60$}
\end{subfigure}
\\[7pt]

\begin{subfigure}{\columnwidth}
    \begin{subfigure}{0.99\linewidth}
        \includegraphics[width=\linewidth]{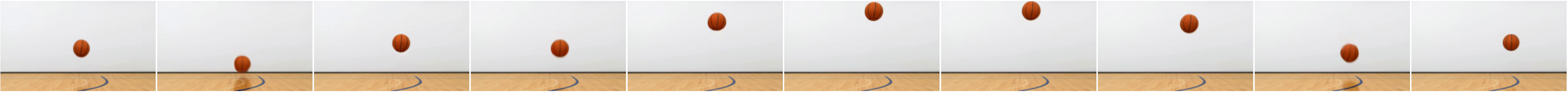}
    \end{subfigure}
    \\
    \begin{subfigure}{0.99\linewidth}
        \includegraphics[width=\linewidth]{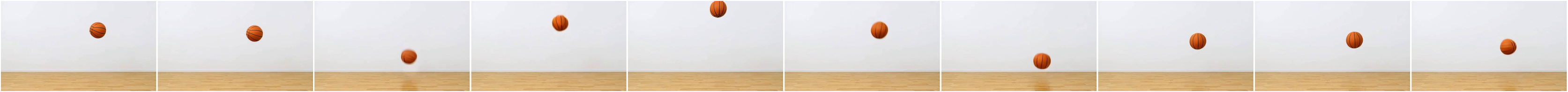}
    \end{subfigure}
\caption{seed = 23, $\lambda^{h/w}=0.85$}
\end{subfigure}
\\[7pt]

\begin{subfigure}{\columnwidth}
    \begin{subfigure}{1.00\linewidth}
        \includegraphics[width=\columnwidth]{iclr2027/images/video_viz/basketball_seed29.pdf}
    \end{subfigure}
    
    \begin{subfigure}{1.00\linewidth}
        \includegraphics[width=\columnwidth]{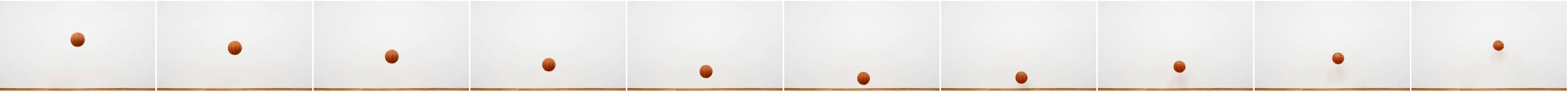}
    \end{subfigure}
\caption{seed = 29, $\lambda^{h/w}=0.50$}
\end{subfigure}

\caption{(Prompt) \textit{``Against a pure white background, a basketball falls vertically from mid-air onto a wooden floor and bounces up several times.''} Each subplot corresponds to the generation result of a random seed, where Top: Wan2.1-T2V-1.3B and Bottom: Wan2.1-T2V-1.3B-modified RoPE. This training-free method requires manually tuning the value of $\lambda^{h/w}$ for each case to achieve optimal results, and the generated content often suffers from instability in terms of motion trajectories and object permanence.}
\label{fig:rope_res_basketball}
\end{figure}

\begin{figure}[!t]
\centering

\begin{subfigure}{\columnwidth}
    \begin{subfigure}{1.00\linewidth}
        \includegraphics[width=\columnwidth]{iclr2027/images/video_viz/basketball_seed8.pdf}
    \end{subfigure}
    
    \begin{subfigure}{1.00\linewidth}
        \includegraphics[width=\columnwidth]{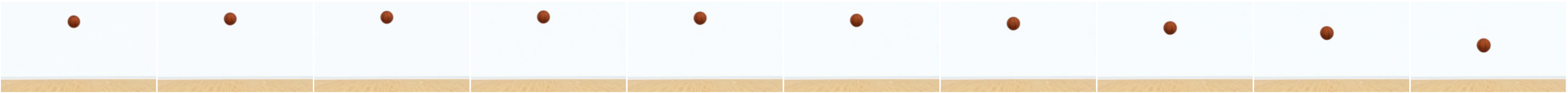}
    \end{subfigure}
\caption{seed 8}
\end{subfigure}
\\[7pt]

\begin{subfigure}{\columnwidth}
    \begin{subfigure}{0.99\linewidth}
        \includegraphics[width=\linewidth]{iclr2027/images/video_viz/basketball_seed20.pdf}
    \end{subfigure}
    \\
    \begin{subfigure}{0.99\linewidth}
        \includegraphics[width=\linewidth]{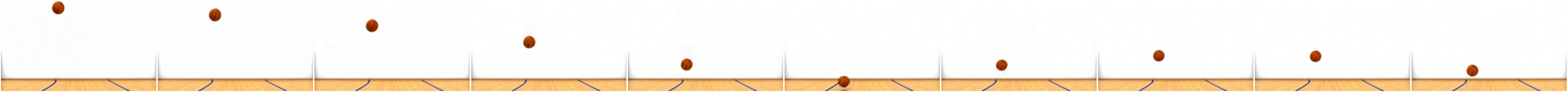}
    \end{subfigure}
\caption{seed 20}
\end{subfigure}
\\[7pt]

\begin{subfigure}{\columnwidth}
    \begin{subfigure}{0.99\linewidth}
        \includegraphics[width=\linewidth]{iclr2027/images/video_viz/basketball_seed23_.pdf}
    \end{subfigure}
    \\
    \begin{subfigure}{0.99\linewidth}
        \includegraphics[width=\linewidth]{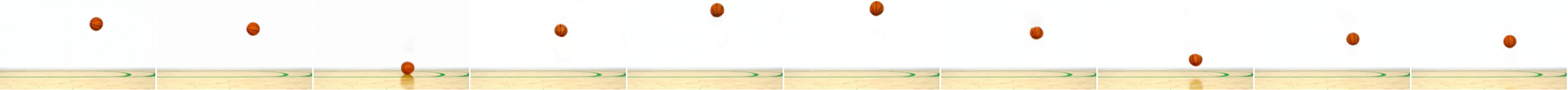}
    \end{subfigure}
\caption{seed 23}
\end{subfigure}
\\[7pt]

\begin{subfigure}{\columnwidth}
    \begin{subfigure}{1.00\linewidth}
        \includegraphics[width=\columnwidth]{iclr2027/images/video_viz/basketball_seed29.pdf}
    \end{subfigure}
    
    \begin{subfigure}{1.00\linewidth}
        \includegraphics[width=\columnwidth]{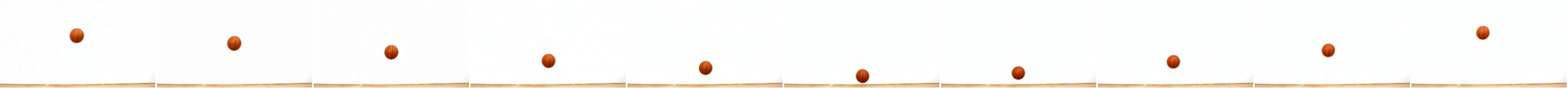}
    \end{subfigure}
\caption{seed 29}
\end{subfigure}

\caption{(Prompt) \textit{``Against a pure white background, a basketball falls vertically from mid-air onto a wooden floor and bounces up several times.''} Each subplot corresponds to the generation result of a random seed, where Top: Wan2.1-T2V-1.3B and Bottom: Wan2.1-T2V-1.3B-LoRA+modified RoPE. During inference, we apply the RoPE modification with $\lambda^{h}=\lambda^{w}=0.70$ on the model trained with modified RoPE during the first 5 steps of denoising. It can be seen that the results of the training-based method are better than those of the training-free method.}
\label{fig:rope_res_train_basketball}
\end{figure}

\clearpage
\subsubsection{Results on several samples from VideoPhy}
\label{appn:rope_modify_viz_videophy}
\vspace{20pt}

\begin{figure}[htbp]
\centering

\begin{subfigure}{\linewidth}
    \includegraphics[width=\columnwidth]{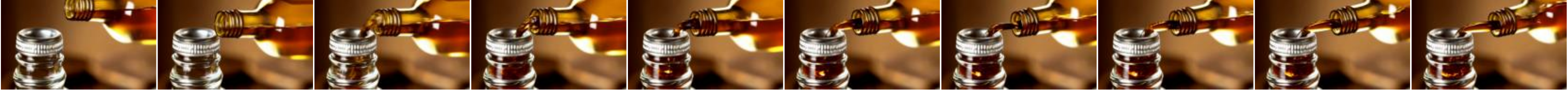}
    \caption{Wan2.1-T2V-1.3B}
\end{subfigure}
\vspace{3pt}

\begin{subfigure}{\linewidth}
    \includegraphics[width=\columnwidth]{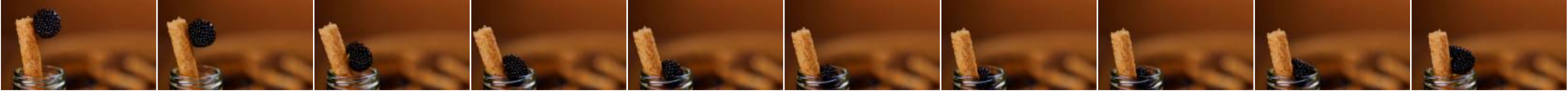}
    \caption{Wan2.1-T2V-1.3B-modified RoPE}
\end{subfigure}
\vspace{3pt}

\begin{subfigure}{\linewidth}
    \includegraphics[width=\columnwidth]{iclr2027/images/wan_eval_videophy/case14_bsz_32-lora_rank_64_alpha_32_modules_attn-lambda_0.75-mixed_0.1_0.9-ckpt_step_800_lambda_steps_1-2-3-4-5_lambda_manual_0.70_0.70.pdf.pdf}
    \caption{Wan2.1-T2V-1.3B-LoRA+modified RoPE}
\end{subfigure}

\caption{(Prompt) \textit{``Cork being twisted out of a bottle.''}}
\end{figure}
\vspace{20pt}

\begin{figure}[htbp]
\centering

\begin{subfigure}{\linewidth}
    \includegraphics[width=\columnwidth]{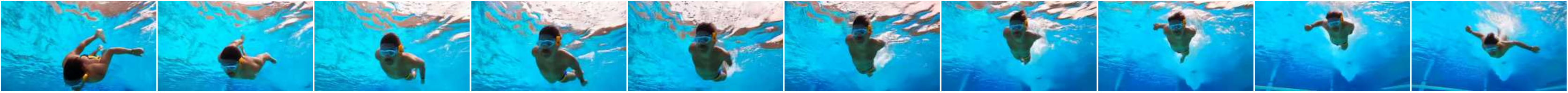}
    \caption{Wan2.1-T2V-1.3B}
\end{subfigure}
\vspace{3pt}

\begin{subfigure}{\linewidth}
    \includegraphics[width=\columnwidth]{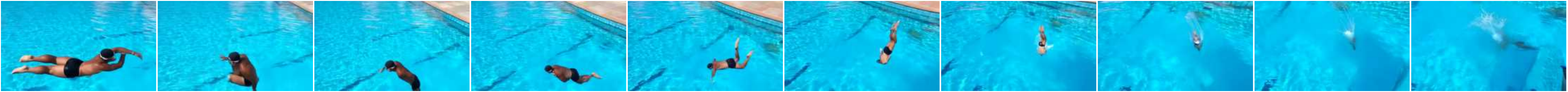}
    \caption{Wan2.1-T2V-1.3B-modified RoPE}
\end{subfigure}
\vspace{3pt}

\begin{subfigure}{\linewidth}
    \includegraphics[width=\columnwidth]{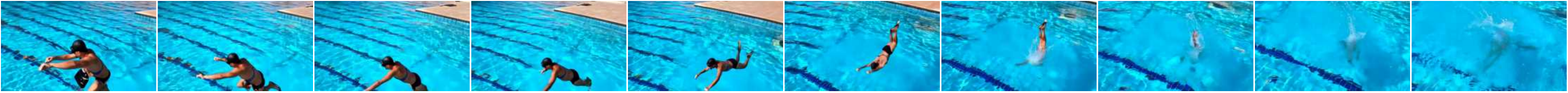}
    \caption{Wan2.1-T2V-1.3B-LoRA+modified RoPE}
\end{subfigure}

\caption{(Prompt) \textit{``A diver plunges headlong into a sparkling pool.''}}
\end{figure}
\vspace{20pt}

\begin{figure}[htbp]
\centering

\begin{subfigure}{\linewidth}
    \includegraphics[width=\columnwidth]{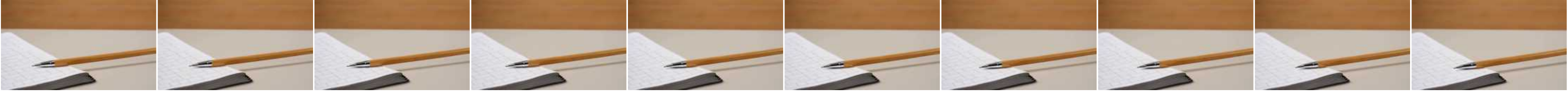}
    \caption{Wan2.1-T2V-1.3B}
\end{subfigure}
\vspace{3pt}

\begin{subfigure}{\linewidth}
    \includegraphics[width=\columnwidth]{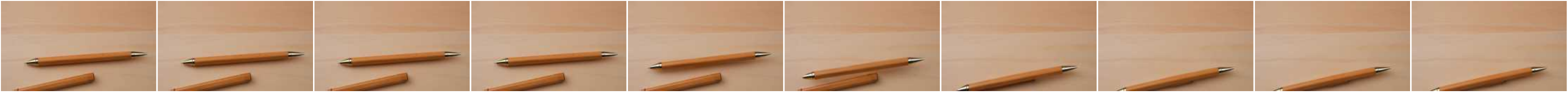}
    \caption{Wan2.1-T2V-1.3B-modified RoPE}
\end{subfigure}
\vspace{3pt}

\begin{subfigure}{\linewidth}
    \includegraphics[width=\columnwidth]{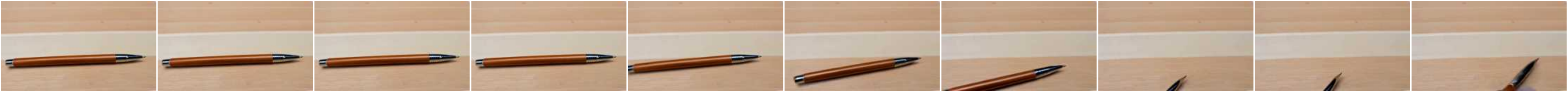}
    \caption{Wan2.1-T2V-1.3B-LoRA+modified RoPE}
\end{subfigure}

\caption{(Prompt) \textit{``Wooden pencil rolls around on a flat desk.''}}
\end{figure}

\begin{figure}[htbp]
\centering

\begin{subfigure}{\linewidth}
    \includegraphics[width=\columnwidth]{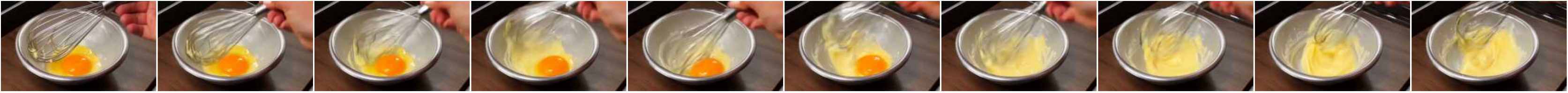}
    \caption{Wan2.1-T2V-1.3B}
\end{subfigure}
\vspace{3pt}

\begin{subfigure}{\linewidth}
    \includegraphics[width=\columnwidth]{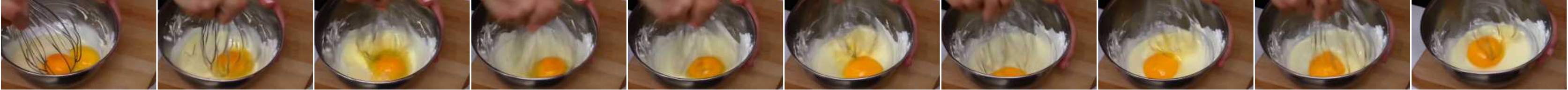}
    \caption{Wan2.1-T2V-1.3B-modified RoPE}
\end{subfigure}
\vspace{3pt}

\begin{subfigure}{\linewidth}
    \includegraphics[width=\columnwidth]{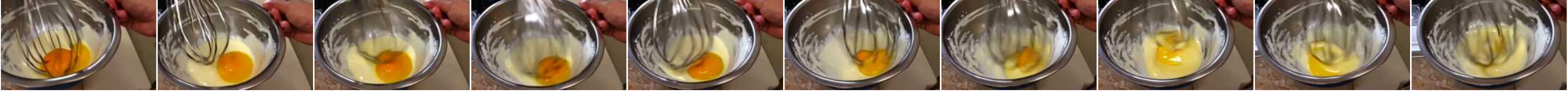}
    \caption{Wan2.1-T2V-1.3B-LoRA+modified RoPE}
\end{subfigure}

\caption{(Prompt) \textit{``A whisk mixes an egg in a bowl.''}}
\end{figure}

\begin{figure}[htbp]
\centering

\begin{subfigure}{\linewidth}
    \includegraphics[width=\columnwidth]{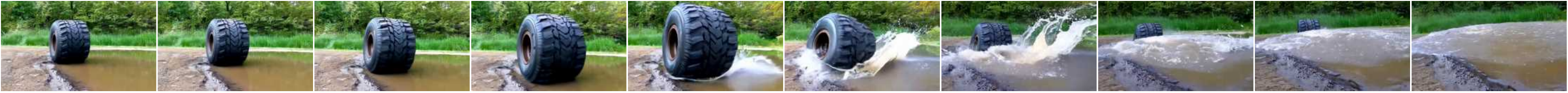}
    \caption{Wan2.1-T2V-1.3B}
\end{subfigure}
\vspace{3pt}

\begin{subfigure}{\linewidth}
    \includegraphics[width=\columnwidth]{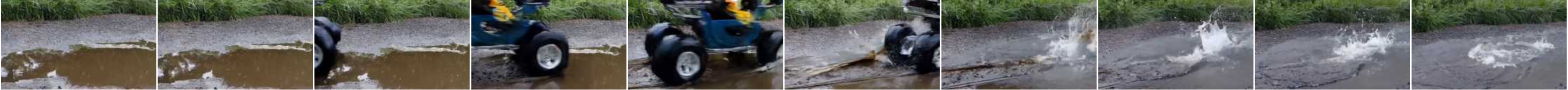}
    \caption{Wan2.1-T2V-1.3B-modified RoPE}
\end{subfigure}
\vspace{3pt}

\begin{subfigure}{\linewidth}
    \includegraphics[width=\columnwidth]{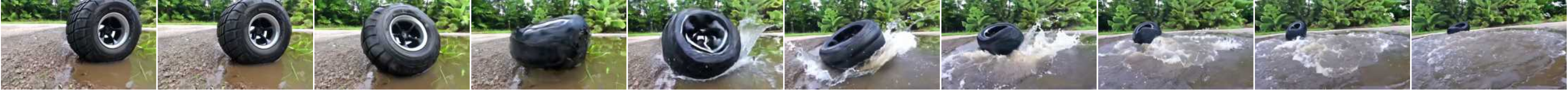}
    \caption{Wan2.1-T2V-1.3B-LoRA+modified RoPE}
\end{subfigure}

\caption{(Prompt) \textit{``A tyre rolls through a large puddle, splashing water.''}}
\end{figure}

\begin{figure}[htbp]
\centering

\begin{subfigure}{\linewidth}
    \includegraphics[width=\columnwidth]{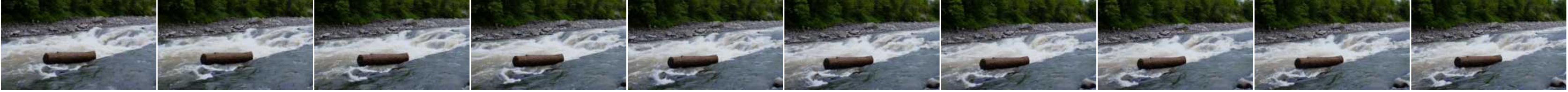}
    \caption{Wan2.1-T2V-1.3B}
\end{subfigure}
\vspace{3pt}

\begin{subfigure}{\linewidth}
    \includegraphics[width=\columnwidth]{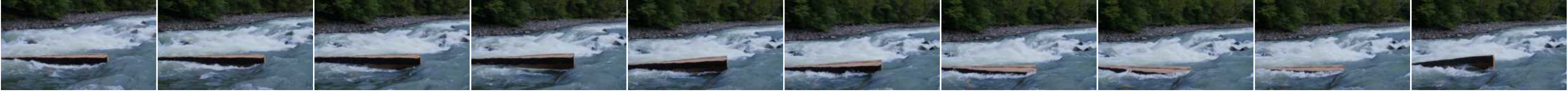}
    \caption{Wan2.1-T2V-1.3B-modified RoPE}
\end{subfigure}
\vspace{3pt}

\begin{subfigure}{\linewidth}
    \includegraphics[width=\columnwidth]{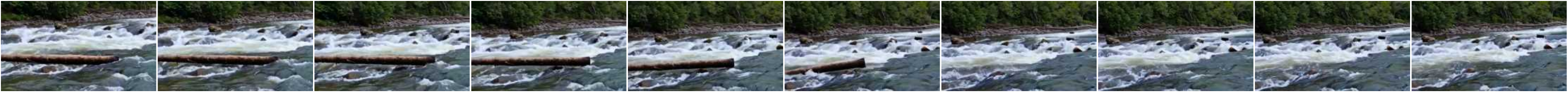}
    \caption{Wan2.1-T2V-1.3B-LoRA+modified RoPE}
\end{subfigure}

\caption{(Prompt) \textit{``A large log floats downstream in a rushing river.''}}
\end{figure}

\begin{figure}[htbp]
\centering

\begin{subfigure}{\linewidth}
    \includegraphics[width=\columnwidth]{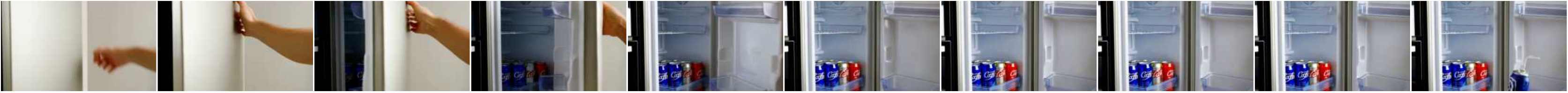}
    \caption{Wan2.1-T2V-1.3B}
\end{subfigure}
\vspace{3pt}

\begin{subfigure}{\linewidth}
    \includegraphics[width=\columnwidth]{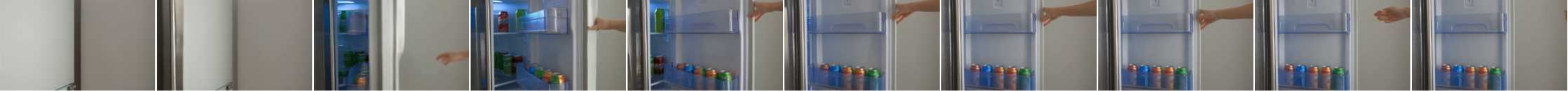}
    \caption{Wan2.1-T2V-1.3B-modified RoPE}
\end{subfigure}
\vspace{3pt}

\begin{subfigure}{\linewidth}
    \includegraphics[width=\columnwidth]{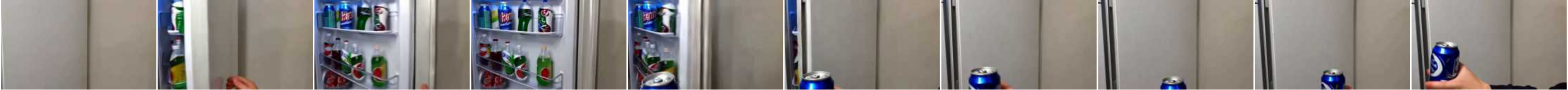}
    \caption{Wan2.1-T2V-1.3B-LoRA+modified RoPE}
\end{subfigure}

\caption{(Prompt) \textit{``Refrigerator door closing after getting a soda.''}}
\end{figure}

\begin{figure}[htbp]
\centering

\begin{subfigure}{\linewidth}
    \includegraphics[width=\columnwidth]{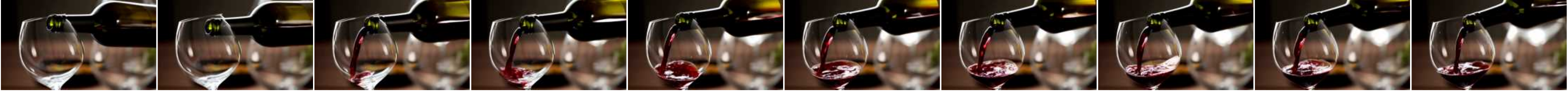}
    \caption{Wan2.1-T2V-1.3B}
\end{subfigure}
\vspace{3pt}

\begin{subfigure}{\linewidth}
    \includegraphics[width=\columnwidth]{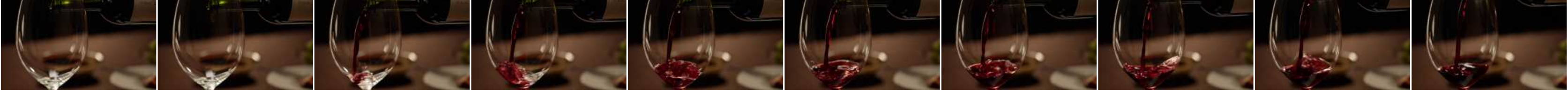}
    \caption{Wan2.1-T2V-1.3B-modified RoPE}
\end{subfigure}
\vspace{3pt}

\begin{subfigure}{\linewidth}
    \includegraphics[width=\columnwidth]{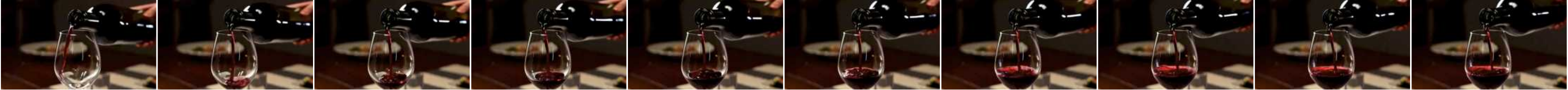}
    \caption{Wan2.1-T2V-1.3B-LoRA+modified RoPE}
\end{subfigure}

\caption{(Prompt) \textit{``Wine pouring from a bottle into a glass.''}}
\end{figure}

\begin{figure}[htbp]
\centering

\begin{subfigure}{\linewidth}
    \includegraphics[width=\columnwidth]{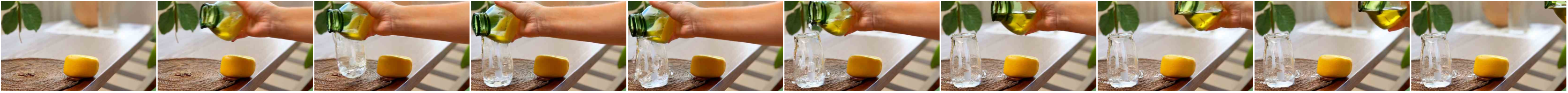}
    \caption{Wan2.1-T2V-1.3B}
\end{subfigure}
\vspace{3pt}

\begin{subfigure}{\linewidth}
    \includegraphics[width=\columnwidth]{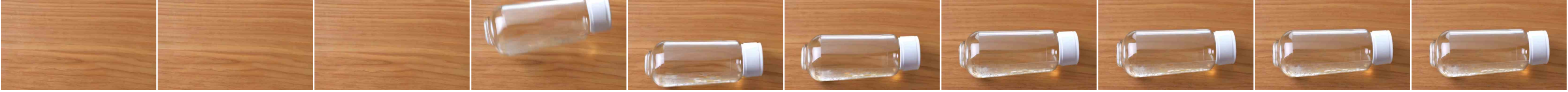}
    \caption{Wan2.1-T2V-1.3B-modified RoPE}
\end{subfigure}
\vspace{3pt}

\begin{subfigure}{\linewidth}
    \includegraphics[width=\columnwidth]{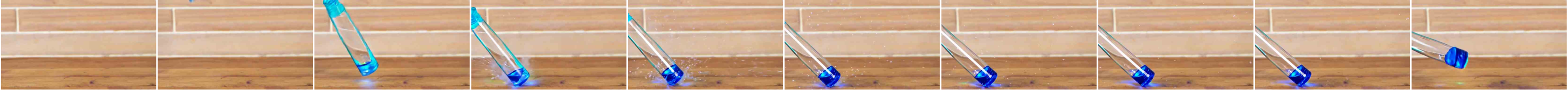}
    \caption{Wan2.1-T2V-1.3B-LoRA+modified RoPE}
\end{subfigure}

\caption{(Prompt) \textit{``Bottle topples off the table.''}}
\end{figure}

\clearpage
The following are videos generated by Wan2.1-T2V-14B, where Top: Wan2.1-T2V-14B, Bottom: Wan2.1-T2V-14B-modified RoPE ($\lambda^{h}=\lambda^{w}=0.70$).
\vspace{20pt}

\begin{figure}[htbp]
\centering

\begin{subfigure}{\linewidth}
    \includegraphics[width=\columnwidth]{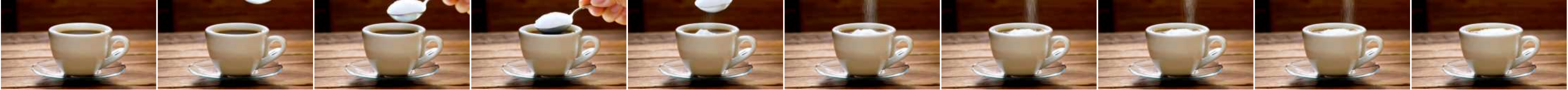}
\end{subfigure}
\vspace{3pt}

\begin{subfigure}{\linewidth}
    \includegraphics[width=\columnwidth]{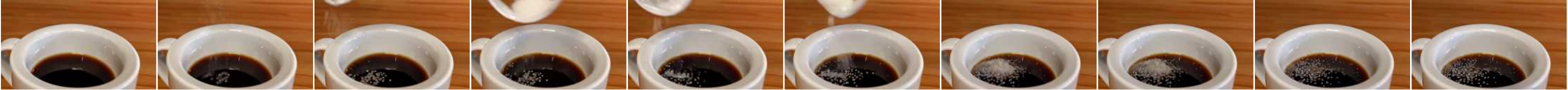}
\end{subfigure}

\caption{(Prompt) \textit{``A teaspoon stirs sugar into a cup of coffee.''}}
\end{figure}
\vspace{15pt}

\begin{figure}[htbp]
\centering

\begin{subfigure}{\linewidth}
    \includegraphics[width=\columnwidth]{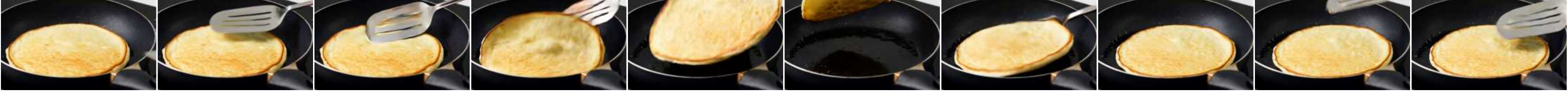}
\end{subfigure}
\vspace{3pt}

\begin{subfigure}{\linewidth}
    \includegraphics[width=\columnwidth]{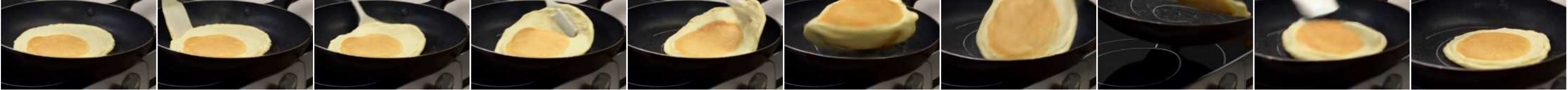}
\end{subfigure}

\caption{(Prompt) \textit{``Spatula flips pancake in air.''}}
\end{figure}
\vspace{15pt}

\begin{figure}[htbp]
\centering

\begin{subfigure}{\linewidth}
    \includegraphics[width=\columnwidth]{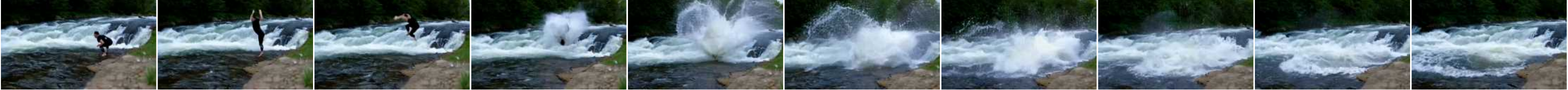}
\end{subfigure}
\vspace{3pt}

\begin{subfigure}{\linewidth}
    \includegraphics[width=\columnwidth]{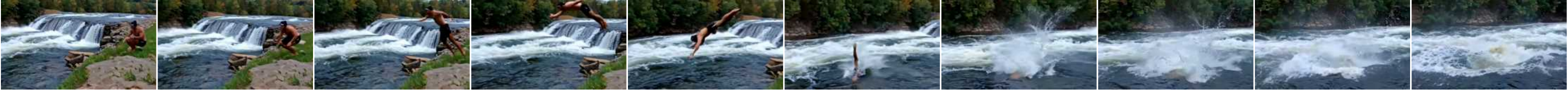}
\end{subfigure}

\caption{(Prompt) \textit{``A diver takes a plunge into a swift river.''}}
\end{figure}
\vspace{15pt}

\begin{figure}[!htbp]
\centering

\begin{subfigure}{\linewidth}
    \includegraphics[width=\columnwidth]{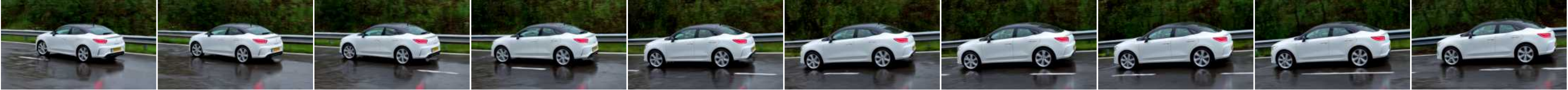}
\end{subfigure}
\vspace{3pt}

\begin{subfigure}{\linewidth}
    \includegraphics[width=\columnwidth]{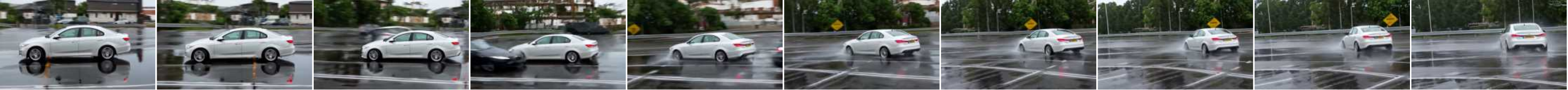}
\end{subfigure}

\caption{(Prompt) \textit{``A car gliding over a road slick with rainwater.''}}
\end{figure}
\vspace{15pt}

\begin{figure}[!htbp]
\centering

\begin{subfigure}{\linewidth}
    \includegraphics[width=\columnwidth]{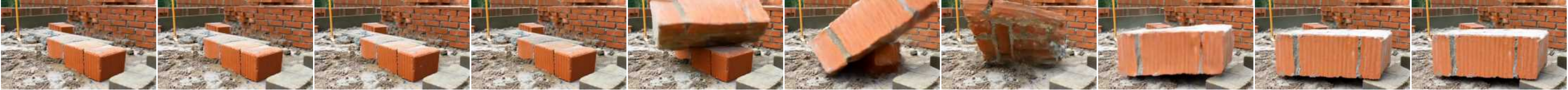}
\end{subfigure}
\vspace{3pt}

\begin{subfigure}{\linewidth}
    \includegraphics[width=\columnwidth]{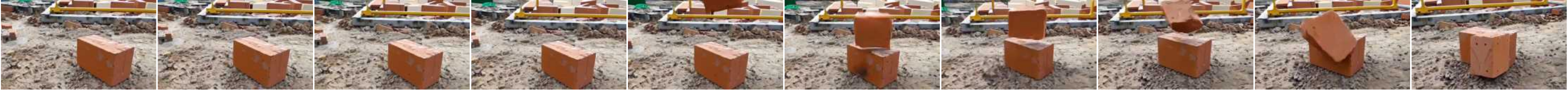}
\end{subfigure}

\caption{(Prompt) \textit{``Brick falling onto another brick.''}}
\end{figure}
\vspace{15pt}

\end{document}